\documentclass[10pt,journal,compsoc]{IEEEtran}

\usepackage{url}       
\usepackage{amsmath}
\usepackage{amssymb}
\usepackage{graphicx}
\usepackage{amsfonts}
\usepackage{bbm}
\usepackage{algorithm}
\usepackage{algorithmic}
\usepackage{parskip}
\usepackage[labelfont=bf,textfont=it]{caption}
\usepackage[marginclue,footnote,silent,draft]{fixme}
\usepackage{xcolor}
\usepackage{pgf,tikz,pgfplots}
\begin{document}
%
%
\title{Fourier-based and Rational Graph Filters\\ for Spectral Processing}
\author{Giuseppe Patan\'e 
\IEEEcompsocitemizethanks{\IEEEcompsocthanksitem G. Patan\'e is with CNR-IMATI, Consiglio Nazionale delle Ricerche, Istituto di Matematica Applicata e Tecnologie Informatiche
Genova, Italy.\protect\\ E-mail: patane@ge.imati.cnr.it}}
\IEEEtitleabstractindextext{
\begin{abstract}
Data are represented as graphs in a wide range of applications, such as Computer Vision (e.g., images) and Graphics (e.g., 3D meshes), network analysis (e.g., social networks), and bio-informatics (e.g., molecules). In this context, our overall goal is the definition of novel Fourier-based and graph filters induced by rational polynomials for graph processing, which generalise polynomial filters and the Fourier transform to non-Euclidean domains. For the efficient evaluation of discrete spectral Fourier-based and wavelet operators, we introduce a spectrum-free approach, which requires the solution of a small set of sparse, symmetric, well-conditioned linear systems and is oblivious of the evaluation of the Laplacian or kernel spectrum. Approximating arbitrary graph filters with rational polynomials provides a more accurate and numerically stable alternative with respect to polynomials. To achieve these goals, we also study the link between spectral operators, wavelets, and filtered convolution with integral operators induced by spectral kernels. According to our tests, main advantages of the proposed approach are (i) its generality with respect to the input data (e.g., graphs, 3D shapes), applications (e.g., signal reconstruction and smoothing, shape correspondence), and filters (e.g., polynomial, rational polynomial), and (ii) a spectrum-free computation with a generally low computational cost and storage overhead.
\end{abstract}
}
\maketitle
\IEEEdisplaynontitleabstractindextext
\IEEEpeerreviewmaketitle
\IEEEraisesectionheading{\section{Introduction}\label{sec:INTRODUCTION}}
Data are represented as graphs in a wide range of applications, such as Computer Vision (e.g., images) and Graphics (e.g., 3D meshes), network analysis (e.g., social networks), and bio-informatics (e.g., molecules). \emph{Spectral graph processing} represents the input signal on a graph in terms of the eigenvectors of a graph operator (e.g., the graph Laplacian, a kernel matrix) in order to define its Fourier transform and convolution with another signal. According to this simple approach, whose discrete counterpart is based mainly on numerical linear algebra, spectral graph processing has been successfully applied to the characterisation of geometric and topological properties of graphs and to dimensionality reduction~\cite{BERARD1984}, through the projection of the input data on low-dimensional subspaces generated by a small set of Laplacian~\cite{BELKIN2003} or kernel~\cite{GHOSH2018} eigenfunctions.
Signal processing on graphs~\cite{SANDRYHAILA2013,SANDRYHAILA2014,CHEN2015} also supports smooth  signal interpolation~\cite{HEIMOWITZ2018,MAHADEVAN2005} and the definition of diffusion wavelets~\cite{LAFON2006,BELKIN2003,SINGER2006,HAMMOND2010}. Main applications of spectral processing are graph embedding through frequency analysis~\cite{BAHONAR2019}, the definition of uncertainty principles~\cite{PERRAUDIND018} and random walks~\cite{SINHA2013}, data representation~\cite{ZHU2003} and classification~\cite{NG2001}. Furthermore, the modelling and training of convolutional neural networks~\cite{KIPPEL2015}, and the design of fast localised convolutional filters in high-dimensional irregular domains (e.g., graphs)~\cite{DEFFERARD2016} have been recently addressed in the spectral domain. Finally, we recall the generalisation of spectral graph theory to 3D data through geometric deep learning~\cite{BRONSTEIN-PAMI2011} on non-Euclidean domains.

In this context, our overall goal is the definition of novel Fourier-based and rational graph filters for graph processing. Firstly (Sect.~\ref{sec:SPECTRAL-FOURIER-OPERATOR}), we show that the \emph{convolution} of signals on a non-Euclidean space is uniquely determined by the linearity and commutativity of the translation operator with respect to convolution. Indeed, the definition of the convolution operator, which is commonly used in the context of spectral graph processing, is unique. 
Through spectral filtering, we define \emph{spectral operators} that map 1D filter functions to signals defined on an arbitrary domain. The spectral operator induces a special class of \emph{Fourier-based spectral operators}, which generalise the notion of Fourier transform to non-Euclidean domains. In particular, we show that the main properties of the 1D Fourier transform, such as dilation, translation, scaling, derivation, localisation, and Parseval's equality, still apply to a signal defined on an arbitrary domain. Combining the spectral operator with convolution, we introduce the \emph{filtered convolution operator}, which is used to show the link between spectral operators, wavelets, and filtered convolution with integral operators induced by spectral kernels (Sect.~\ref{sec:SPECTRAL-WAVELETS}). The filtered convolution operator reduces to well-known Laplacian spectral operators (e.g., harmonic, bi-harmonic, diffusion, wave operators) for specific filters.

To efficiently compute spectral operators, it is necessary to apply fast polynomial approximations that are oblivious of the evaluation of the Laplacian spectrum, which is computationally expensive and numerically unstable. Main examples include polynomials~\cite{HAMMOND2010} (e.g., Chebyshev polynomials) and algebraic~\cite{WILSON2005} filters, which are evaluated through recursive relations and without diagonalising the Laplacian matrix associated with the input data.  To overcome the time-consuming computation of the filter coefficients in terms of the Chebyshev polynomials through the evaluation of integrals, we approximate the \mbox{$\mathcal{L}^{2}(\mathbb{R})$} scalar product with a discrete scalar product defined in terms of the Chebyshev nodes. In this way, we guarantee a high approximation accuracy and a linear computational cost for the evaluation of the Chebyshev coefficients of the input filter. 
Generalising these results, we propose a novel class of Laplacian spectral wavelets induced by \emph{rational polynomial filters}, whose evaluation is recursively expressed in terms of Chebyshev rational polynomials and requires the solution of a set of Laplace equations (Sect.~\ref{sec:POL-RATIONAL-FILTERS}). Rational polynomials are then applied to approximate arbitrary filters within a given tolerance, thus providing a more accurate and numerically stable alternative with respect to polynomials. In fact, rational polynomials are a reacher class of functions with respect to polynomials, improve the approximation accuracy of polynomials, are more stable with respect to oscillations, as the errors in the numerator and denominator compensate each others~\cite{GOLUB1989,LITVINOV1993}. Furthermore, rational polynomials have been computed analytically for filters (e.g., sin/cos, exponential, logarithm) commonly used in spectral graph processing. 

The definition of the spectral, filtered convolution, and Fourier-based wavelets through the filtering of the Laplacian spectrum faces a high computational cost for the evaluation of the spectrum in case of large graphs and numerical instabilities, which are associated with multiple or close eigenvalues for spectral graph processing. Even though multiple or close eigenvalues are quite common in real applications, the evaluation of the characteristic polynomial in these situations has deserved a little attention in spectral graph theory. For instance, close eigenvalues are associated with symmetries or perturbations of the input graph, or with a low accuracy of the eigensolver with respect to the spectral gap among eigenvalues.
To address the aforementioned numerical instabilities associated with the evaluation of the spectrum (Sect.~\ref{sec:NUMERICAL-METHODS}), we discuss the definition of the \emph{pseudo-spectrum} with respect to a given threshold and introduce the approximation of the characteristic polynomial through \emph{spectral densities}, which are efficiently computed through the evaluation of the trace of the Chebyshev polynomials of the input matrix. In particular, spectral densities allow us to apply the Caley-Hamilton theorem for the reduction of the degree of polynomial filters and to extract properties of the underlying graph. 

Analogously to the continuous case, we introduce a \emph{discrete spectrum-free approach} for the efficient evaluation of (discrete) spectral Fourier-based, and wavelet operators, which requires the solution of a small set of sparse, symmetric, and well-conditioned linear systems and is oblivious of the evaluation of the Laplacian or kernel spectrum. To further investigate these aspects, we evaluate the \emph{numerical stability} of the linear systems associated with the spectrum-free approximation, which confirms that the coefficient matrices involved in the computation are well-conditioned.
In this setting (Sect.~\ref{sec:DISCUSSION}), we show the generality of the proposed approach with respect to the input data (e.g., graphs, 3D shapes, etc) and to different applications, such as \emph{signal reconstruction} and \emph{smoothing}, and \emph{shape correspondence}. We also discuss the higher computational stability and accuracy of the rational approximation of spectral wavelets with respect to spectral approximations (e.g., low pass filters). 
Finally, rational filters, and in particular rational Chebyshev polynomials, are particularly useful to enlarge the class of learning networks, as a generalisation of networks based on polynomial filters (e.g., PolyNet~\cite{ZHANG2017}, ChebNet~\cite{KIPF2016}, CayleyNet~\cite{LEVIER2019}) in order to improve the discriminative capabilities of networks for 3D geometric deep learning.

\section{Spectral and Fourier-based operators\label{sec:SPECTRAL-FOURIER-OPERATOR}}
For the definition of spectral operators, we introduce the Laplace-Beltrami operator and its spectrum, which provide a generalisation of the Fourier basis to non-Euclidean domains. Let~$\mathcal{M}$ be an input domain (e.g., a manifold, a graph) and \mbox{$\mathcal{F}(\mathcal{M})$} the space of signals defined on~$\mathcal{M}$ (e.g., the space \mbox{$\mathcal{L}^{2}(\mathcal{M})$} of square integrable functions or the space \mbox{$\mathcal{C}^{0}(\mathcal{M})$} of continuous functions) equipped with the inner product \mbox{$\langle f,g\rangle_{2}:=\int_{\mathcal{M}}fg\textrm{d}\mu$} and the corresponding norm \mbox{$\|\cdot\|_{2}$}. The \emph{Laplace-Beltrami operator}~$\Delta$ is self-adjoint, positive semi-definite, and admits the \emph{Laplacian orthonormal eigensystem} \mbox{$(\lambda_{n},\phi_{n})_{n=0}^{+\infty}$}, \mbox{$\Delta\phi_{n}=\lambda_{n}\phi_{n}$}, \mbox{$\lambda_{0}=0$}, \mbox{$\lambda_{n}\leq\lambda_{n+1}$}, in \mbox{$\mathcal{L}^{2}(\mathcal{M})$}.

We show that the convolution of signals defined on a non-Euclidean space is uniquely determined by the linearity and commutativity of the translation operator with respect to convolution (Sect.~\ref{sec:CONVOLUTION-OPERATOR}). Then, we define a spectral operator that extend 1D filter functions to signals through the filtering of the spectrum and generalises well-known Laplacian spectral operators, such as the harmonic, bi-harmonic, and diffusion operators (Sect.~\ref{sec:SPECTRAL-OPERATOR}). Considering the continuous Fourier transform of 1D filters, the spectral operator induces a special class of Fourier-based spectral operators that extend the notion of Fourier transform to non-Euclidean domains (Sect.~\ref{sec:GENERALISED-SPECTRAL-OP}). Finally, we show that the main properties of the 1D Fourier transform, such as dilation, translation, scaling, Fourier transform, derivation, localisation, and Parseval's equality, still apply to the Fourier-based operator.

\subsection{Convolution operator\label{sec:CONVOLUTION-OPERATOR}}
Applying the linearity and commutativity with respect to convolution, we show that the spectral representation of the convolution operator, which is commonly used in the context of spectral graph processing, is unique. According to the continuous definition of the \emph{convolution operator}
\begin{equation}\label{eq:CONVOLUTION-OP}
(f\star g)(\mathbf{p})
:=\langle f,\mathcal{T}_{\mathbf{p}}g\rangle_{2},
\end{equation}
in terms of the~$\mathcal{L}^{2}(\mathcal{M})$ scalar product and of the translation operator~$\mathcal{T}_{\mathbf{p}}$, the convolution operator is uniquely defined by the translation operator. Indeed, it is enough to derive the spectral representation of the translation operator from its properties (i.e., linearity, commutativity). Noting that~$\mathcal{T}_{\mathbf{p}}$ is commutative with respect to convolution (i.e., \mbox{$\mathcal{T}_{\mathbf{p}}f\star g=f\star\mathcal{T}_{\mathbf{p}}g$}), and applying Eq. (\ref{eq:CONVOLUTION-OP}), we get that
\begin{equation}\label{eq:WEAK-CONVOLUTION}
\langle\mathcal{T}_{\mathbf{p}}f,\mathcal{T}_{\mathbf{q}}g\rangle_{2}
=\langle f,\mathcal{T}_{\mathbf{q}}\mathcal{T}_{\mathbf{p}}g\rangle_{2},\qquad
\forall\mathbf{p},\mathbf{q}\in\mathcal{M}.
\end{equation}
Considering the spectral representations of the functions \mbox{$f=\sum_{n=0}^{+\infty}\langle f,\phi_{n}\rangle_{2}\phi_{n}$}, and \mbox{$\mathcal{T}_{\mathbf{p}}f=\sum_{n=0}^{+\infty}\langle\mathcal{T}_{\mathbf{p}}f,\phi_{n}\rangle_{2}\phi_{n}$}, and \mbox{$\mathcal{T}_{\mathbf{q}}g=\sum_{n=0}^{+\infty}\langle\mathcal{T}_{\mathbf{q}}g,\phi_{n}\rangle_{2}\phi_{n}$}, Eq. (\ref{eq:WEAK-CONVOLUTION}) is equivalent to \mbox{$\langle\mathcal{T}_{\mathbf{p}}f,\phi_{n}\rangle_{2}\langle\mathcal{T}_{\mathbf{q}}g,\phi_{n}\rangle_{2}
=\langle f,\phi_{n}\rangle_{2}\langle\mathcal{T}_{\mathbf{q}}\mathcal{T}_{\mathbf{p}}g,\phi_{n}\rangle_{2}$}. Choosing \mbox{$f=g=\phi_{m}$}, the previous relation reduces to
\begin{equation*}
\langle\mathcal{T}_{\mathbf{p}}\phi_{m},\phi_{n}\rangle_{2}\langle\mathcal{T}_{\mathbf{q}}\phi_{m},\phi_{n}\rangle_{2}
=\delta_{mn}\langle\mathcal{T}_{\mathbf{q}}\mathcal{T}_{\mathbf{p}}\phi_{m},\phi_{n}\rangle_{2}.
\end{equation*}
The resulting identity \mbox{$\langle\mathcal{T}_{\mathbf{p}}\phi_{m},\phi_{n}\rangle_{2}=0$}, \mbox{$m\neq n$}, implies that \mbox{$\mathcal{T}_{\mathbf{p}}\phi_{n}=\alpha_{n}(\mathbf{p})\phi_{n}$}. Applying this last relation to the spectral representation of~$f$, we get that
\begin{equation}\label{eq:GENERAL-TRANSLATION}
\mathcal{T}_{\mathbf{p}}f
=\sum_{n=0}^{+\infty}\langle f,\phi_{n}\rangle_{2}\mathcal{T}_{\mathbf{p}}\phi_{n}\\
=\sum_{n=0}^{+\infty}\alpha_{n}(\mathbf{p})\langle f,\phi_{n}\rangle_{2}\phi_{n},
\end{equation}
i.e., the \emph{general spectral representation} of the translation operator. Assuming the commutativity of the translation operator with respect to the~$\delta$-function (i.e., \mbox{$\mathcal{T}_{\mathbf{p}}\delta_{\mathbf{q}}=\mathcal{T}_{\mathbf{q}}\delta_{\mathbf{p}}$}, for any \mbox{$\mathbf{p},\mathbf{q}\in\mathcal{M}$}), we have that \mbox{$\alpha_{n}(\mathbf{p})\phi_{n}(\mathbf{q})=\alpha_{n}(\mathbf{q})\phi_{n}(\mathbf{p})$}, for any \mbox{$\mathbf{p},\mathbf{q}\in\mathcal{M}$}, i.e., \mbox{$\alpha_{n}=\phi_{n}$}. Indeed, the translation operator in Eq. (\ref{eq:GENERAL-TRANSLATION}) becomes
\begin{equation}\label{eq:SPECTRAL-TRANSLATION}
\mathcal{T}_{\mathbf{p}}f
=\sum_{n=0}^{+\infty}\langle f,\phi_{n}\rangle_{2}\phi_{n}(\mathbf{p})\phi_{n}\\
=\sum_{n=0}^{+\infty}\widehat{f}(n)\phi_{n}(\mathbf{p})\phi_{n},
\end{equation}
with \mbox{$\widehat{f}(n):=\langle f,\phi_{n}\rangle_{2}$}~$n$-th Fourier coefficient of~$f$. Applying the spectral representation (\ref{eq:SPECTRAL-TRANSLATION}) of the translation operator to Eq. (\ref{eq:CONVOLUTION-OP}), we obtain the identity
\begin{equation}\label{eq:CONVOLUTION-DERIVATION}
(f\star g)(\mathbf{p})
=\langle f,\mathcal{T}_{\mathbf{p}}g\rangle_{2}
=\sum_{n=0}^{+\infty}\widehat{f}(n)\widehat{g}(n)\phi_{n}(\mathbf{p}).
\end{equation}
%
%
\subsection{Spectral and filtered convolution operators\label{sec:SPECTRAL-OPERATOR}}
Firstly, we introduce the definition and main properties of the spectral operator (Sect.~\ref{sec:SPECTRAL-OPERATOR-DETAIL}), which is applied to interpret filtered operators as convolution operators (Sect.~\ref{sec:FILTERED-CONVOLUTION}) and to extend the Fourier transform of 1D functions to signals defined on non-Euclidean domains (Sect.~\ref{sec:GENERALISED-SPECTRAL-OP}). 
\begin{figure}[t]
\begin{tabular}{ccc}
(a)\includegraphics[height=85pt]{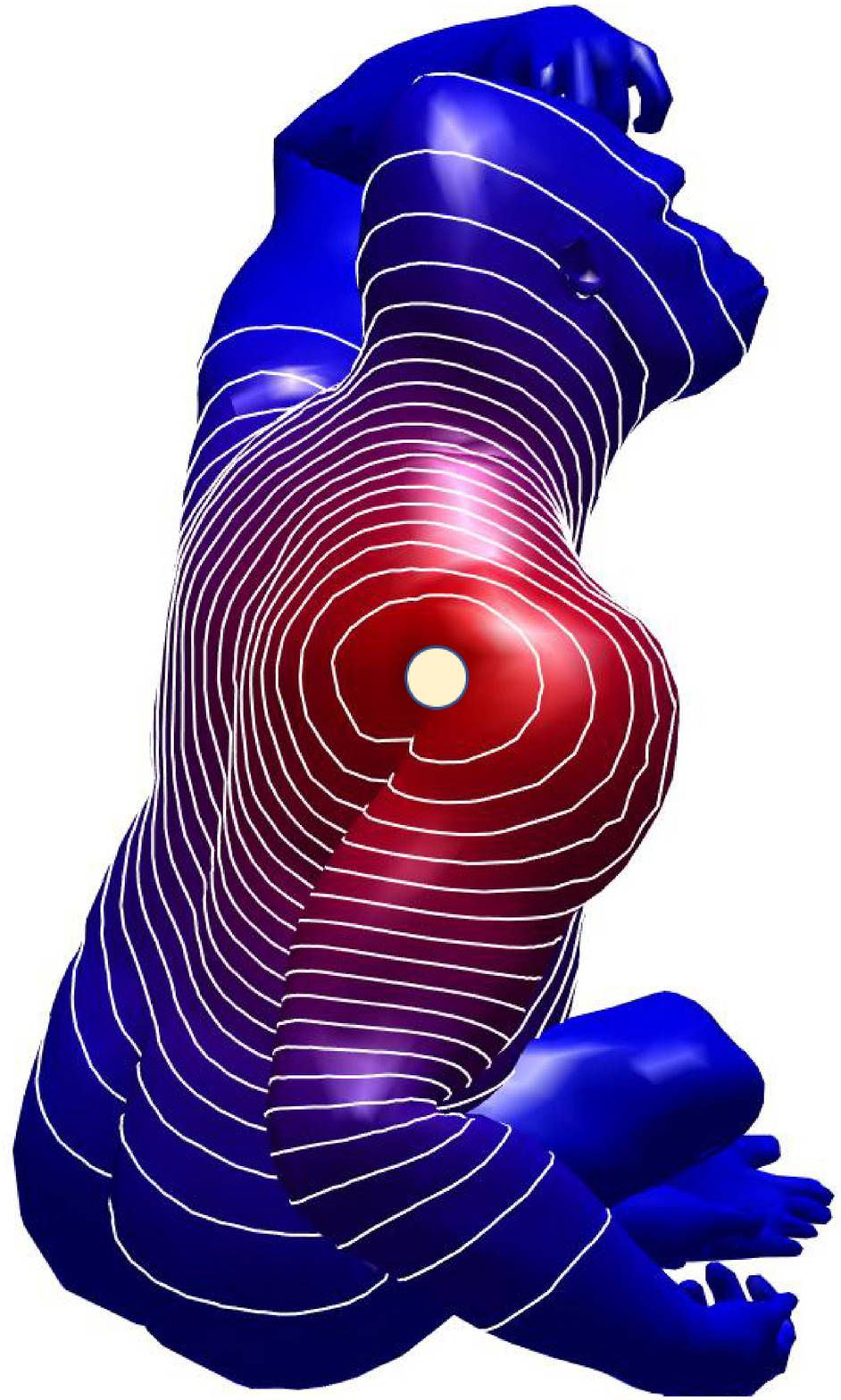}
&(b)\includegraphics[height=85pt]{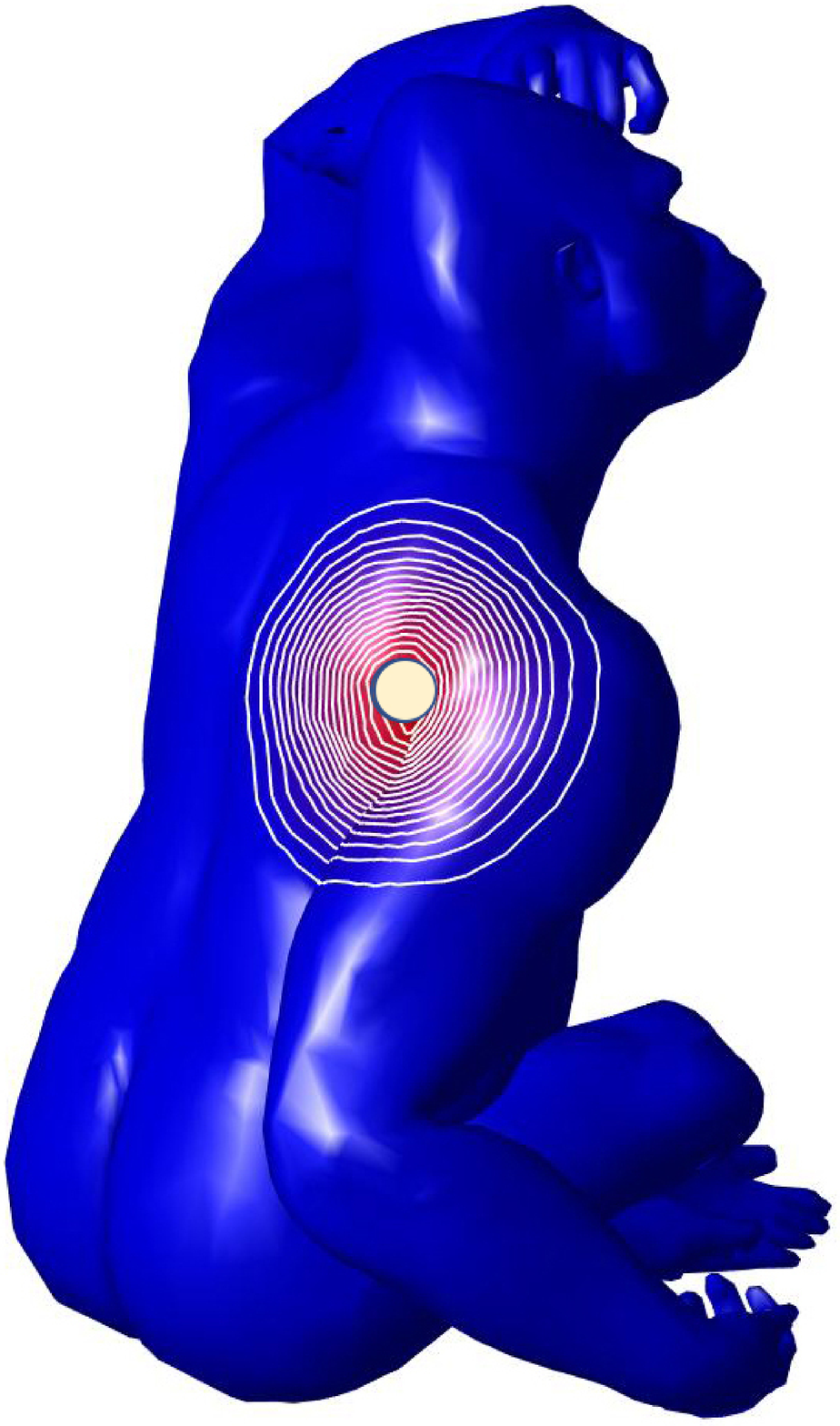}
&(c)\includegraphics[height=85pt]{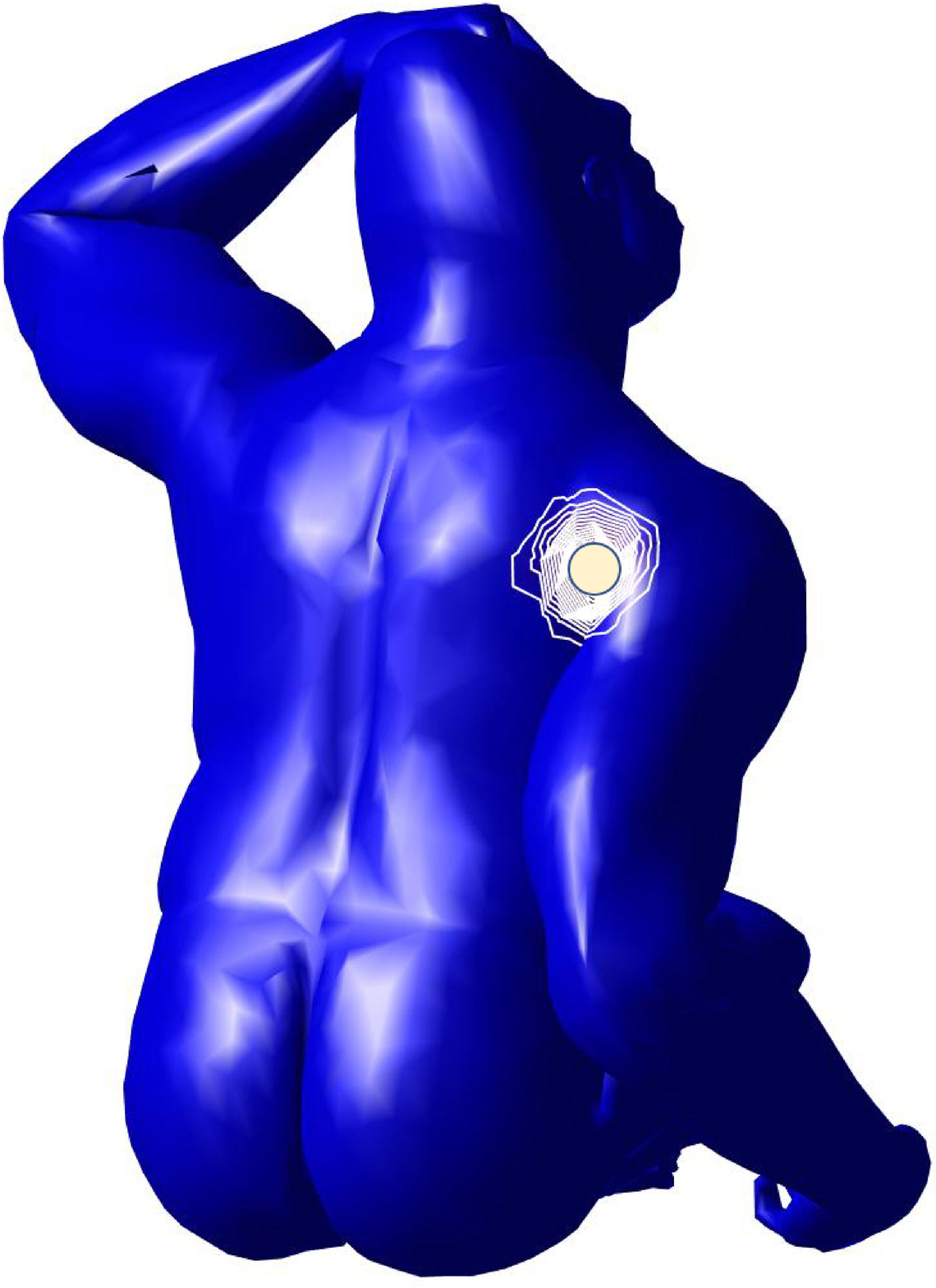}
\end{tabular}
\caption{Level-sets of spectral kernels induced by the filter \mbox{$\phi(s):=(1-s)^{-4}_{+}(4s+1)^{-1}$} with (a) \mbox{$s:=r$}, (b) \mbox{$s:=r^{2}$}, and (c) \mbox{$s:=r^{3}$} centred at a seed point (yellow dot). A larger value of~$s$ corresponds to a faster decay of~$\phi$, thus providing a smaller support of the corresponding kernel. (c) Reducing~$s$, the filter becomes constant and the corresponding kernels resembles the behaviour of the geodesic distance in a neighbour of the seed point.\label{fig:HEAD-DIF_PROBLEMS}} 
\end{figure}
\subsubsection{Spectral operator\label{sec:SPECTRAL-OPERATOR-DETAIL}}
Let \mbox{$\mathcal{H}:=\mathcal{L}^{2}(\mathbb{R})\cap\mathcal{C}^{0}(\mathbb{R})$} be the \emph{filter space} and \mbox{$\varphi:\mathbb{R}\rightarrow\mathbb{R}$} be a positive, continuous, and square integrable \emph{filter} in~$\mathcal{H}$. Then, we define the \emph{spectral operator}
\begin{equation}\label{eq:SPECTRAL-OP}
\Phi:\mathcal{H}\rightarrow\mathcal{F}(\mathcal{M}),\qquad
\varphi\mapsto\Phi_{\varphi}:=\sum_{n=0}^{+\infty}\varphi(\lambda_{n})\phi_{n},
\end{equation}
where~$\Phi_{\varphi}$ is the \emph{spectral function} induced by~$\varphi$. The spectral operator is \emph{linear} and \emph{continuous}, according to the following upper bound
\begin{equation}\label{eq:SPECTRAL-OP-WP}
\|\Phi_{\varphi}\|_{2}^{2}
=\sum_{n=0}^{+\infty}\vert\varphi(\lambda_{n})\vert^{2}
\leq\int_{0}^{+\infty}\vert\varphi(s)\vert^{2}ds
=\|\varphi\|_{2}^{2}.
\end{equation}
The continuity of the filter function allows us to evaluate its values \mbox{$(\varphi(\lambda_{n}))_{n=0}^{+\infty}$} at the eigenvalues and the~$\mathcal{L}^{2}(\mathbb{R})$-integrability of the filter ensures the well-posedness of the spectral operator. The properties of the spectral operator depends only on the behaviour of the filter in the spectral domain and on the Laplacian/kernel spectrum; increasing or decreasing the filter decay to zero encodes global or local properties of the input graph, respectively.
\begin{figure}[t]
\centering
\begin{tabular}{c|c|c}
\hline
\end{tabular}
\begin{tabular}{cc}
\includegraphics[height=60pt]{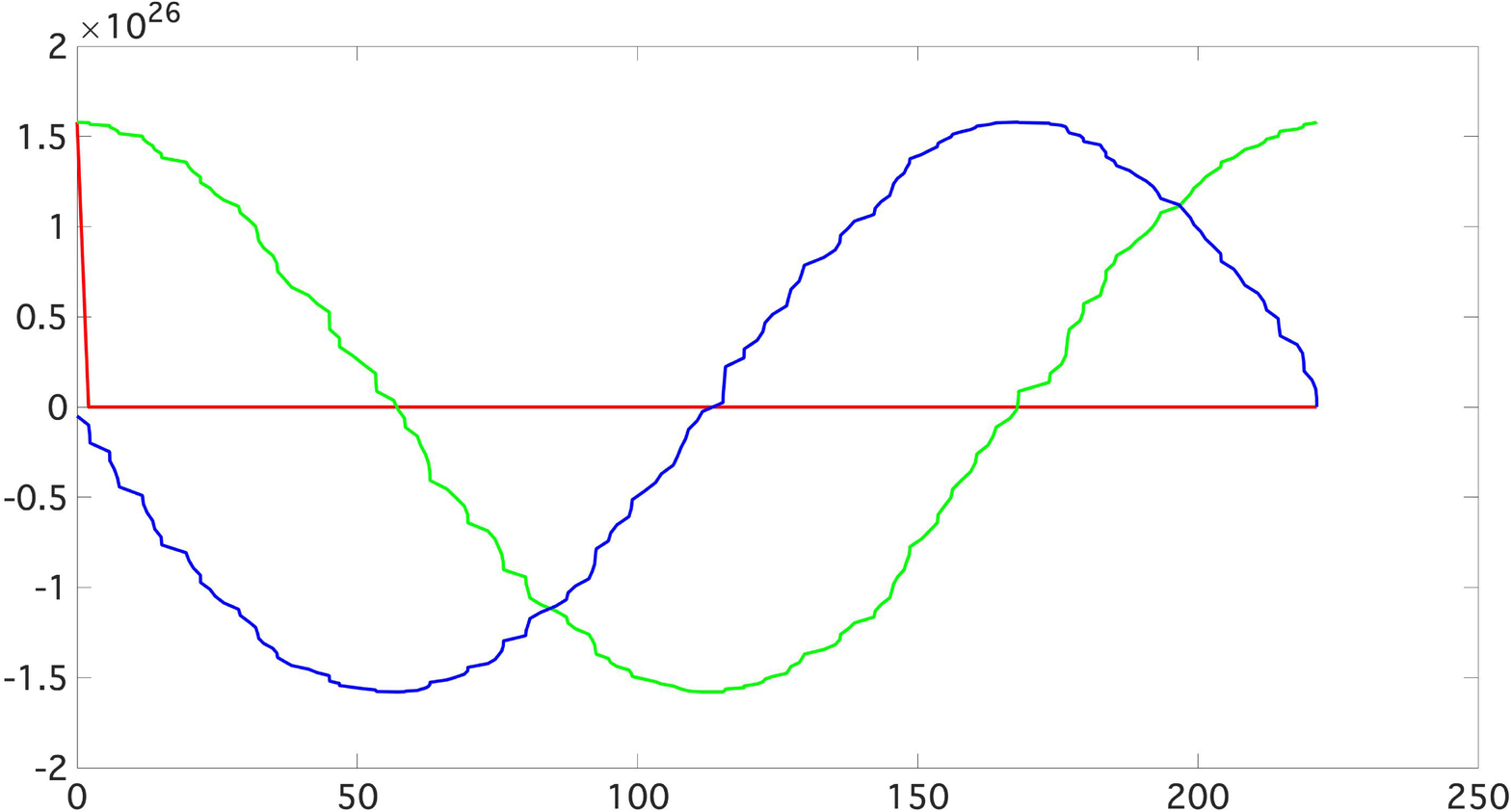}
&\includegraphics[height=60pt]{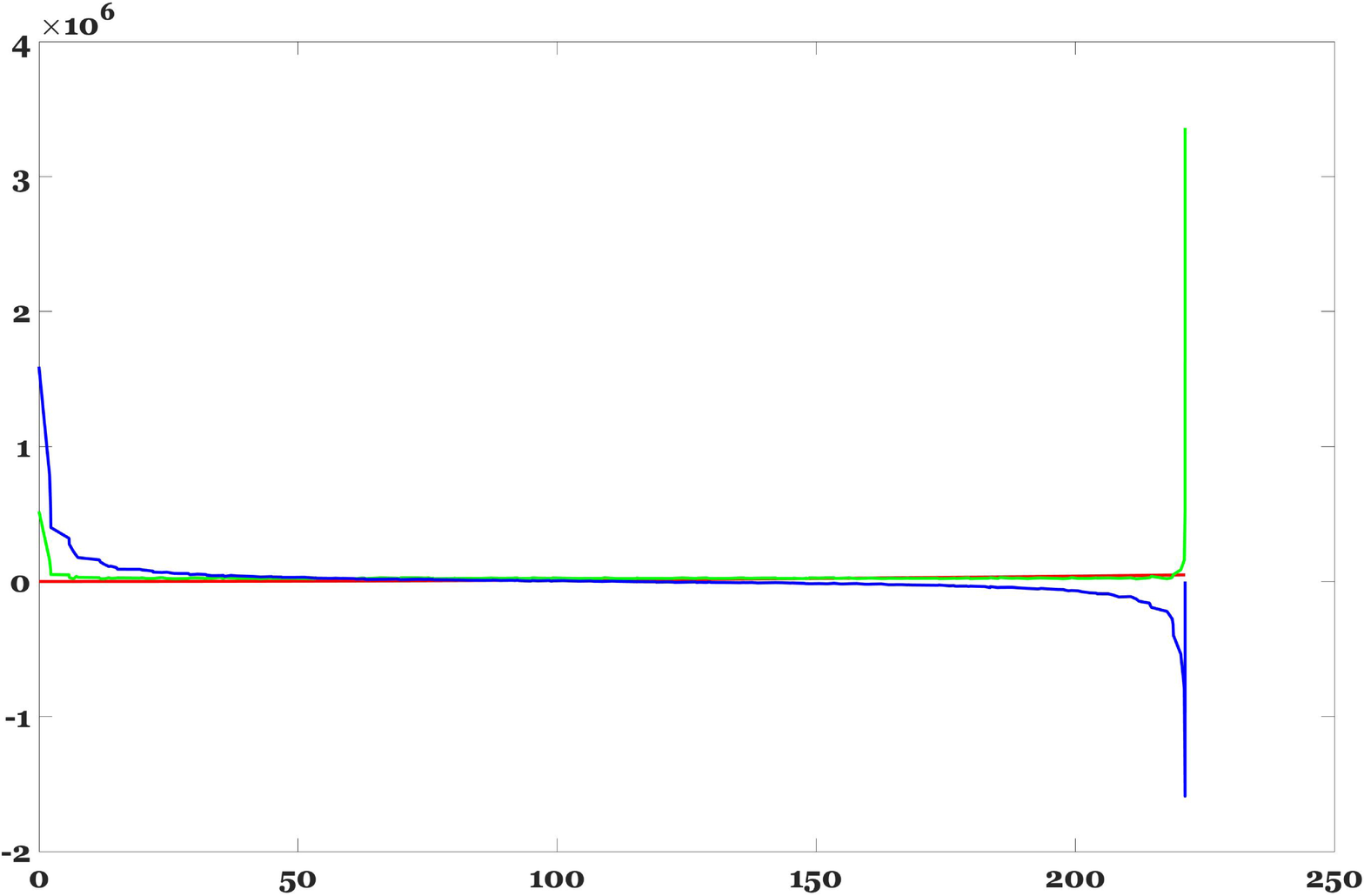}
\end{tabular}
\begin{tabular}{c|c|c}
$\Phi_{\varphi}$
&$\Phi_{\textrm{Real}(\widehat{\varphi})}$
&$\Phi_{\textrm{Im}(\widehat{\varphi})}$\\
\hline
\includegraphics[height=60pt]{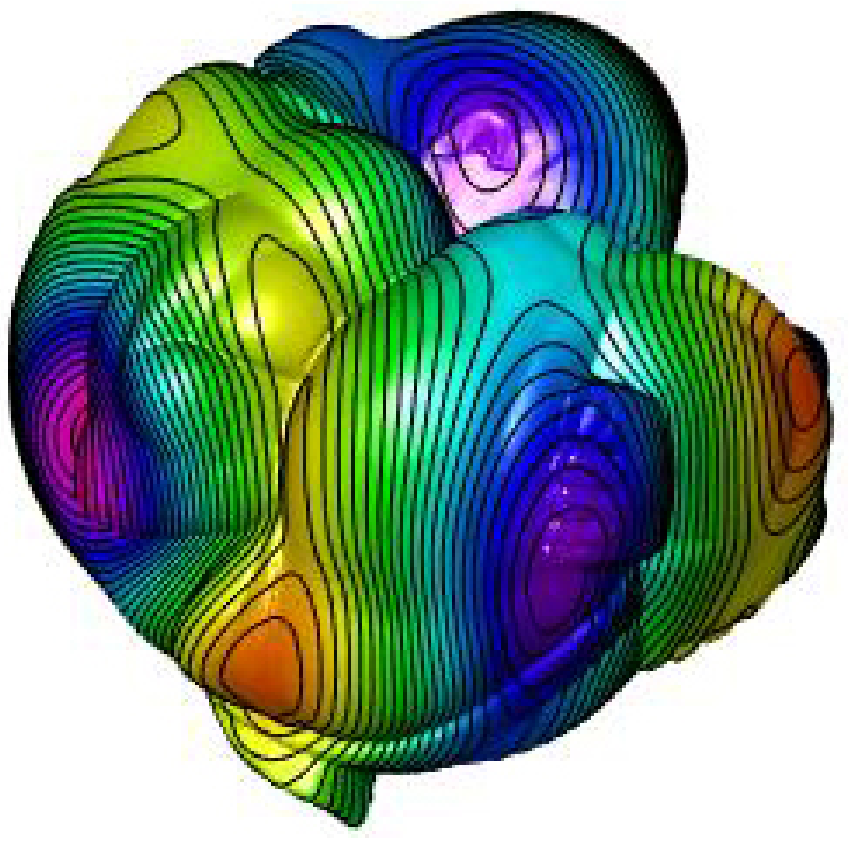}
&\includegraphics[height=60pt]{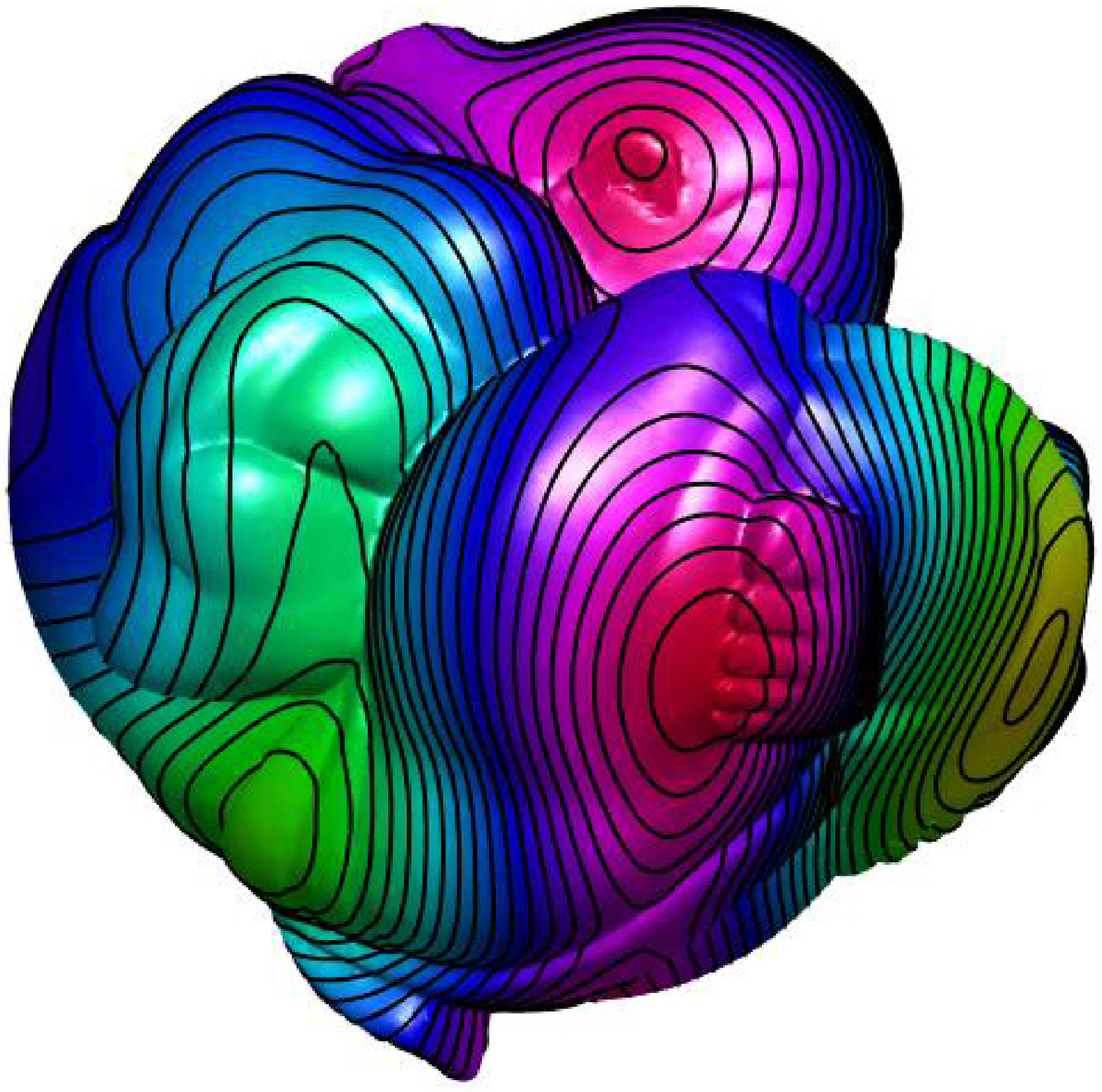}
&\includegraphics[height=60pt]{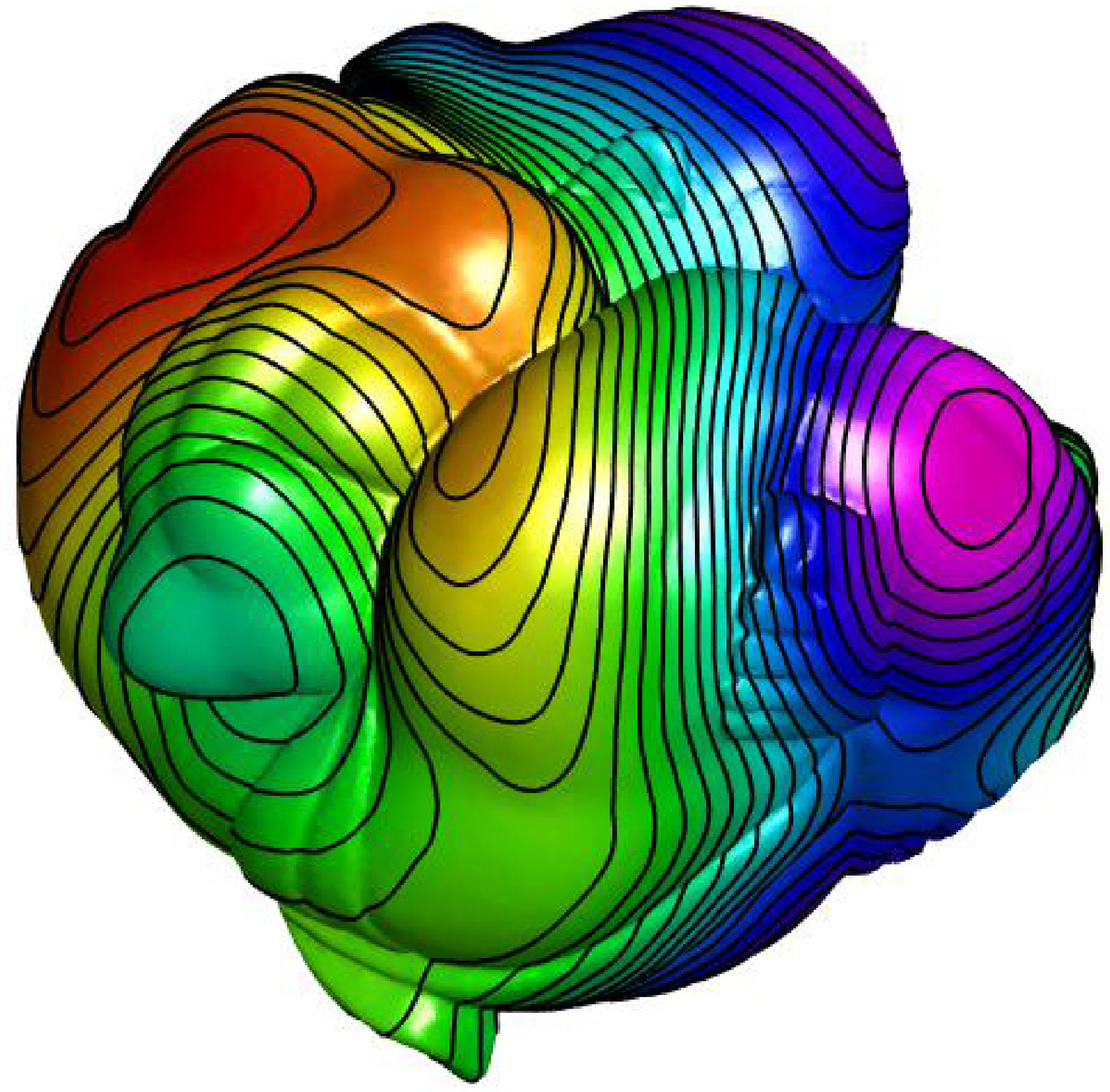}\\
\includegraphics[height=60pt]{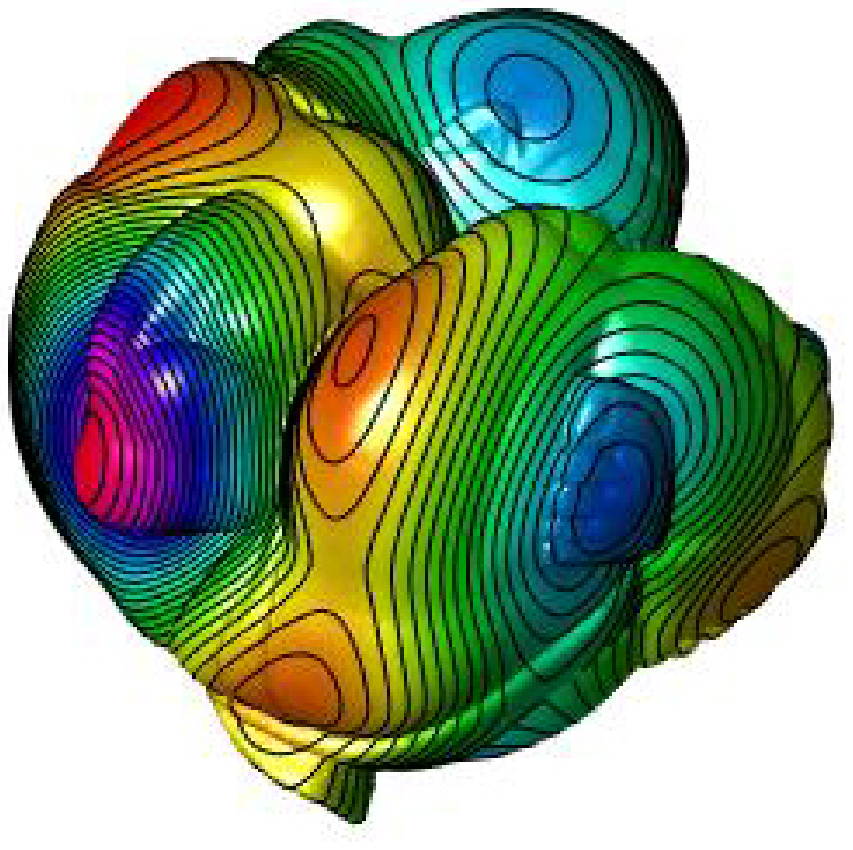}
&\includegraphics[height=60pt]{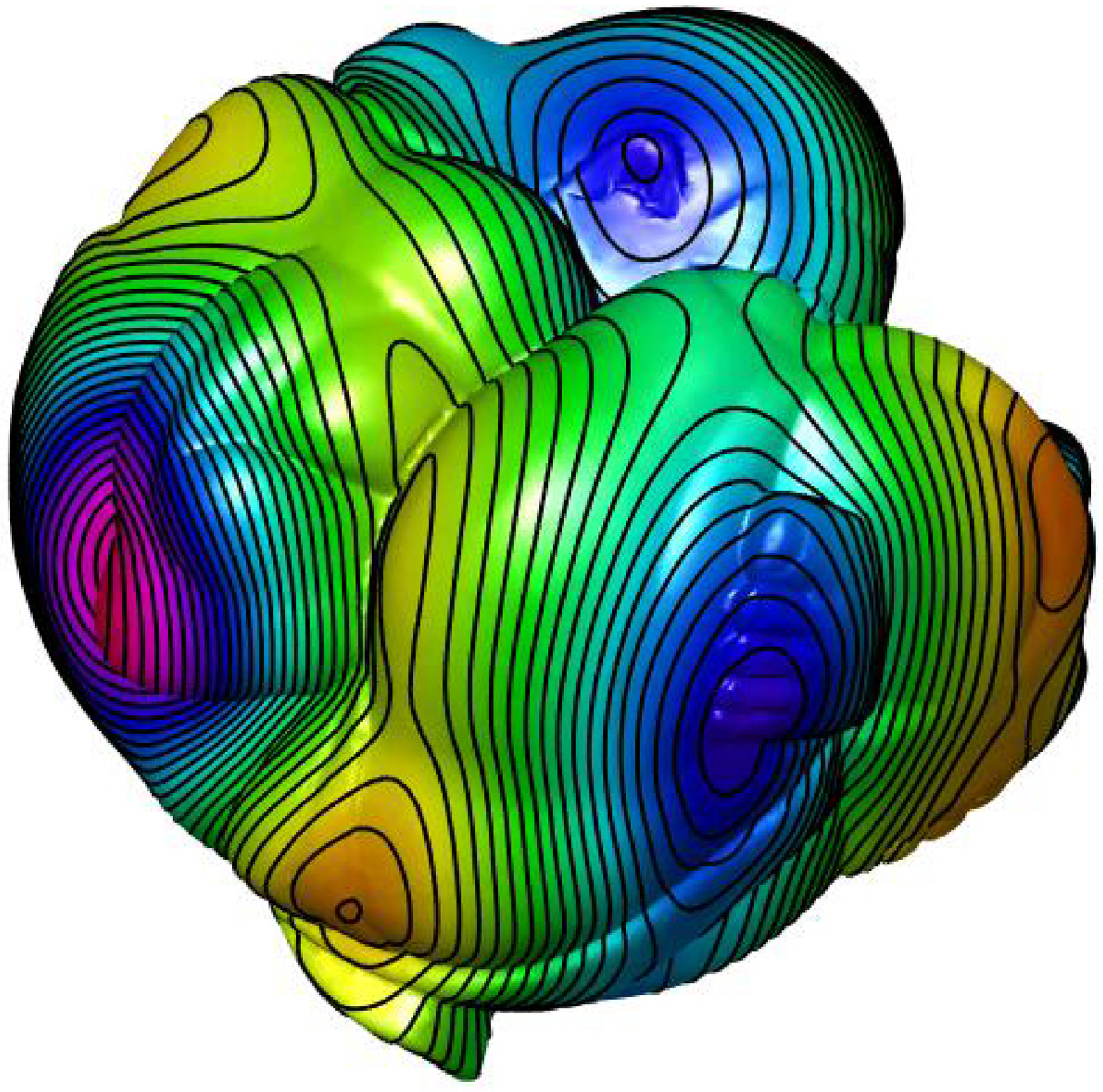}
&\includegraphics[height=60pt]{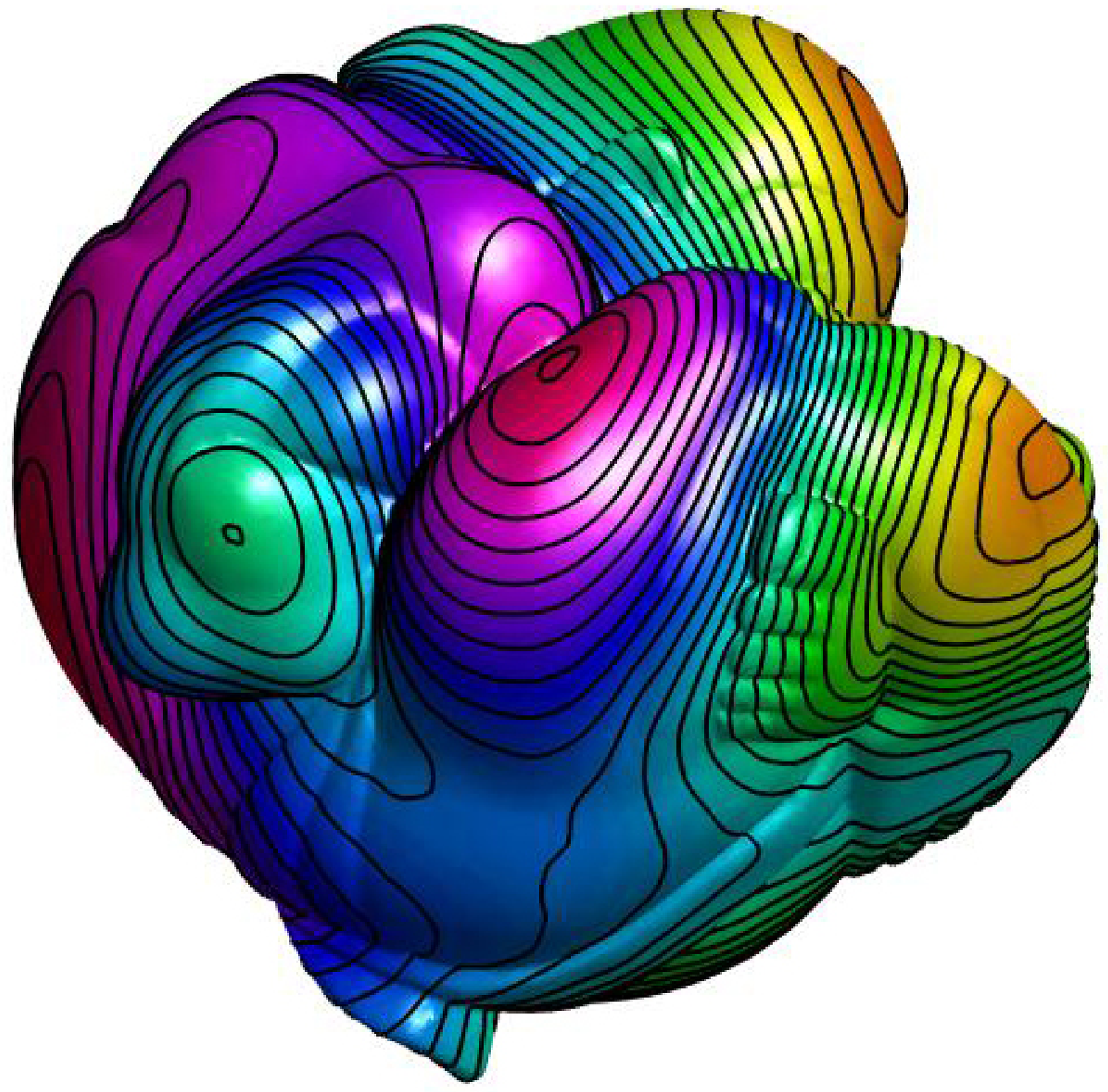}
\end{tabular}
\caption{Input filters~$\varphi$ (red), real (green) and imagery (blue) parts of its Fourier transform~$\widehat{\varphi}$ computed with the Fast Fourier Transform. Level-sets of the spectral operator induced by~$\varphi$, \mbox{$\textrm{Im}(\widehat{\varphi})$}, and \mbox{$\textrm{Real}(\widehat{\varphi})$}: (first row) \mbox{$\varphi(s):=s^{-2}$}, (second row) \mbox{$\varphi(s):=s^{2}$}.\label{fig:FOURIER-FILTER}}
\end{figure}

\textbf{Scalar product and norm of spectral operators}
The~$\mathcal{L}^{2}(\mathcal{M})$ scalar product between the spectral functions~$\Phi_{\varphi_{1}}$,~$\Phi_{\varphi_{2}}$ reduces to the~$\ell_{2}$ scalar product of the filtered eigenvalues and is bounded by the~$\mathcal{L}^{2}(\mathbb{R})$ scalar product of the corresponding filters~$\varphi_{1}$,~$\varphi_{2}$, i.e., 
\begin{equation*}
\begin{split}
&\langle\Phi_{\varphi_{1}},\Phi_{\varphi_{2}}\rangle_{2}
=\sum_{n=0}^{+\infty}\varphi_{1}(\lambda_{n})\varphi_{2}(\lambda_{n})=\langle(\varphi_{1}(\lambda_{n}))_{n}(\varphi_{2}(\lambda_{n}))_{n}\rangle_{2}\\
&\leq\int_{0}^{+\infty}\varphi_{1}(s)\varphi_{2}(s)ds
\leq\langle\varphi_{1},\varphi_{2}\rangle_{2},\quad (\varphi_{1},\varphi_{2}\geq 0).
\end{split}
\end{equation*}
The~$\mathcal{L}^{2}(\mathcal{M})$-distance of the spectral functions is bounded by~$\mathcal{L}^{2}(\mathbb{R})$-distance of the corresponding filter functions, i.e.,
\begin{equation}\label{eq:SPECTRAL-BOUND}
\|\Phi_{\varphi_{1}}-\Phi_{\varphi_{2}}\|_{2}
=\left\|\Phi_{\varphi_{1}-\varphi_{2}}\right\|_{2}
=_{\textrm{Eq. }(\ref{eq:SPECTRAL-OP-WP})}\|\varphi_{1}-\varphi_{2}\|_{2}.
\end{equation}
Indeed, the approximation of~$\Phi_{\varphi}$ reduces to the approximation of the 1D filter~$\varphi$; this remark will be applied to the spectrum-free computation of spectral and wavelet operators with rational polynomials (Sect.~\ref{sec:POL-RATIONAL-FILTERS}).
\begin{figure}[t]
\centering
\begin{tabular}{ccc}
\includegraphics[height=75pt]{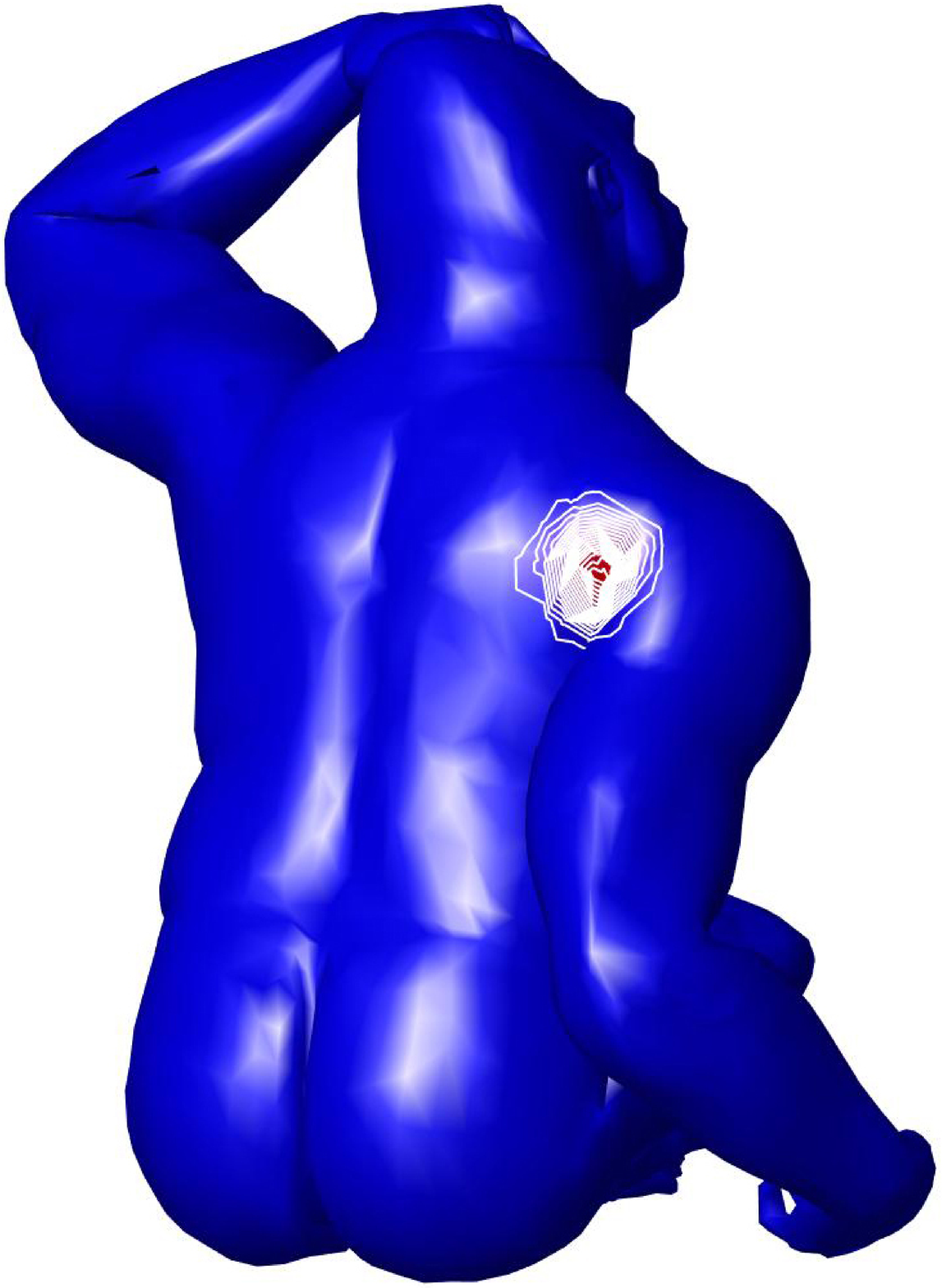}
&\includegraphics[height=75pt]{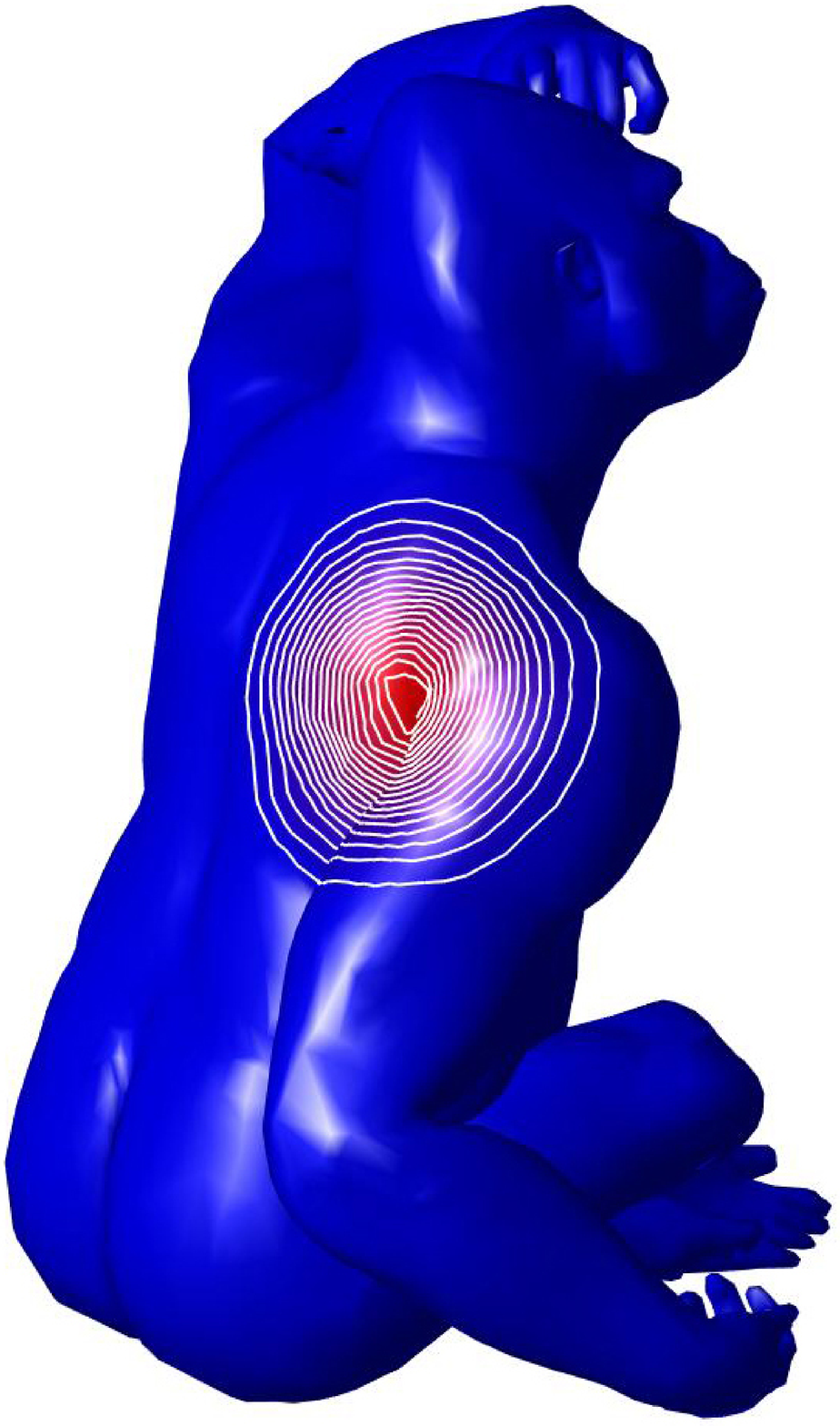}
&\includegraphics[height=75pt]{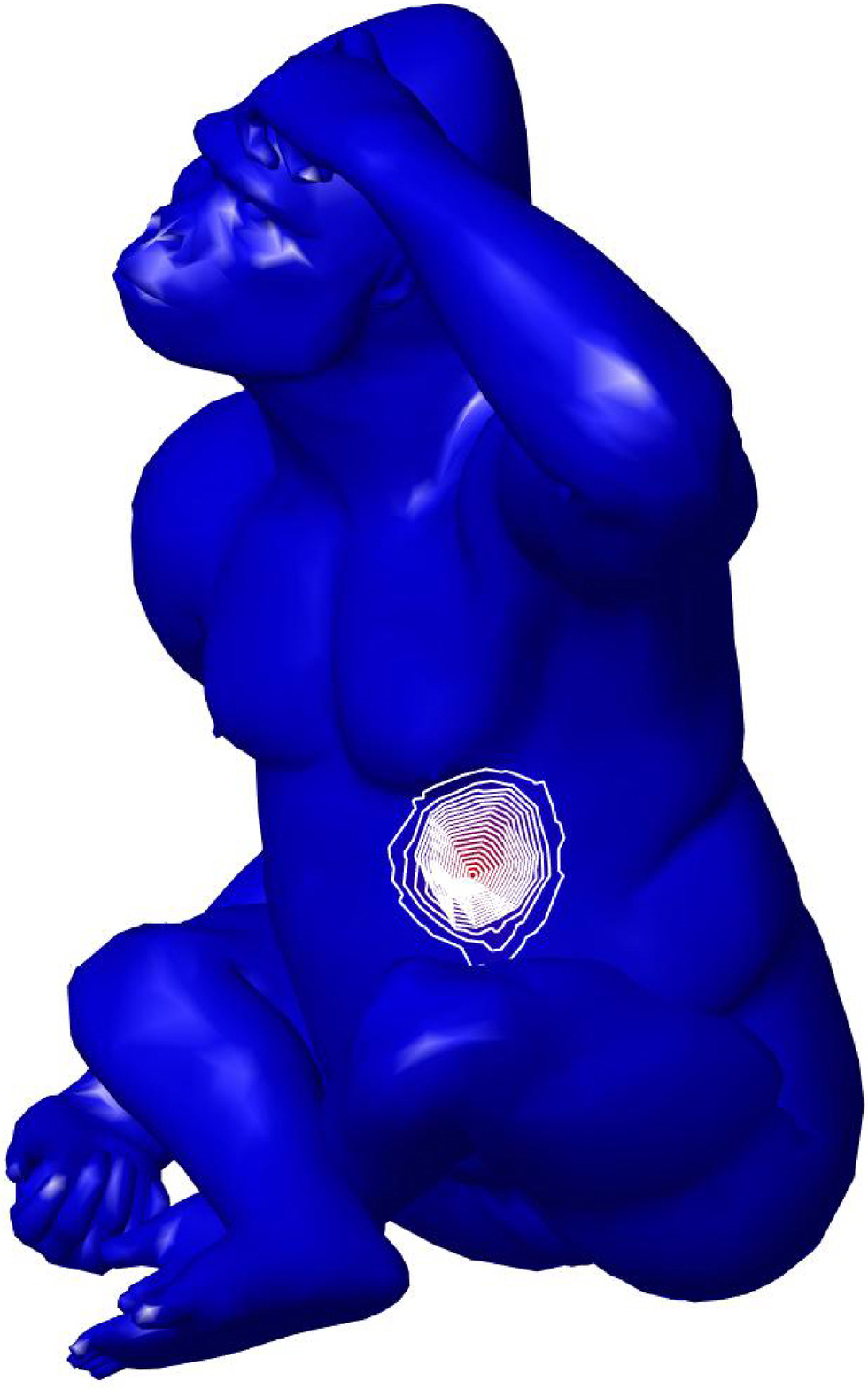}\\
(a)~$t=10^{-3}$ &(b)~$t=10^{-3}$ &(c)~$t=10^{-2}$\\
\includegraphics[height=75pt]{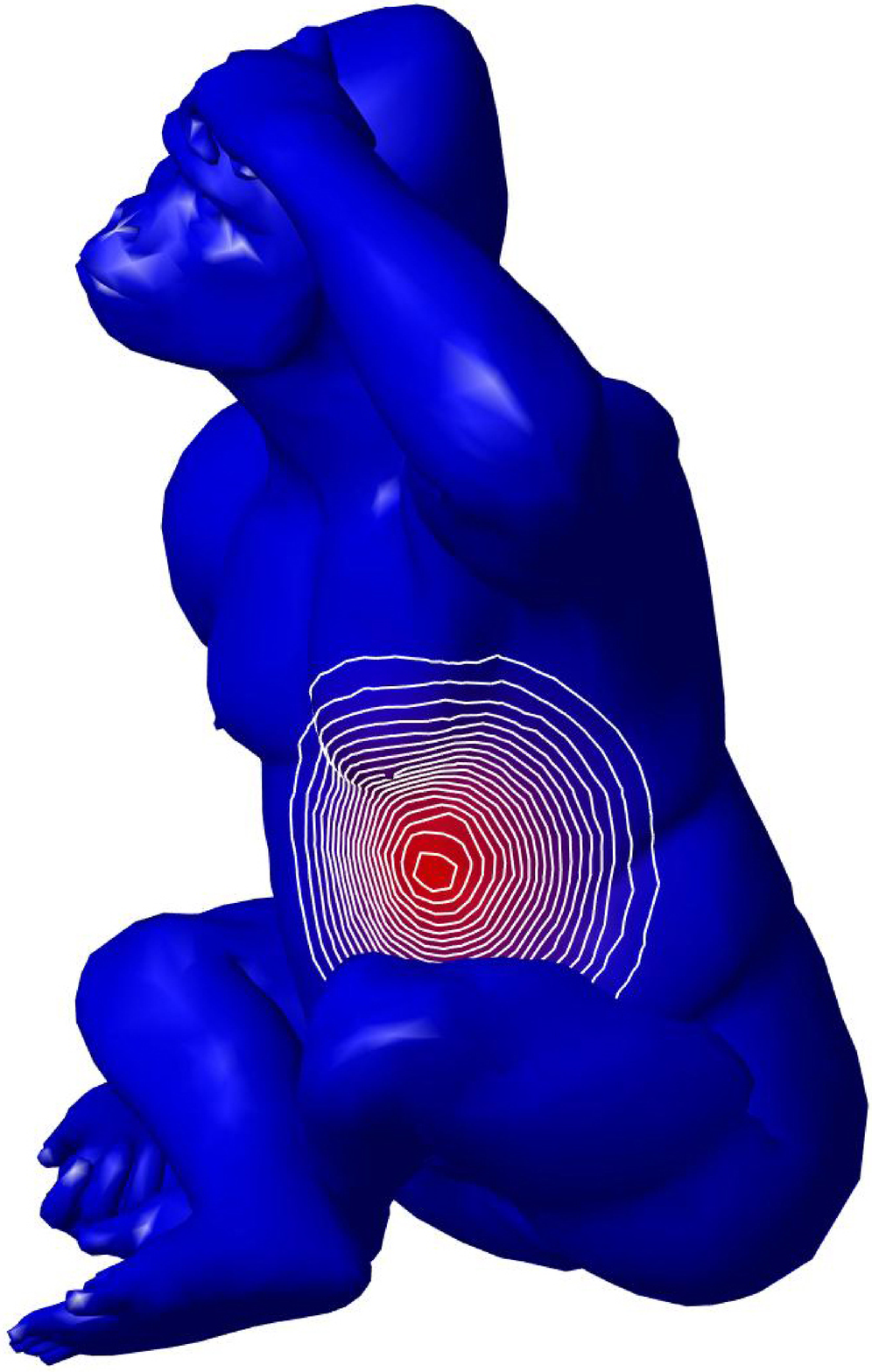}
&\includegraphics[height=75pt]{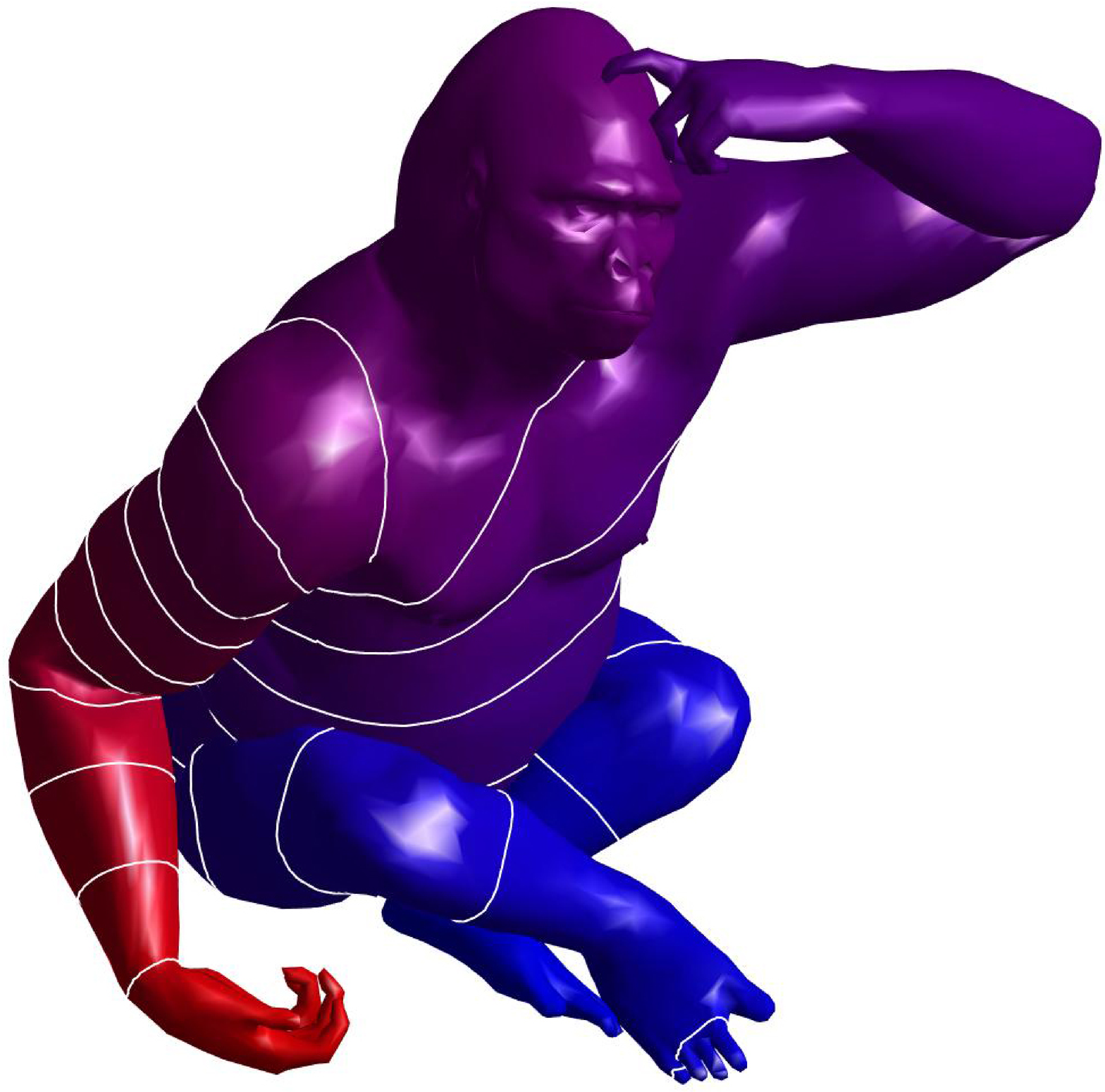}
&\includegraphics[height=75pt]{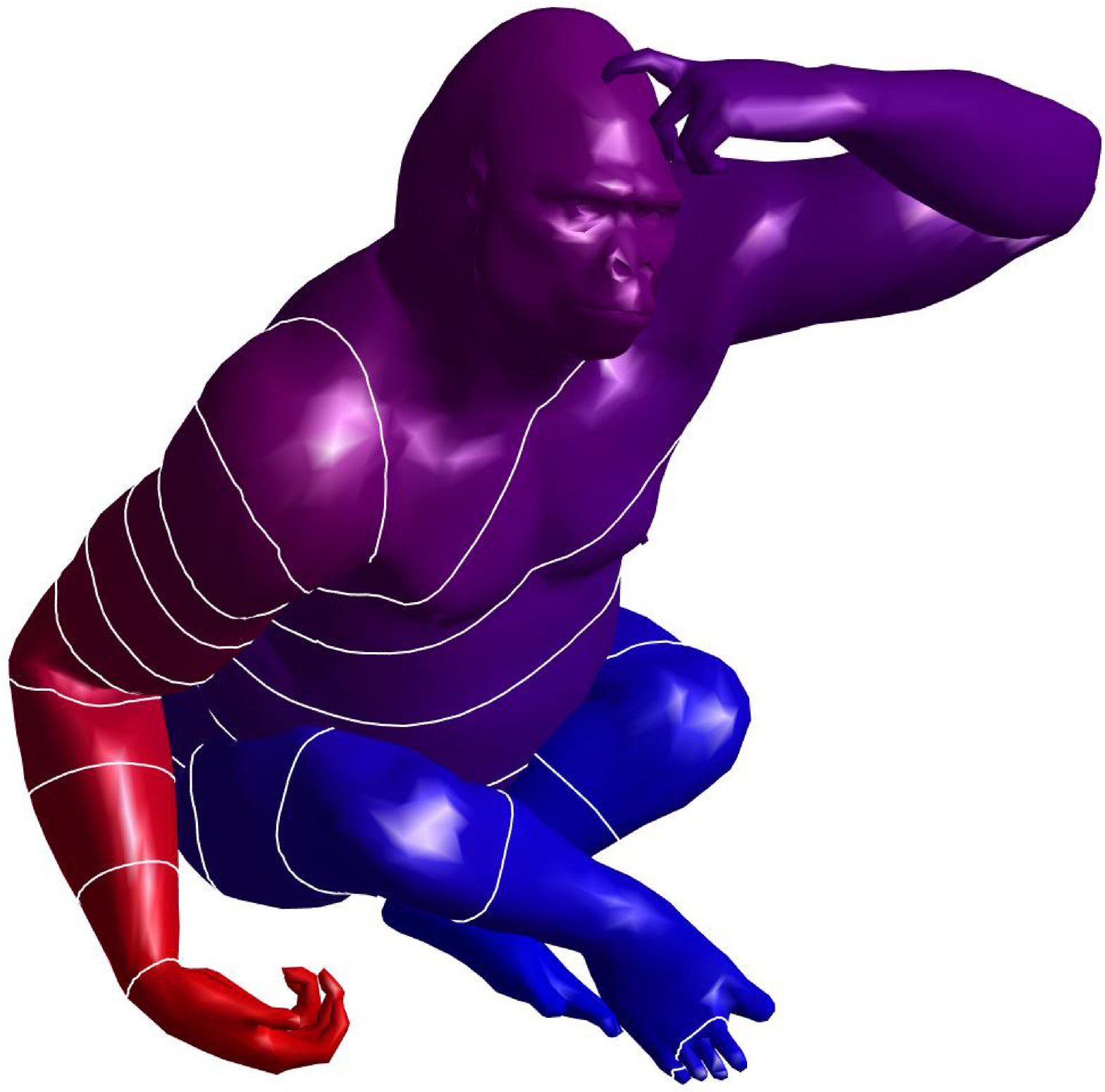}\\
(d)~$t=10^{-2}$ &(e)~$t=0.5$ &(f)~$t=1$
\end{tabular}
\caption{At small scales (\mbox{$t=10^{-3}$}, \mbox{$t=10^{-2}$}), the distribution of the level-sets of the diffusion kernel have a local and multi-scale behaviour at the same seed points (a,b), (c,d). (e,f) At large scales (\mbox{$t=10^{-2}$}, \mbox{$t=0.5$}, \mbox{$t=1$}), the diffusion kernels are no more centred at the seed points and have an analogous global behaviour (e,f). The kernel has been approximated with a rational polynomial of the exp filter of degree \mbox{$(r,r)$}, \mbox{$r:=7$} (c.f. Eq. (\ref{eq:CHEBYSHEV-BASIS})).\label{fig:MONKEY-DIFFUSION}}
\end{figure}
\subsubsection{Filtered convolution operator\label{sec:FILTERED-CONVOLUTION}}
We now show the link between spectral and filtered convolution operators with filtered Laplacian operators and integral operators induced by the spectral kernels. Introducing the \emph{filtered convolution operator} \mbox{$\Psi_{\varphi}:\mathcal{F}(\mathcal{M})\rightarrow\mathcal{F}(\mathcal{M})$}
\begin{equation}\label{eq:SPECTRAL-CONV}
\Psi_{\varphi}f
:=\Phi_{\varphi}\star f
=\sum_{n=0}^{+\infty}\varphi(\lambda_{n})\langle f,\phi_{n}\rangle_{2}\phi_{n}
=\Phi_{\varphi_{1}}
=\Psi_{\varphi_{1}}f,
\end{equation}
with \mbox{$\varphi_{1}:=\varphi\widehat{f}$}, the convolution between the spectral function~$\Phi_{\varphi}$ and a function~$f$ is induced by the pointwise product \mbox{$\varphi\widehat{f}$} between the filter~$\varphi$ and the Fourier transform of~$f$. In particular, the convolution of the spectral functions~$\Phi_{\varphi_{1}}$,~$\Phi_{\varphi_{2}}$ is induced by their pointwise product \mbox{$\varphi_{1}\varphi_{2}$}, i.e., \mbox{$\Phi_{\varphi_{1}}\star \Phi_{\varphi_{2}}=\Phi_{\varphi_{1}\varphi_{2}}$}. Assuming that the input filter admits the power-series representation \mbox{$\varphi(s):=\sum_{n=0}^{+\infty}\alpha_{n}s^{n}$}, we define the \emph{filtered Laplacian operator} \mbox{$\varphi(\Delta):=\sum_{n=0}^{+\infty}\alpha_{n}\Delta^{n}$}, which is equal to the filtered convolution operator \mbox{$\Psi_{\varphi}=\varphi(\Delta)$}; in fact, \mbox{$\varphi(\Delta)f=\sum_{n=0}^{+\infty}\varphi(\lambda_{n})\langle f,\phi_{n}\rangle_{2}\phi_{n}=_{\textrm{Eq. }(\ref{eq:SPECTRAL-CONV})}\Psi_{\varphi}f$}. Defining \mbox{$K_{\varphi}(\mathbf{p},\mathbf{q}):=
\Psi_{\varphi}\delta_{\mathbf{p}}
=\sum_{n=0}^{+\infty}\varphi(\lambda_{n})\phi_{n}(\mathbf{p})\phi_{n}(\mathbf{q}),$} as  \emph{spectral kernel} and noting that
\begin{equation}\label{eq:SPECTRAL-INT-CONV}
\Psi_{\varphi}f(\mathbf{p})=\langle K_{\varphi}(\mathbf{p},\cdot),f\rangle_{2}, 
\end{equation}
the filtered convolution operator is equal to the \emph{integral operator} induced by the spectral kernel, i.e., \mbox{$\Psi_{\varphi}f=\langle K_{\varphi},f\rangle_{2}$}. Fig.~\ref{fig:HEAD-DIF_PROBLEMS} shows different spectral kernels centred at a seed point on a 3D surface represented as a triangle mesh. Indeed, the input graph is arbitrary and the nodes have a 3D embedding, i.e., the coordinates of the mesh vertices. The spectral kernel well encodes the local geometry, as confirmed by the shape and distribution of the level-sets at all scales.

\textbf{Approximation of filtered convolution operators}
The norm of the filtered convolution operator \mbox{$\Psi_{\varphi}$} is bounded in terms of the~$\mathcal{L}^{2}(\mathbb{R})$ and~$\mathcal{L}^{\infty}(\mathbb{R})$ norm of the input filter function as \mbox{$\|\Psi_{\varphi}\|\leq\|\varphi\|_{k}$}, \mbox{$k=2,\infty$}. In fact, we have that
\begin{equation}\label{eq:NORM-SPECTRAL-OPERATOR}
\|\Psi_{\varphi}f\|^{2}
=\sum_{n=0}^{+\infty}\vert\varphi(\lambda_{n})\vert^{2}\vert\langle f,\phi_{n}\rangle_{2}\vert^{2}
\leq\|f\|_{2}^{2}\|\varphi\|_{2}^{2}.
\end{equation}
Choosing the signal \mbox{$f:=\sum_{n}\phi_{n}$}, we get \mbox{$\|\Psi_{\varphi}f\|_{2}=\|\varphi\|_{2}$}, i.e., \mbox{$\|\Psi_{\varphi}\|=\|\varphi\|_{2}$}. From the equality (\ref{eq:NORM-SPECTRAL-OPERATOR}), it follows that 
\begin{equation}\label{eq:GENERAL-BOUND}
\|\Psi_{\varphi}f\|^{2}
\leq\|\varphi\|_{\infty}^{2}\sum_{n=0}^{+\infty}\vert\langle f,\phi_{n}\rangle_{2}\vert^{2}
=\|f\|_{2}^{2}\|\varphi\|_{\infty}^{2}.
\end{equation}
Choosing the signal \mbox{$f:=\|\varphi\|_{\infty}\sum_{n}\phi_{n}$}, we get \mbox{$\|\Psi_{\varphi}f\|_{2}=\|\varphi\|_{\infty}$}, i.e., \mbox{$\|\Psi_{\varphi}\|=\|\varphi\|_{\infty}$}. Analogously to Eq. (\ref{eq:SPECTRAL-BOUND}), the~$\mathcal{L}_{k}~$distance, \mbox{$k=2,\infty$},
\begin{equation}\label{eq:CONVOLUTION-BOUND}
\|\Psi_{\varphi_{1}}-\Psi_{\varphi_{2}}\|
=\|\Psi_{\varphi_{1}-\varphi_{2}}\|
\leq_{\textrm{Eqs. }(\ref{eq:NORM-SPECTRAL-OPERATOR}),(\ref{eq:GENERAL-BOUND})}\|\varphi_{1}-\varphi_{2}\|_{k},
\end{equation}
between two filtered convolution operators is bounded by the \mbox{$\mathcal{L}^{k}(\mathbb{R})$} distance of the corresponding filters  (Fig.~\ref{fig:FOURIER-FILTER}).
\begin{figure}[t]
\centering
\begin{tabular}{ccc}
\includegraphics[height=80pt]{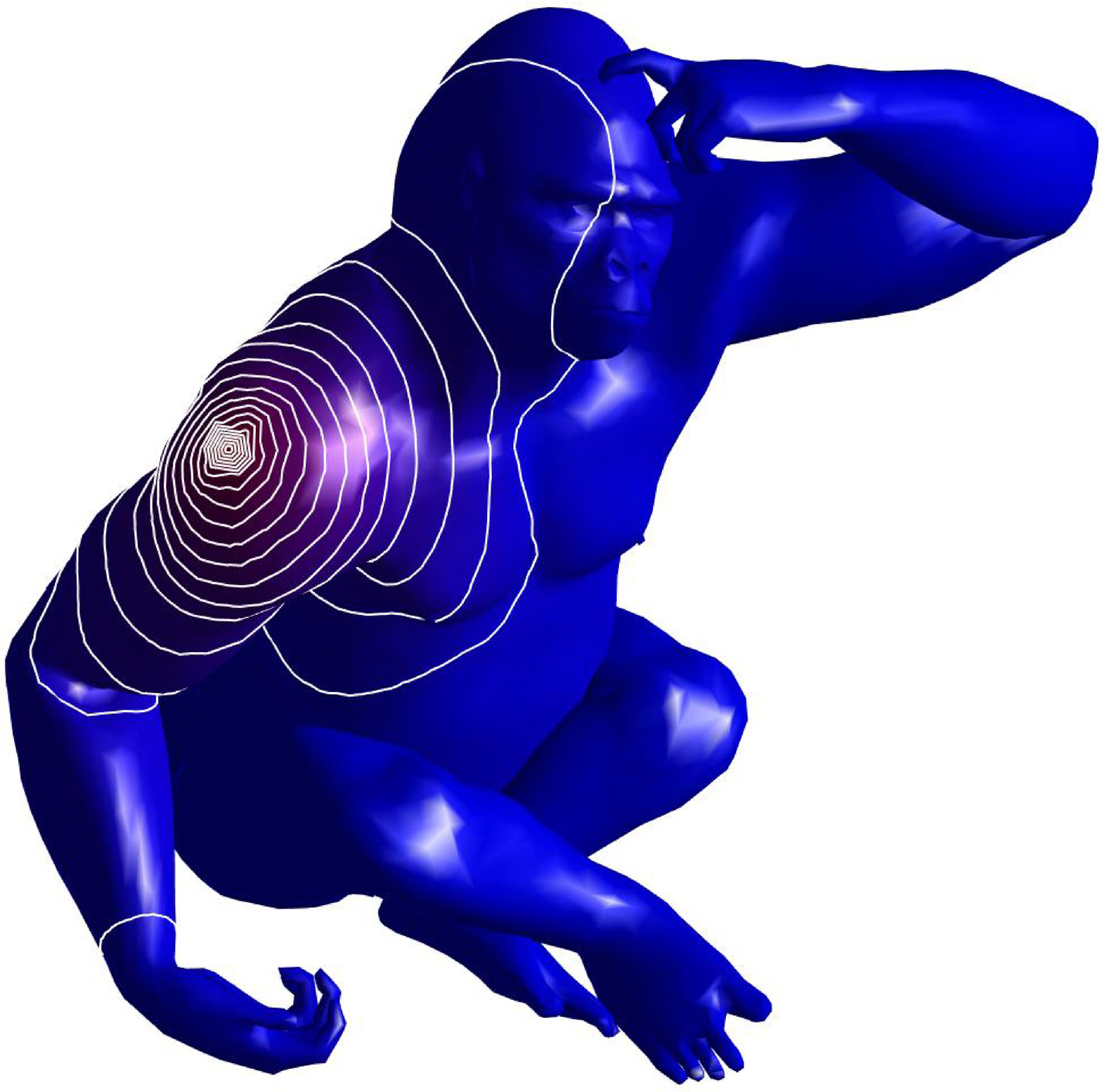}
&\includegraphics[height=80pt]{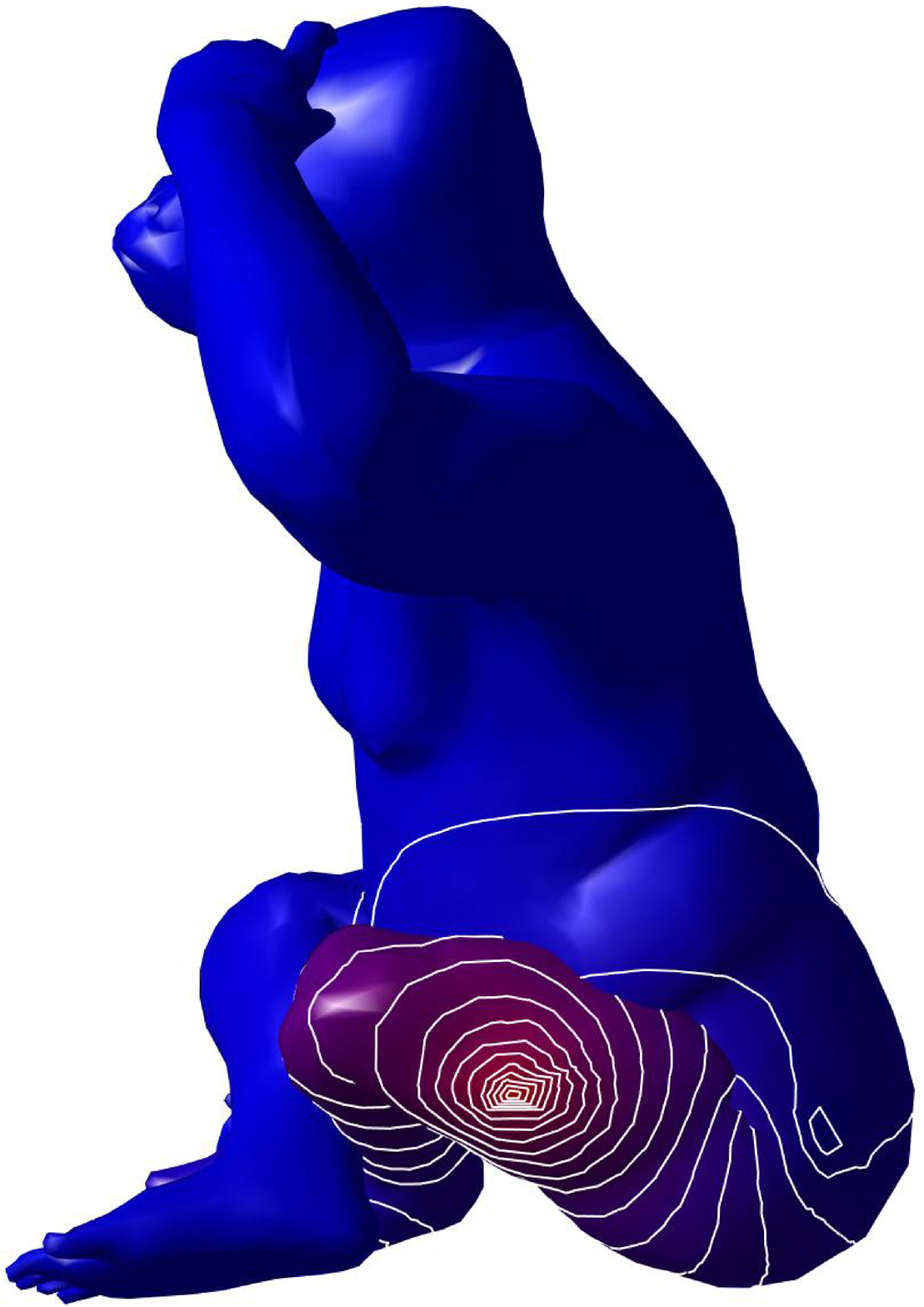}
&\includegraphics[height=80pt]{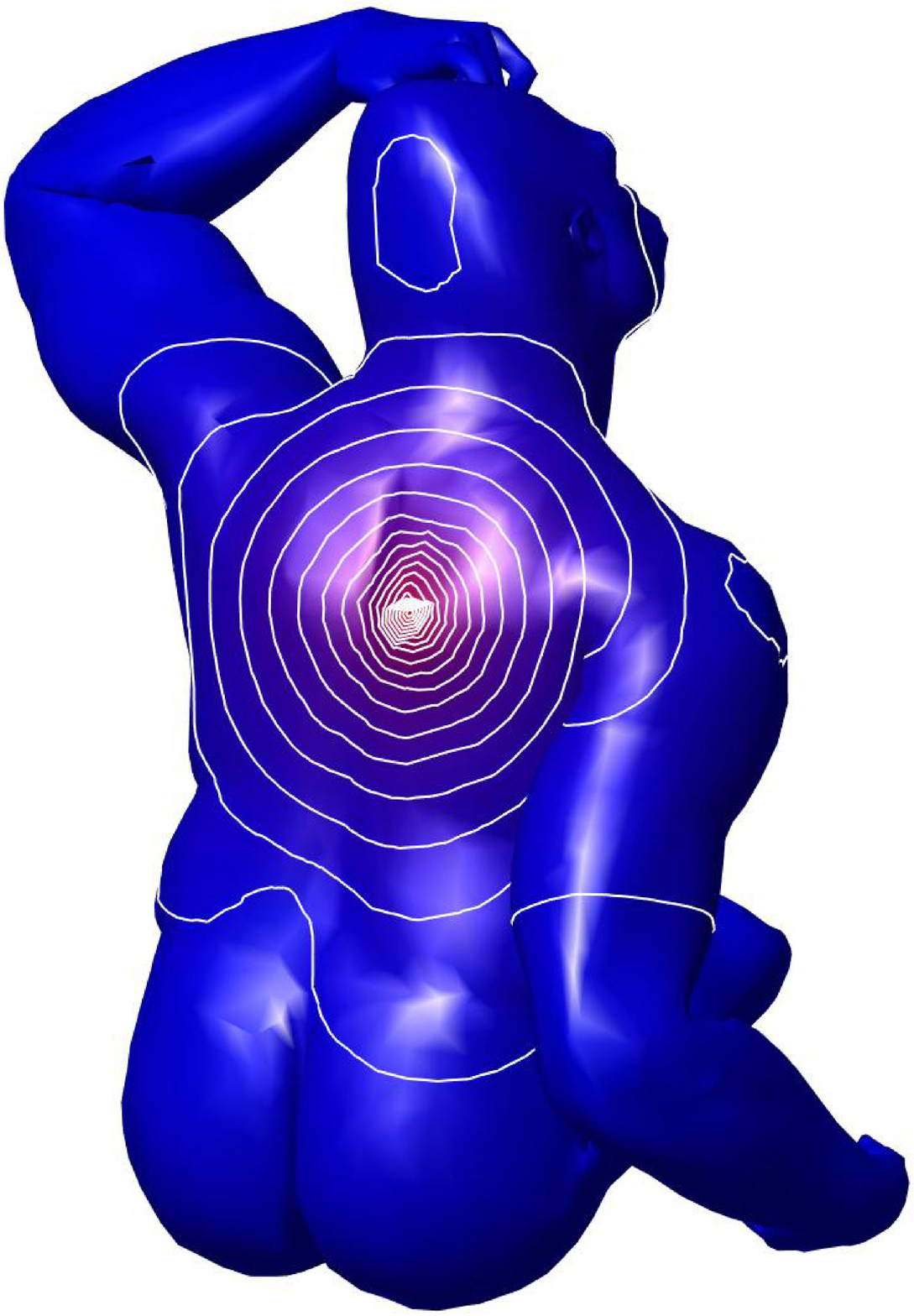}
\end{tabular}
\caption{Level-sets of bi-harmonic functions centred at 3 seed points on flat (spine) and rounded (harm, leg) surface regions.\label{fig:MONKEY-HARM}}
\end{figure}
\subsection{Fourier-based spectral operator\label{sec:GENERALISED-SPECTRAL-OP}}
Given a filter \mbox{$\varphi:\mathbb{R}\rightarrow\mathbb{R}$}, we consider the filter \mbox{$\widehat{\varphi}:\mathbb{R}\rightarrow\mathbb{R}$} induced by its \emph{continuous Fourier transform} \mbox{$\widehat{\varphi}(t):=\int\varphi(s)\exp(-ist)ds$} , and the \emph{Fourier-based spectral function} (c.f., Eq. (\ref{eq:SPECTRAL-OP})) \mbox{$\Phi_{\widehat{\varphi}}=\sum_{n=0}^{+\infty}\widehat{\varphi}(\lambda_{n})\phi_{n}$}, with \emph{real} \mbox{$\Phi_{\textrm{real}}(\widehat{\varphi})$} and \emph{imagery} \mbox{$\Phi_{\textrm{im}}(\widehat{\varphi})$} \emph{components} induced by the real and imagery parts of~$\widehat{\varphi}$. The spectral operator generalises the Fourier transforms of 1D filters on~$\mathbb{R}$ to signals defined on~$\mathcal{M}$, thus introducing a further flexibility in the design of spectral functions on arbitrary data that resembles the 1D case, as shown by the following properties.

\emph{Dilatation}~$\&$ \emph{Translation}: the Fourier-based spectral function of the filter \mbox{$\tilde{\varphi}(s):=\exp(is\alpha_{0})\varphi(s-s_{0})$} is \mbox{$F(\mathbf{p})=\sum_{n=0}^{+\infty}\exp(-i\lambda_{n}s_{0})\widehat{\varphi}(\lambda_{n}-\alpha_{0})\phi_{n}(\mathbf{p})$}.

\emph{Scaling}: the Fourier-based spectral function of the filter \mbox{$\tilde{\varphi}(s):=\varphi(\alpha s)$} is \mbox{$F(\mathbf{p})=\sum_{n=0}^{+\infty}\vert\alpha\vert^{-1}\widehat{\varphi}(\lambda_{n}/\alpha)\phi_{n}(\mathbf{p})$}.

\emph{Fourier transform}: the Fourier-based spectral function of the filter \mbox{$\tilde{\varphi}(s):=\widehat{\varphi}(s)$} is \mbox{$F(\mathbf{p})=\sum_{n=0}^{+\infty}\varphi(-\lambda_{n})\phi_{n}(\mathbf{p})$}.

\emph{Derivation}: the Fourier-based spectral function of the filter \mbox{$\tilde{\varphi}(s):=\varphi^{(k)}(s)$} is  \mbox{$F(\mathbf{p})=\sum_{n=0}^{+\infty}(2\pi i \lambda_{n})^{k}\widehat{\varphi}(\lambda_{n})\phi_{n}(\mathbf{p})$}.

\emph{Filter locality}: the Fourier-based spectral function of the filter \mbox{$\tilde{\varphi}(s):=\delta_{s_{0}}\varphi(s)$} is \mbox{$F(\mathbf{p})=\varphi(s_{0})\sum_{n=0}^{+\infty}\exp(-i \lambda_{n}s_{0})\phi_{n}(\mathbf{p})$}.

\emph{Exponential filter}: the Fourier-based spectral function of the filter  \mbox{$\tilde{\varphi}(s):=\exp(i\alpha_{0}s$)} is \mbox{$F(\mathbf{p})=2\pi\sum_{n:\,\lambda_{n}=\alpha_{0}}\phi_{n}(\mathbf{p})$}.

According to Eq. (\ref{eq:SPECTRAL-OP-WP}), we have the \emph{Parseval's equality} \mbox{$\|\Phi_{\widehat{\varphi}}\|_{2}=\|\widehat{\varphi}\|_{2}
=\|\varphi\|_{2}$}; in particular, the integrability of~$\varphi$ ensures the well-posedness of~$\Phi_{\varphi}$ and~$\Phi_{\widehat{\varphi}}$. Under the previous assumption and recalling the results in  Sect.~\ref{sec:SPECTRAL-OPERATOR},~$\Phi_{\widehat{\varphi}}$ is a linear and continuous operator and Eq. (\ref{eq:CONVOLUTION-DERIVATION}) still holds by replacing \mbox{$\widehat{\varphi_{1}}(n):=\langle f,\phi_{n}\rangle_{2}$}, \mbox{$\widehat{\varphi_{2}}(n):=\langle g,\phi_{n}\rangle_{2}$} with \mbox{$\widehat{\varphi_{1}}(\lambda_{n})$}, \mbox{$\widehat{\varphi_{2}}(\lambda_{n})$}, respectively. Applying the relation \mbox{$\widehat{\Phi_{\widehat{\varphi}}}=\Phi_{\widehat{\widehat{\varphi}}}=\Phi_{\varphi(-\cdot)}$} and assuming that input filter is even, the Fourier-based spectral operator induced by~$\widehat{\varphi}$ is equal to the spectral operator induced by the input filter, i.e., \mbox{$\widehat{\Phi}_{\widehat{\varphi}}=\Phi_{\varphi}$}. As we work with positive semi-definite operators and kernels, we always deal with filter functions defined on the positive real semi-axis. Indeed, we consider an even filter or redefine the filter on the negative semi-axis by symmetry in such a way that the resulting filter is even on~$\mathbb{R}$. 

\section{Spectral wavelets~$\&$ integral operators\label{sec:SPECTRAL-WAVELETS}}
We derive equivalent representations of the spectral wavelet operator (Sect.~\ref{sec:SPECTRAL-WAVELETS-DETAIL}), and its link with the spectral and filtered convolution operators and with existing kernels (Sect.~\ref{sec:SPECTRAL-EXAMPLES}).

\subsection{Continuous spectral wavelet\label{sec:SPECTRAL-WAVELETS-DETAIL}}
We define the \emph{continuous spectral wavelet} \mbox{$\Psi_{\varphi,\mathbf{p}}:\mathcal{M}\rightarrow\mathbb{R}$} centred at~$\mathbf{p}$ and associated with the filter \mbox{$\varphi:\mathbb{R}\rightarrow\mathbb{R}$} as~\cite{HAMMOND2010}
\begin{equation}\label{eq:SPECTRAL-WAVELET}
\Psi_{\varphi,\mathbf{p}}
:=\Psi_{\varphi}\delta_{\mathbf{p}}
=\sum_{n=0}^{+\infty}\varphi(\lambda_{n})\phi_{n}(\mathbf{p})\phi_{n}.
\end{equation}
Indeed, the filtered convolution operator (\ref{eq:SPECTRAL-CONV}) generalises the spectral wavelet (\ref{eq:SPECTRAL-WAVELET}); in fact, the spectral convolution operator~$\Psi_{\varphi}$ is achieved as the action of the spectral operator on the~$\delta$-function at~$\mathbf{p}$. Applying the identity \mbox{$(\Psi_{\varphi}f)(\mathbf{p})=\langle\Psi_{\varphi,\mathbf{p}}, f\rangle_{2}$} and Eq. (\ref{eq:SPECTRAL-INT-CONV}), the spectral convolution operator is also interpreted as the integral operator induced by the continuous spectral wavelet. Finally, the spectral wavelet (\ref{eq:SPECTRAL-WAVELET}) is continuous and its \mbox{$\mathcal{L}^{2}$}-energy is bounded by the \mbox{$\mathcal{L}^{2}(\mathbb{R})$} norm of the filter, i.e.,
\begin{equation}\label{eq:FUNCT-OPER-BOUND}
\|\Psi_{\varphi,\mathbf{p}}\|_{2}^{2}
=\sum_{n=0}^{+\infty}\vert\varphi(\lambda_{n})\vert^{2}\,\vert\phi_{n}(\mathbf{p})\vert^{2}
\leq\|\varphi\|^{2}_{2}.
\end{equation}
\textbf{Comparing wavelets induced by the same filter}
The~$\mathcal{L}^{2}(\mathcal{M})$ scalar product of the continuous wavelets induced by~$\varphi$ and centred at~$\mathbf{p}$,~$\mathbf{q}$ is bounded by the \mbox{$\mathcal{L}^{2}(\mathbb{R})$}-norm of~$\varphi$, i.e.,
\begin{equation*}
\begin{split}
&\left\vert\langle\Psi_{\varphi,\mathbf{p}},\Psi_{\varphi,\mathbf{q}}\rangle_{2}\right\vert
=\left\vert\sum_{n=0}^{+\infty}\vert\varphi(\lambda_{n})\vert^{2}\int_{\mathcal{M}\times\mathcal{M}}\phi_{n}(\mathbf{p})\phi_{n}(\mathbf{q})d\mathbf{p}d\mathbf{q}\right\vert\\
&=\sum_{n=0}^{+\infty}\vert\varphi(\lambda_{n})\vert^{2}\left(\int_{\mathcal{M}}\phi_{n}(\mathbf{p})d\mathbf{p}\right)^{2}
\leq \sum_{n=0}^{+\infty}\vert\varphi(\lambda_{n})\vert^{2}
\leq \|\varphi\|_{2}^{2}.
\end{split}
\end{equation*}
\textbf{Comparing wavelets induced by different filters}
The variation of the spectral wavelets induced by the filters~$\varphi_{1}$,~$\varphi_{2}$ and centred at~$\mathbf{p}$ is bounded as
\begin{equation*}
{\small{
\begin{split}
&\vert\Psi_{\varphi_{1},\mathbf{p}}(\mathbf{q})-\Psi_{\varphi_{2},\mathbf{p}}(\mathbf{q})\vert
\leq\sum_{n=0}^{+\infty}\vert\varphi_{1}(\lambda_{n})-\varphi_{2}(\lambda_{n})\vert\,\vert\phi_{n}(\mathbf{p})\vert\,\vert\phi_{n}(\mathbf{q})\vert\\
&\leq C_{\infty}\,C_{\mathbf{p}}\,C_{\mathbf{q}},\,
C_{\infty}:=\|\varphi_{1}-\varphi_{2}\|_{\infty},\,
C_{\mathbf{p}}:=\left[\sum_{n=0}^{+\infty}\vert\phi_{n}(\mathbf{p})\vert^{2}\right]^{1/2},
\end{split}}}
\end{equation*}
where we have applied the Cauchy-Scwartz inequality in the last relation. Indeed, the maximum variation of the spectral wavelets~$\Psi_{\varphi_{1},\mathbf{p}}$,~$\Psi_{\varphi_{2},\mathbf{p}}$ is bounded by the maximum variation~$C_{\infty}$ of the corresponding filters and by the values of the function \mbox{$C(\cdot)$} at the center~$\mathbf{p}$ and at the evaluation point~$\mathbf{q}$.
Finally, the~$\mathcal{L}^{2}(\mathcal{M})$-distance between the wavelets~$\Psi_{\varphi_{1},\mathbf{p}}$ and~$\Psi_{\varphi_{2},\mathbf{p}}$ is bounded as
\begin{equation*}\label{eq:WAVELET-BOUND}
\begin{split}
\|\Psi_{\varphi_{1},\mathbf{p}}-\Psi_{\varphi_{2},\mathbf{p}}\|_{2}^{2}
&=\sum_{n=0}^{+\infty}\vert\varphi_{1}(\lambda_{n})-\varphi_{2}(\lambda_{n})\vert^{2}\,\vert\phi_{n}(\mathbf{p})\vert^{2}\\
&\leq C_{\infty}^{2}\sum_{n=0}^{+\infty}\vert\phi_{n}(\mathbf{p})\vert^{2},
\end{split}
\end{equation*}
and its integral over~$\mathcal{M}$
\begin{equation}\label{eq:INTEGRAL-BOUND}
\begin{split}
\int_{\mathcal{M}}\|\Psi_{\varphi_{1},\mathbf{p}}-\Psi_{\varphi_{2},\mathbf{p}}\|_{2}^{2}d\mathbf{p}
&=\sum_{n=0}^{+\infty}\vert\varphi_{1}(\lambda_{n})-\varphi_{2}(\lambda_{n})\vert^{2}\\
&\leq\|\varphi_{1}-\varphi_{2}\|_{2}^{2}
\end{split}
\end{equation}
is bounded by the~$\mathcal{L}^{2}(\mathbb{R})$-distance of the corresponding filters. Indeed, the distance of the filters determines the maximum distance of the corresponding spectral wavelets, and close filters generally correspond to close spectral wavelets. Finally, we notice the analogy of Eq. (\ref{eq:INTEGRAL-BOUND}) with the upper bounds related to the approximation of the spectral (\ref{eq:SPECTRAL-BOUND}) and filtered convolution (\ref{eq:CONVOLUTION-BOUND}) operators.
\begin{figure}[t]
\centering
\begin{tabular}{c|ccc}
\multicolumn{1}{c}{Spectral kernel} &\multicolumn{3}{c}{Diffusion kernel}\\
\includegraphics[height=60pt]{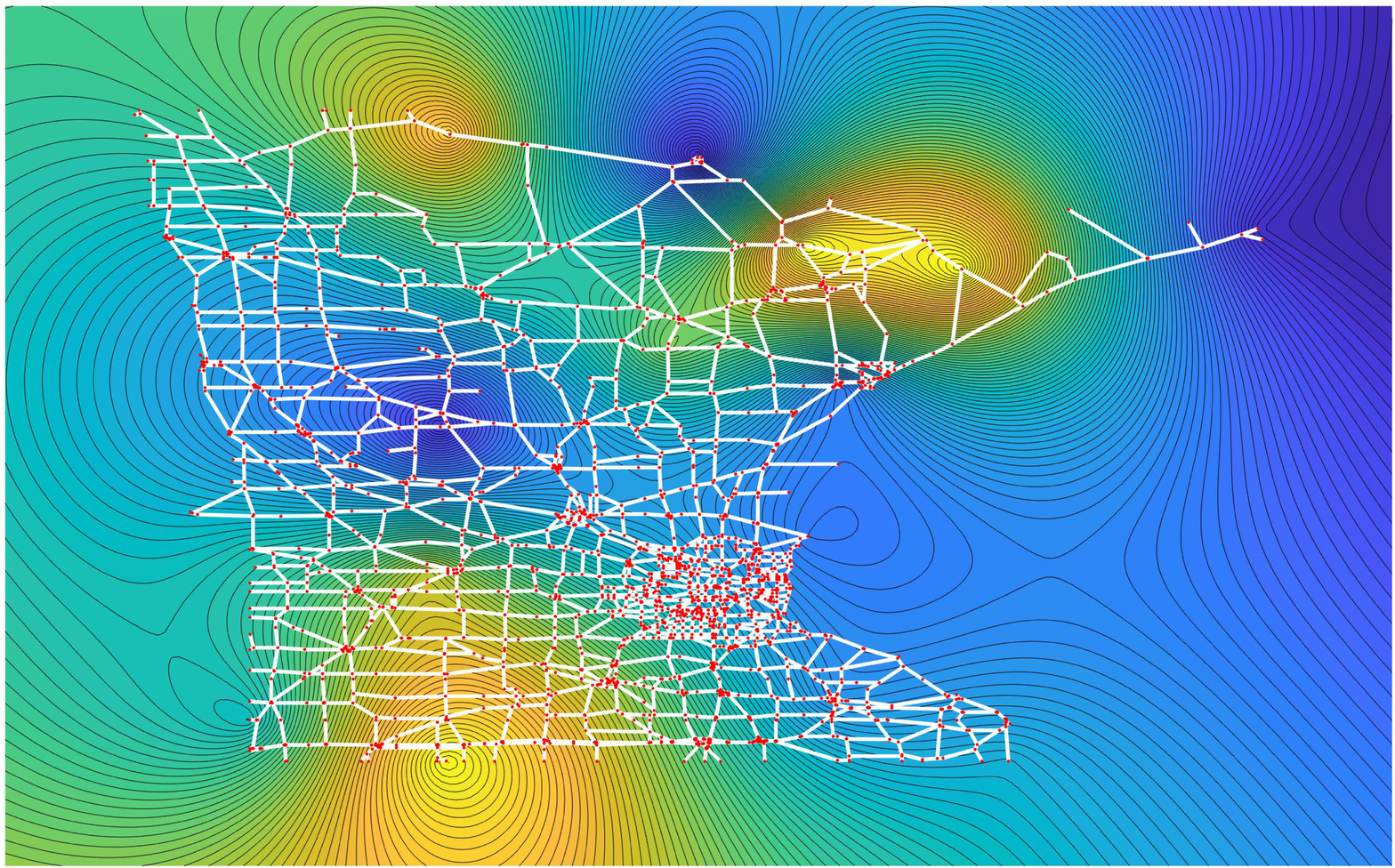}
&\includegraphics[height=60pt]{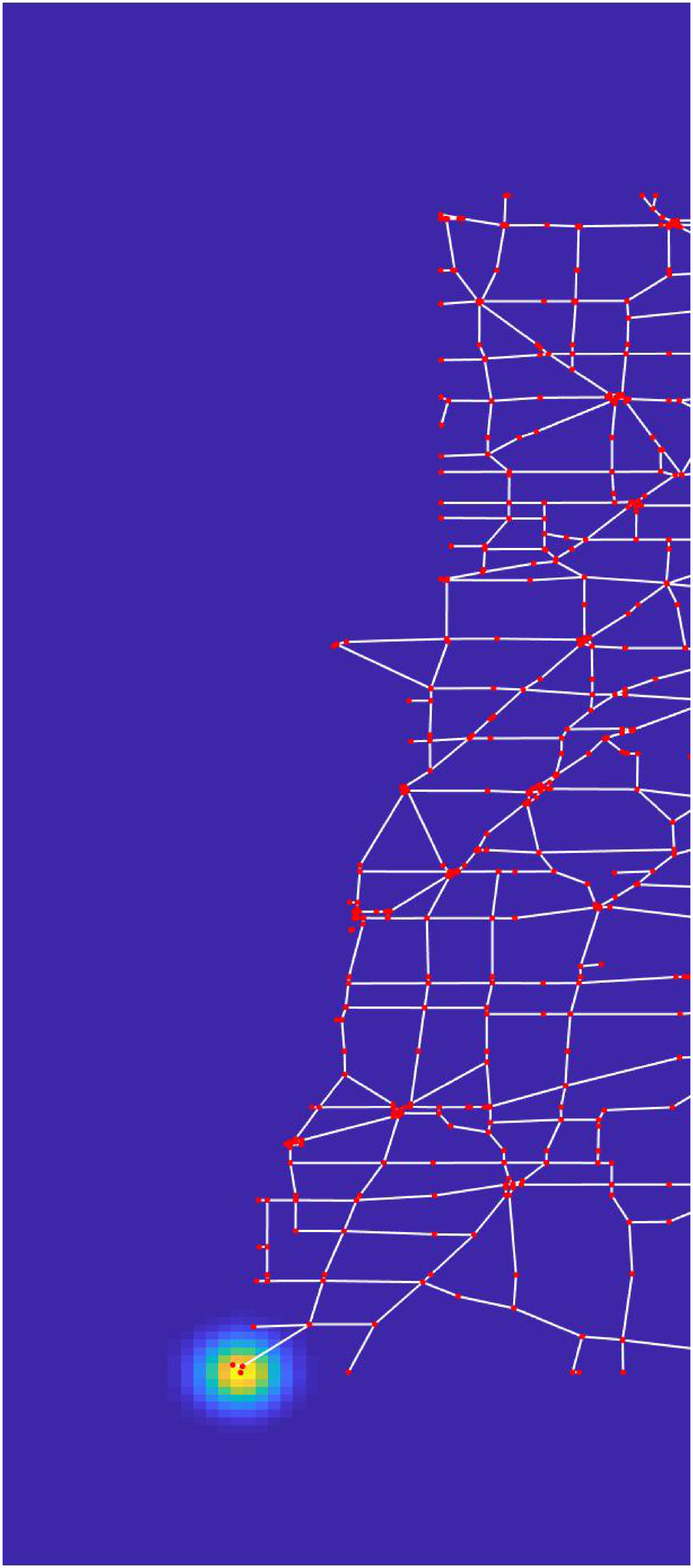}
&\includegraphics[height=60pt]{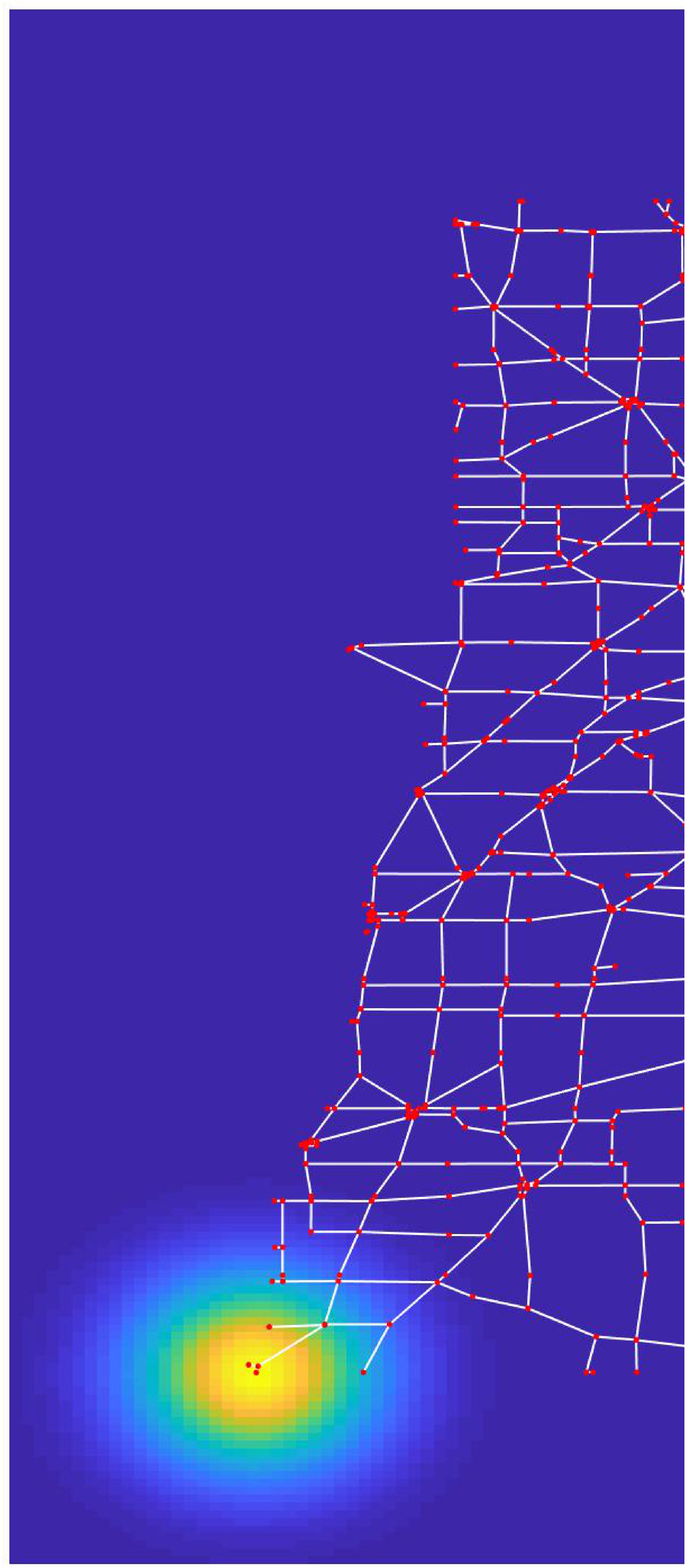}
&\includegraphics[height=60pt]{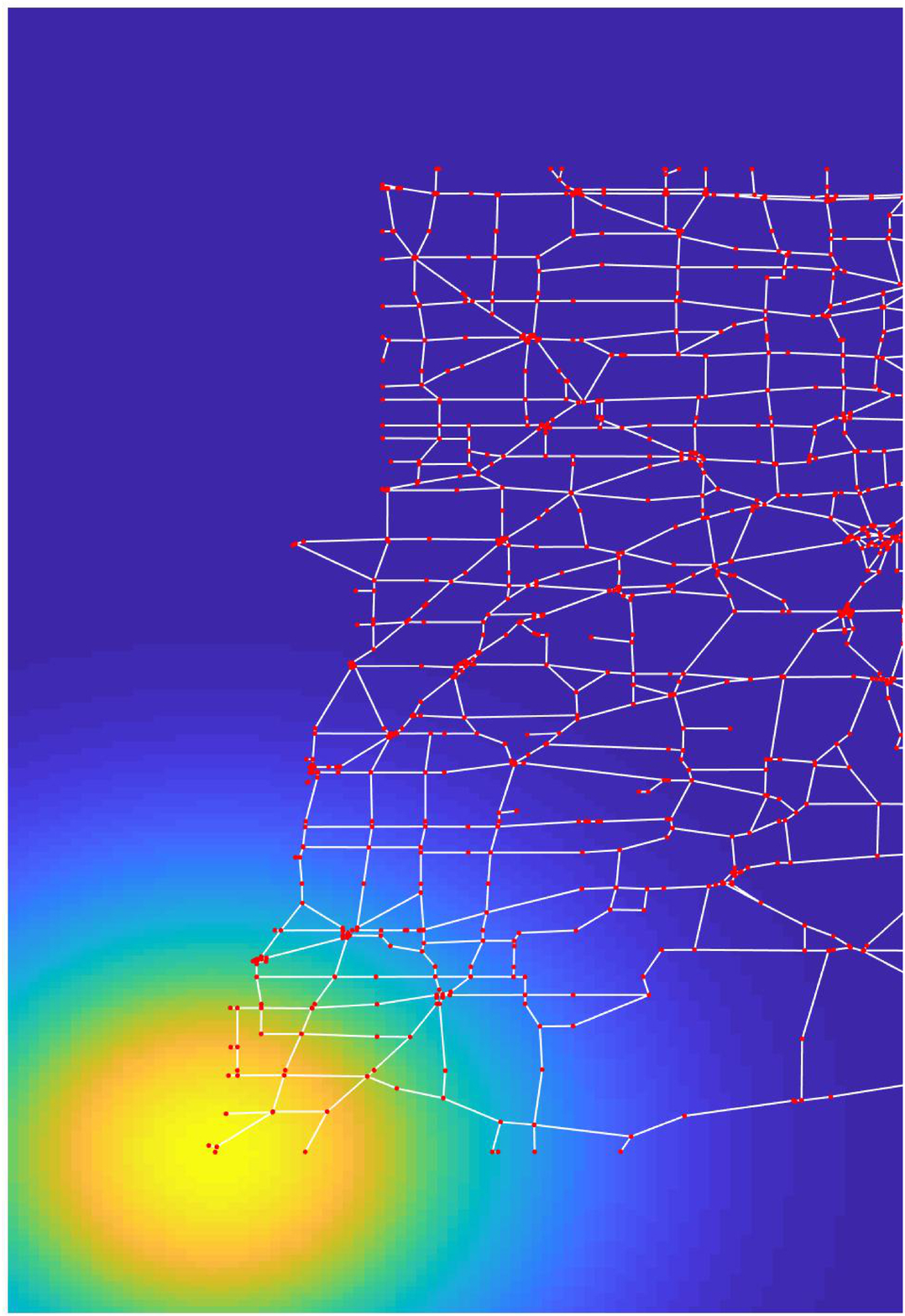}\\
~$\varphi$ &$t=0.001$ &$t=0.01$ &$t=0.1$
\end{tabular}
\caption{Input graph, colour-map, and level-sets of spectral kernels induced by (a) an analytic and (b-d) diffusion filters. (b-d) At small scales (\mbox{$t=10^{-3}$}, \mbox{$t=10^{-2}$}), the distribution of the level-sets of the diffusion kernels have a local and multi-scale behaviour at the same seeds. At large scales (\mbox{$t=10^{-1}$}), diffusion kernels are no more centred at seeds and have an analogous global behaviour.\label{fig:BIHARMONIC-MEMBER-FUNCTION}}
\end{figure}
\subsection{Examples of spectral wavelet operators\label{sec:SPECTRAL-EXAMPLES}}
We report main examples of spectral wavelets (\ref{eq:SPECTRAL-WAVELET}), defined by properly selecting the filter function, thus showing their link with filtered convolution and spectral kernels.

\emph{Commute-time wavelet/kernel} induced by the spectral operator \mbox{$\Psi_{\varphi}=\Delta^{\dag}$} and the filter \mbox{$\varphi(s):=s^{-1}$}, i.e., \mbox{$\Psi_{\varphi,\mathbf{p}}(\mathbf{q})=K_{\varphi}(\mathbf{p},\mathbf{q})=\sum_{n=1}^{+\infty}\lambda_{n}^{-1}\phi_{n}(\mathbf{p})\phi_{n}(\mathbf{q})$} (Fig.~\ref{fig:MONKEY-DIFFUSION}).

\emph{Bi-harmonic wavelet/kernel} \mbox{$\Psi_{\varphi,\mathbf{p}}(\mathbf{q})
=\sum_{n=1}^{+\infty}\lambda_{n}^{-2}\phi_{n}(\mathbf{p})\phi_{n}(\mathbf{q})$} induced by \mbox{$\Psi_{\varphi}:=(\Delta^{\dag})^{2}$} and \mbox{$\varphi(s):=s^{2}$} (Fig.~\ref{fig:MONKEY-HARM}).

\emph{Diffusion wavelet/kernel} induced by the spectral operator \mbox{$\Psi_{\varphi}=\exp(-t\Delta)$} and the filter \mbox{$\varphi(s):=\exp(-st)$} (Fig.~\ref{fig:MONKEY-DIFFUSION}), i.e. \mbox{$\Psi_{\varphi,\mathbf{p}}(\mathbf{q}):=\sum_{n=1}^{+\infty}\exp(-\lambda_{n}t)\phi_{n}(\mathbf{p})\phi_{n}(\mathbf{q})$}.

The commute-time and bi-harmonic wavelets, or equivalently kernels, are globally-supported. Increasing or reducing the time scale~$t$ of the exponential filter, we easily define globally-supported and locally-supported diffusion wavelets. In fact, as~$t$ becomes smaller the support of the corresponding diffusion wavelet centred at a seed point reduces until it degenerates to the seed itself.

\begin{figure}[t]
\centering
\begin{tabular}{ccc}
\includegraphics[height=60pt]{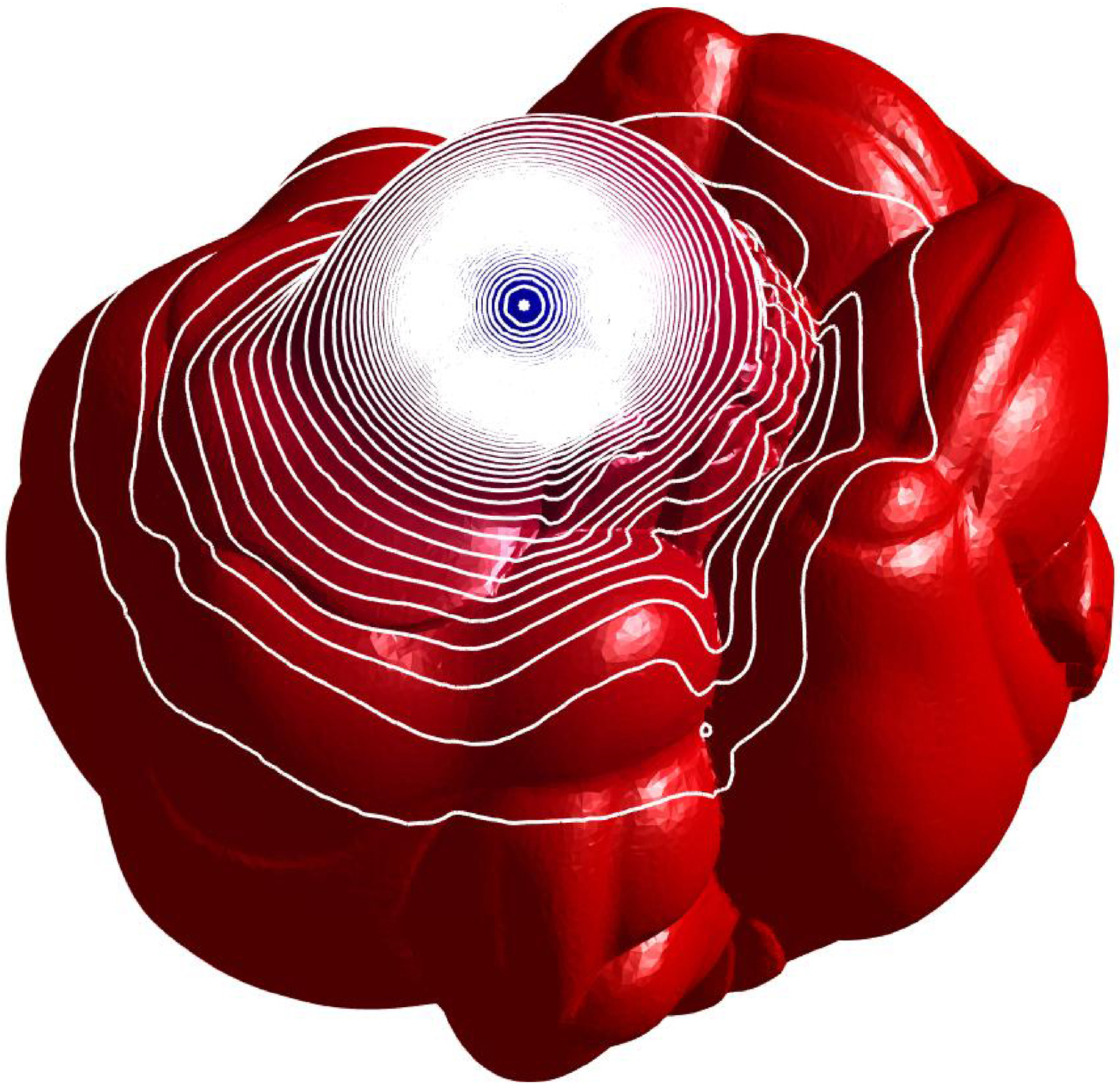}
&\includegraphics[height=60pt]{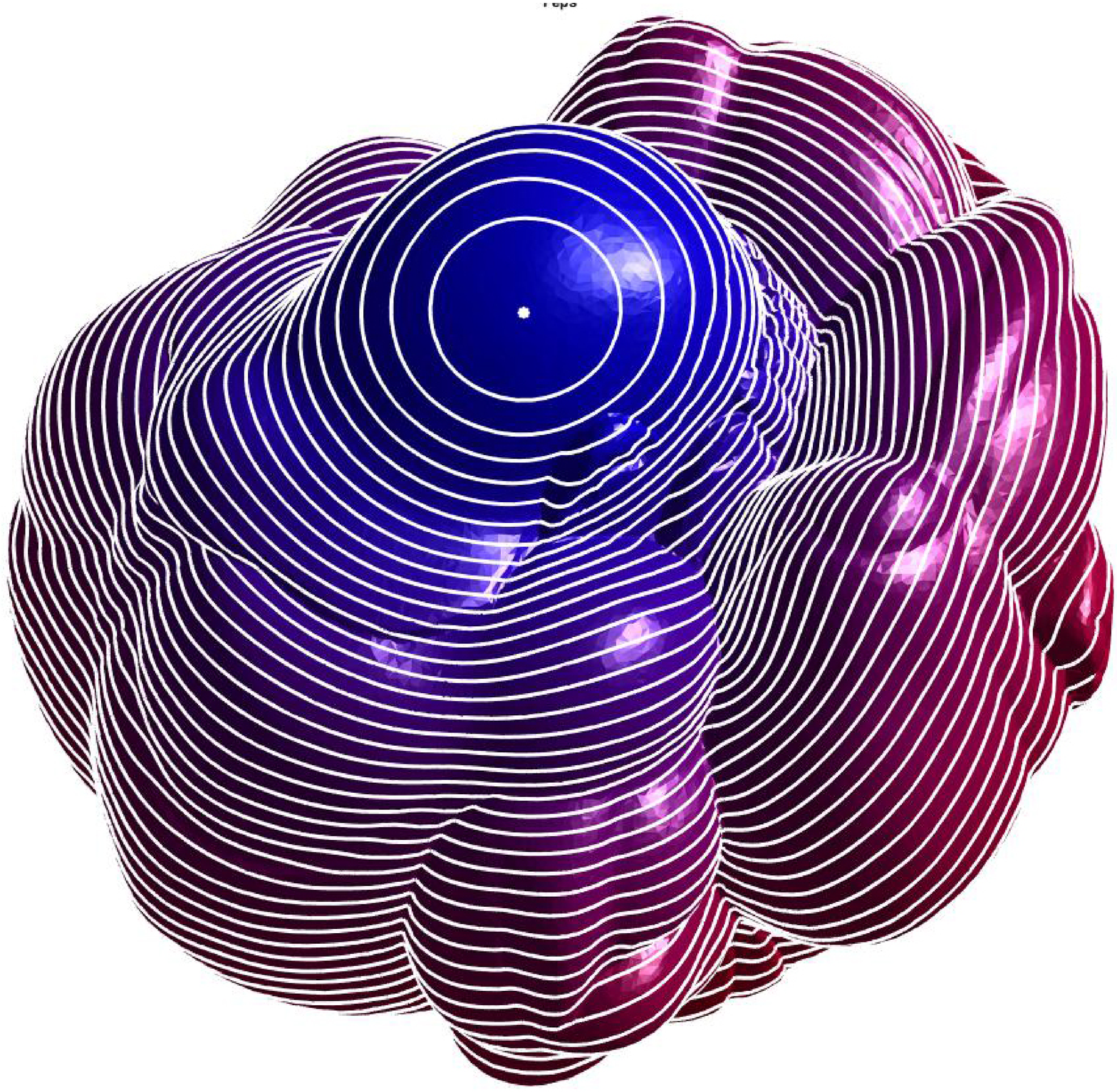}
&\includegraphics[height=60pt]{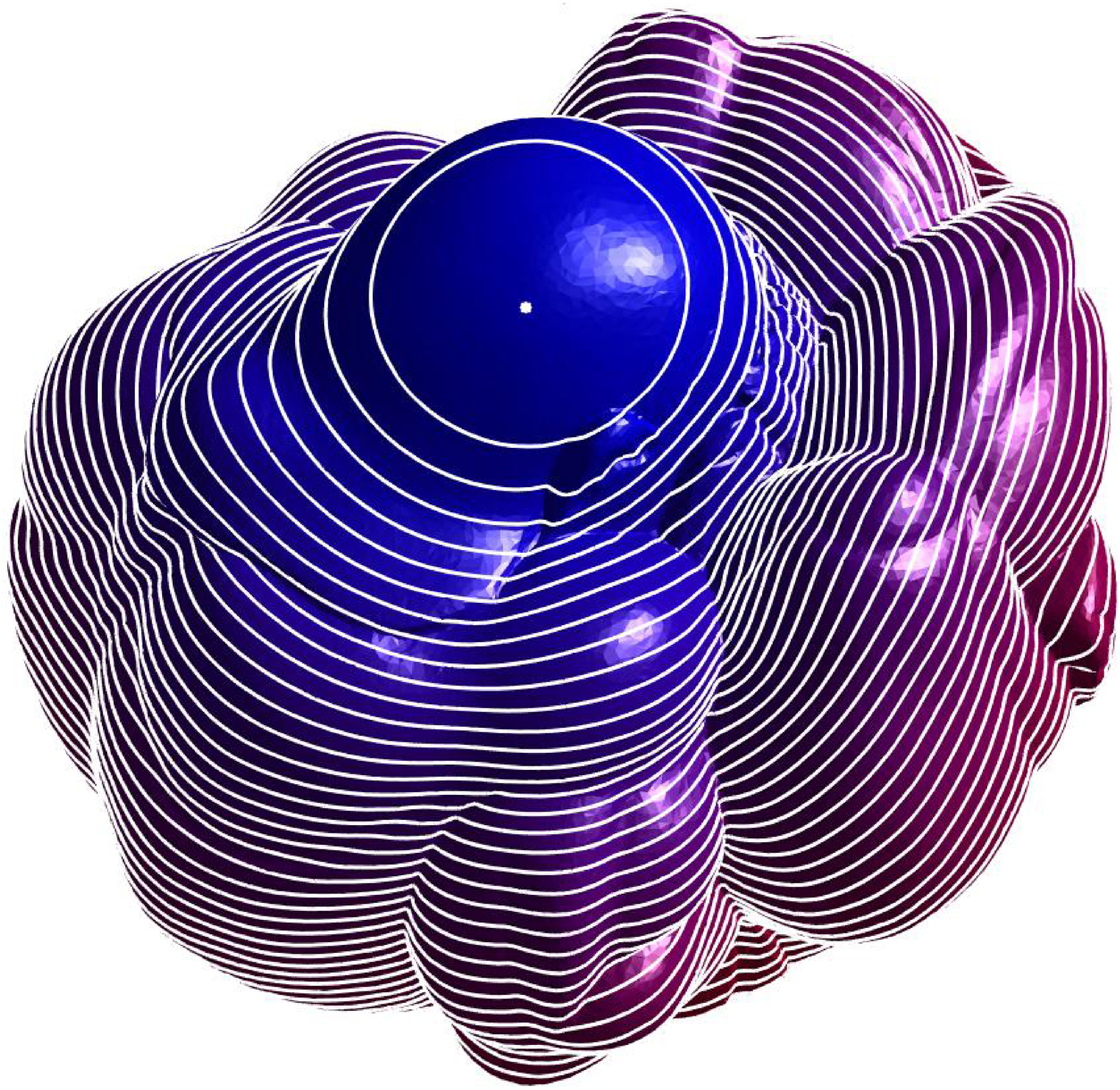}\\
$\varphi_{t}(s)=s^{-1}$ &$\varphi_{t}(s)=s^{-2}$
&$\varphi(s)=s^{-3}$ 
\end{tabular}
\caption{Level sets of the spectral distances from a source point (white dot) induced by the filter~$\varphi$ and evaluated with the Pad\'e-Chebyshev approximation (\mbox{$r=5$}).\label{fig:DIFF-EQUATION}}
\end{figure}

The previous definitions and properties of the spectral kernels/wavelets apply to any signal on a discrete domain, where we can discretise the Laplace-Beltrami operator. To show this generality of the proposed approach, we compute the spectral kernels on graphs (Fig.~\ref{fig:BIHARMONIC-MEMBER-FUNCTION}) and visualise their behaviour with density maps. Finally, through the rational polynomial basis we also approximate the \emph{spectral distance} \mbox{$d^{2}(\mathbf{p},\mathbf{q})
=\sum_{n=0}^{+\infty}\varphi^{2}(\lambda_{n})\vert\phi_{n}(\mathbf{p})-\phi_{n}(\mathbf{q})\vert^{2}$}, induced by arbitrary filers (Fig.~\ref{fig:DIFF-EQUATION}).
\section{Rational graph filters\label{sec:POL-RATIONAL-FILTERS}}
Firstly, we propose a fast computation of polynomial filters, which is based on a discrete scalar product induced by Chebyshev nodes (Sect.~\ref{sec:FAST-COMPUTATION}). Then, we introduce the class of \emph{rational filters} (Sect.~\ref{sec:DATA-DRIVEN-COMPUTATION}), which generalise the polynomial filters and are efficiently evaluated through recursive relations, in analogy to Chebyshev polynomials. As main benefits with respect to polynomial filters, we mention a higher approximation accuracy and stability, as they are not affected by undulations that typically affect polynomial approximations as the degree increases. According to these properties, we introduce a spectrum-free computation of spectral operators induced by arbitrary filters through their approximation with with rational polynomials.

\subsection{Fast computation of polynomial approximations\label{sec:FAST-COMPUTATION}}
The \emph{Chebyshev polynomials of the first kind} \mbox{$(T_{n})_{n=0}^{+
\infty}$} are defined by the identity \mbox{$\cos(n\theta)=T_{n}(\cos\theta)$}, e.g., \mbox{$T_{0}(x)=1$}, \mbox{$T_{1}(x)=x$}, \mbox{$T_{2}(x)=2x^{2}-1$}, etc.. Equivalently, these polynomials are defined through the recursive relation
\begin{equation}\label{eq:RECURSIVE-RELATION}
T_{0}(x)=1,\quad
T_{1}(x)=x,\quad
T_{n+1}(x)=2xT_{n}(x)-T_{n-1}(x).
\end{equation}
The Chebyshev polynomials of the first kind are orthogonal with respect to the weighted \mbox{$\mathcal{L}^{2}(\mathbb{R})$} scalar product induced by \mbox{$(1-x^{2})^{-1/2}$} on the interval \mbox{$[-1,1]$}. Indeed, a function \mbox{$f:[-1,1]\rightarrow\mathbb{R}$} is represented in terms of the Chebyshev polynomials as \mbox{$f(x)=\sum_{n=0}^{+\infty}a_{n}T_{n}(x)$}, whose coefficients are \mbox{$a_{0}=\pi^{-1}\langle f,1\rangle_{w}$}, \mbox{$a_{n}=(2\pi)^{-1}\langle f,T_{n}\rangle_{w}$}, \mbox{$n\neq 0$}.

The computation of the coefficients of the Chebyshev representation is generally time-consuming, as it requires to evaluate several integrals for an arbitrary filter. To overcome this limitation, we propose to approximate the weighted scalar product through the Chebyshev nodes in order to accurately compute the Chebyshev coefficients in linear time. Indicating with \mbox{$x_{k}:=\cos\left(\pi\frac{2k+1}{2N}\right)$}, \mbox{$k=0,\ldots,N-1$}, the~$N$ Chebyshev nodes, the Chebyshev polynomials are orthogonal with respect to the discrete scalar product
\begin{equation*}\label{eq:CHEB-ORTHO-DISCRETE}
\langle T_{n},T_{m}\rangle_{d}
:=\sum_{k=0}^{N-1}T_{n}(x_{k})T_{m}(x_{k})
=\left\{
\begin{array}{ll}
0	&n\neq m;\\
N	&n=m=0;\\
N/2	&n=m\neq 0.
\end{array}
\right.
\end{equation*}
According to the relations
\begin{equation*}
\langle f,T_{i}\rangle_{d}
\approx\sum_{k=0}^{N-1}a_{k}\langle T_{k},T_{i}\rangle_{d}
=\left\{
\begin{array}{ll}
0			&k\neq i;\\
Na_{0}		&k=i=0;\\
Na_{i}/2	&k\neq i\neq 0;
\end{array}
\right.
\end{equation*}
the coefficients are evaluated in linear time as
\begin{equation*}
a_{0}
=\frac{1}{N}\sum_{k=0}^{N-1}f(x_{k}),\qquad
a_{i}=\frac{2}{N}\sum_{k=0}^{N-1}f(x_{j})T_{i}(x_{j}).
\end{equation*}
\subsection{Rational spectrum-free computation\label{sec:DATA-DRIVEN-COMPUTATION}}
Noting that \mbox{$\|\Psi_{\varphi}-\Psi_{\rho}\|_{2}^{2}\leq\|\varphi-\rho\|_{2}$} (c.f., Eq. (\ref{eq:FUNCT-OPER-BOUND})), the approximation of~$\Psi_{\varphi}$ is reduced to the computation of a function \mbox{$\rho:\mathbb{R}\rightarrow\mathbb{R}$} that well approximates the filter~$\varphi$ with respect to the \mbox{$\mathcal{L}^{2}(\mathbb{R})$}-norm, i.e., thus solving 1D approximation problem. The representation of~$\rho$ must guarantee that~$\Psi_{\rho}$ can be efficiently computed through numerically robust algorithms, e.g., direct/iterative solvers of sparse, symmetric, and well-conditioned linear systems. 

\textbf{Rational approximation}
According to the Weierstrass approximation theorem, polynomial and rational polynomial approximations allow us to approximate any continuous function on an interval within an arbitrary tolerance. Even though polynomials are easily evaluated at arbitrary values through the Horner's method, and support an easy computation of derivatives and integrals, polynomial approximations tend to be oscillatory as their degree increases. Since rational polynomials are a reacher class of functions with respect to polynomials, as novel contribution, we propose to apply rational filters, which generally improve the approximation accuracy of arbitrary filters with respect to polynomials and are more stable with respect to oscillations. 
To approximate the input filter with a polynomials, we consider a rational polynomial approximation \mbox{$R(s):=P_{n}(s)/Q_{m}(s)$}, where~$P_{n}$,~$Q_{m}$ are polynomials of degree~$n$ and~$m$, respectively. In particular, we consider rational polynomials \mbox{$R(s)$} of the form
\begin{equation}\label{eq::RATIONAL-MONOMIAL}
R(s)
=\frac{P_{n}(s)}{Q_{m}(s)}
=\frac{\sum_{i=1}^{n}p_{i}s^{i}}{1+\sum_{i=1}^{m}q_{i}s^{i}},
\end{equation}
of degree \mbox{$N:=n+m$}. If \mbox{$q_{i}=0$}, \mbox{$i=1,\ldots,m$}, then the rational polynomial~$R$ reduces to a polynomial~$P_{n}$ of degree~$n$. The rational approximation is a generalisation of the polynomial expansion (i.e., \mbox{$Q_{m}:=1$}), and more generally, of the Taylor series. As rational polynomials generate polynomials, we expect that rational approximations of degree \mbox{$(n,m)$} are as good as polynomial approximations of degree \mbox{$(m+n)$}. 

\textbf{Pad\`e approximant}
The \emph{Pad\`e approximant} is the best approximation of a function with a rational polynomial such that the power series of the rational polynomial is the Taylor polynomial of the input function. Indeed, the Pad\`e approximation \mbox{$R(s)$} of order \mbox{$(n,m)$} of \mbox{$\varphi:\mathbb{R}\rightarrow\mathbb{R}$} in a neighbour of~$0$ is such that \mbox{$R^{(k)}(0)=\varphi^{(k)}(0)$}, \mbox{$k=0,\ldots,m+n$}. In particular, \mbox{$\varphi(s)-R^{(k)}(s)=\mathcal{O}(s^{m+n+1})$}, \mbox{$s\rightarrow 0$}. 
The Pad\`e approximant is uniquely defined if the constant term at the denominator has been set equal to~$1$ (c.f., Eq. (\ref{eq::RATIONAL-MONOMIAL})); otherwise, the approximant is defined up to a multiplication by a constant. For the computation of the coefficients of the polynomials~$P_{n}$,~$Q_{m}$, we can apply the Wylm's epsilon algorithm or the extended Euclidean algorithm~\cite{BAKER1996}, and the sequence transformation~\cite{BREZINSKY1991}. 
According~\cite{GOLUB1989,LITVINOV1993}, a rational approximation is more stable than a polynomial approximation, as the errors in the numerator and denominator of a rational polynomial compensate each other. Furthermore, rational polynomial approximations have been computed analytically for commonly used filters (e.g., sin/cos, exponential, logarithm). 
For the approximation and evaluation of the rational polynomial, we consider (i) the \emph{canonical polynomial basis} and the \emph{Chebyshev polynomial basis}, applied to~$P_{n}$ and~$Q_{m}$ (Sect.~\ref{sec:RAT-POL-BASIS}) and (ii) the \emph{Chebyshev rational polynomials} of the first kind, applied to~$R$ (Sect.~\ref{sec:RAT-CHEB-APPROX}). Among these options, the recursive representation of the Chebyshev rational polynomials of the first kind allows us to reduce the computation cost for the evaluation of the approximation and resembles the polynomial case. Finally, we discuss the accuracy and convergence of the corresponding rational approximation (Sect.~\ref{sec:ACCURACY}).

\subsubsection{Rational approximation with polynomial basis\label{sec:RAT-POL-BASIS}}
\textbf{Approximation through the canonical polynomial basis applied to~$P_{n}$ and~$Q_{m}$}
We discuss the computation of the coefficients of the rational polynomial in terms of the Taylor coefficients \mbox{$a_{k}:=\varphi^{(k)}(0)/k!$} of the input filter. Considering the power series \mbox{$\varphi(s):=\sum_{n=0}^{+\infty}a_{n}s^{n}$}, we have that
\begin{equation*}
\begin{split}
\varphi(s)-R(s)
&=\frac{\varphi(s)Q_{m}(s)-P_{n}(s)}{Q_{m}(s)}\\
&=\frac{\sum_{i=0}^{+\infty}a_{i}s^{i}\sum_{j=0}^{N}q_{j}s^{j}-\sum_{i=0}^{N}p_{i}s^{i}}{Q_{m}(s)}=0.
\end{split}
\end{equation*}
Defining the vectors \mbox{$(p_{i})_{i=n+1}^{N}=\mathbf{0}$}, \mbox{$(q_{i})_{i=m+1}^{N}=\mathbf{0}$}, the previous relation is equivalent to \mbox{$\sum_{i=0}^{+\infty}a_{i}q_{j}s^{i+j}
=\sum_{i=0}^{N}p_{i}s^{i}$}, \mbox{$j=0,\ldots,N$}. Introducing the index \mbox{$k:=i+j$}, this last relation is rewritten as \mbox{$\sum_{i=0}^{k}a_{i}q_{k-i}-p_{k}=0$}, \mbox{$k=0,\ldots,N$}. Recalling that \mbox{$q_{0}=1$}, we get \mbox{$a_{0}=p_{0}$} and the remaining unknowns are computed as the solution to the linear system
{\small{
\begin{equation*}
\left[
\begin{array}{llllll}
a_{1} 		&a_{0} 			&\ldots			&\ldots			&0\\
a_{2} 		&a_{1} 			&a_{0}			&\ldots 		&0\\
\vdots		&\vdots			&\vdots			&\vdots			&\vdots\\
a_{N}		&a_{N-1} 		&\ldots			&\ldots			&a_{0}
\end{array}
\right]
\left[
\begin{array}{l}
q_{1}\\
q_{2}\\
\vdots\\
q_{N}
\end{array}
\right]
-
\left[
\begin{array}{l}
p_{1}\\
p_{2}\\
\vdots\\
p_{N}
\end{array}
\right]
=
\left[
\begin{array}{l}
a_{1}\\
a_{2}\\
\vdots\\
a_{N}
\end{array}
\right].
\end{equation*}}}

\textbf{Approximation through the Chebyshev polynomial basis applied to~$P_{n}$ and~$Q_{m}$}
Instead of expanding numerator and denominator in terms of the monomial basis, we use the Chebyshev polynomials \mbox{$T_{k}$}, through the representation
\begin{equation}\label{eq:CHEB-POL-REL}
R:=\frac{\sum_{k=0}^{n}p_{k}T_{k}}{\sum_{k=0}^{m}q_{k}T_{k}},\qquad
N=n+m,\quad q_{0}=1.
\end{equation}
Expanding the input filter \mbox{$\varphi=\sum_{n=0}^{+\infty}a_{n}T_{n}$} in a series of Chebyshev polynomials, we get the relation
\begin{equation*}
\varphi-R
=\frac{\sum_{n=0}^{+\infty}a_{n}T_{n}\sum_{k=0}^{m}q_{k}T_{k}-\sum_{k=0}^{n}p_{k}T_{k}}{\sum_{k=0}^{m}q_{k}T_{k}}.
\end{equation*}
The coefficients \mbox{$(p_{i})_{i=0}^{n}$}, \mbox{$(q_{i})_{i=0}^{m}$} are chosen in such a way that the numerator has zero coefficient for \mbox{$(T_{k})_{k=0}^{k}$}, i.e., 
\begin{equation*}
\varphi-R
\sum_{n=0}^{+\infty}a_{n}T_{n}\sum_{k=0}^{m}q_{k}T_{k}-\sum_{k=0}^{n}p_{k}T_{k}
=\sum_{k=N+1}^{+\infty}\gamma T_{k}.
\end{equation*}
To this end, we apply the following relations
\begin{equation*}
\left\{
\begin{array}{ll}
T_{i}T_{j}=\frac{1}{2}[T_{i+j}-T_{\vert i-j\vert}], 
&a_{0}=\frac{1}{\pi}\int_{-1}^{-1}\frac{f(s)}{(1-s^{2})^{1/2}}ds,\\
a_{k}=\frac{2}{\pi}\int_{-1}^{1}\frac{f(s)T_{k}(s)}{(1-s^{2})^{1/2}}ds,
&k\geq 1,
\end{array}
\right.
\end{equation*}
whose evaluation is generally faster than monomial basis functions.

\textbf{Computation}
Once the rational approximation~$P_{n}/Q_{m}$ of~$\varphi$ has been computed with one of the two previous approaches, any signal \mbox{$\varphi(\Delta)f$} is computed as the solution to the problem \mbox{$Q_{m}(\Delta)g=P_{n}(\Delta)f$}. In case of the canonical basis (\ref{eq::RATIONAL-MONOMIAL}), we evaluate the right- and left-side terms of
\begin{equation}\label{eq:CANONICAL-POL-BASIS}
\sum_{k=0}^{m}q_{k}\Delta^{k}g=\sum_{k=0}^{n}p_{k}\Delta^{k}f.
\end{equation}
In case of the Chebyshev polynomials (\ref{eq:CHEB-POL-REL}), we evaluate the right- and left-side term of 
\begin{equation}\label{eq:CHEBYSHEV-BASIS}
\sum_{k=0}^{m}q_{k}T_{k}(\Delta)g=\sum_{k=0}^{n}p_{k}T_{k}(\Delta)f,
\end{equation}
through the recursive relations (\ref{eq:RECURSIVE-RELATION}).

\subsubsection{Rational approximation with rational basis\label{sec:RAT-CHEB-APPROX}}
\begin{figure}[t]
\centering
\begin{tabular}{cccc}
\includegraphics[height=50pt]{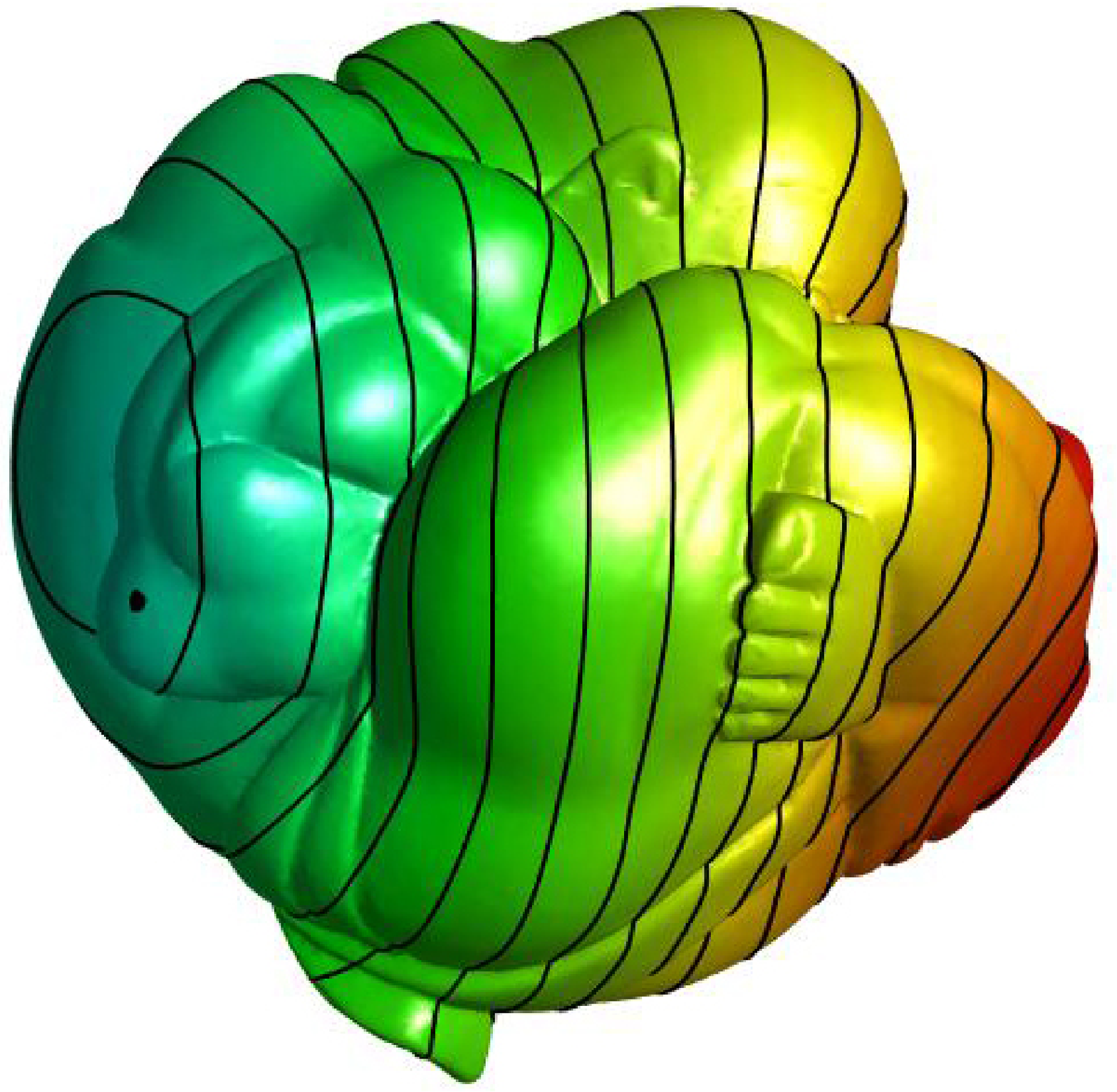}
&\includegraphics[height=50pt]{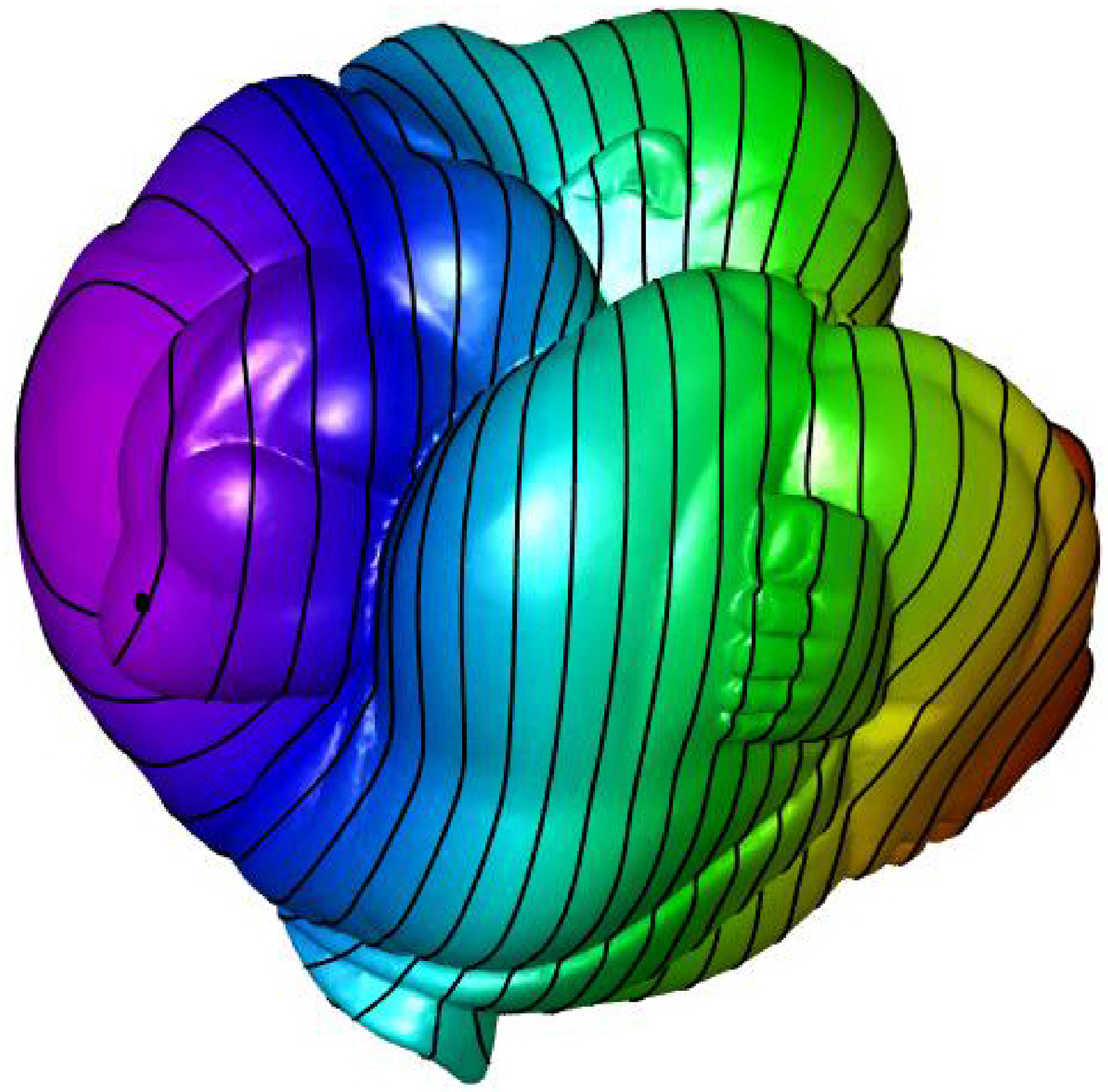}
&\includegraphics[height=50pt]{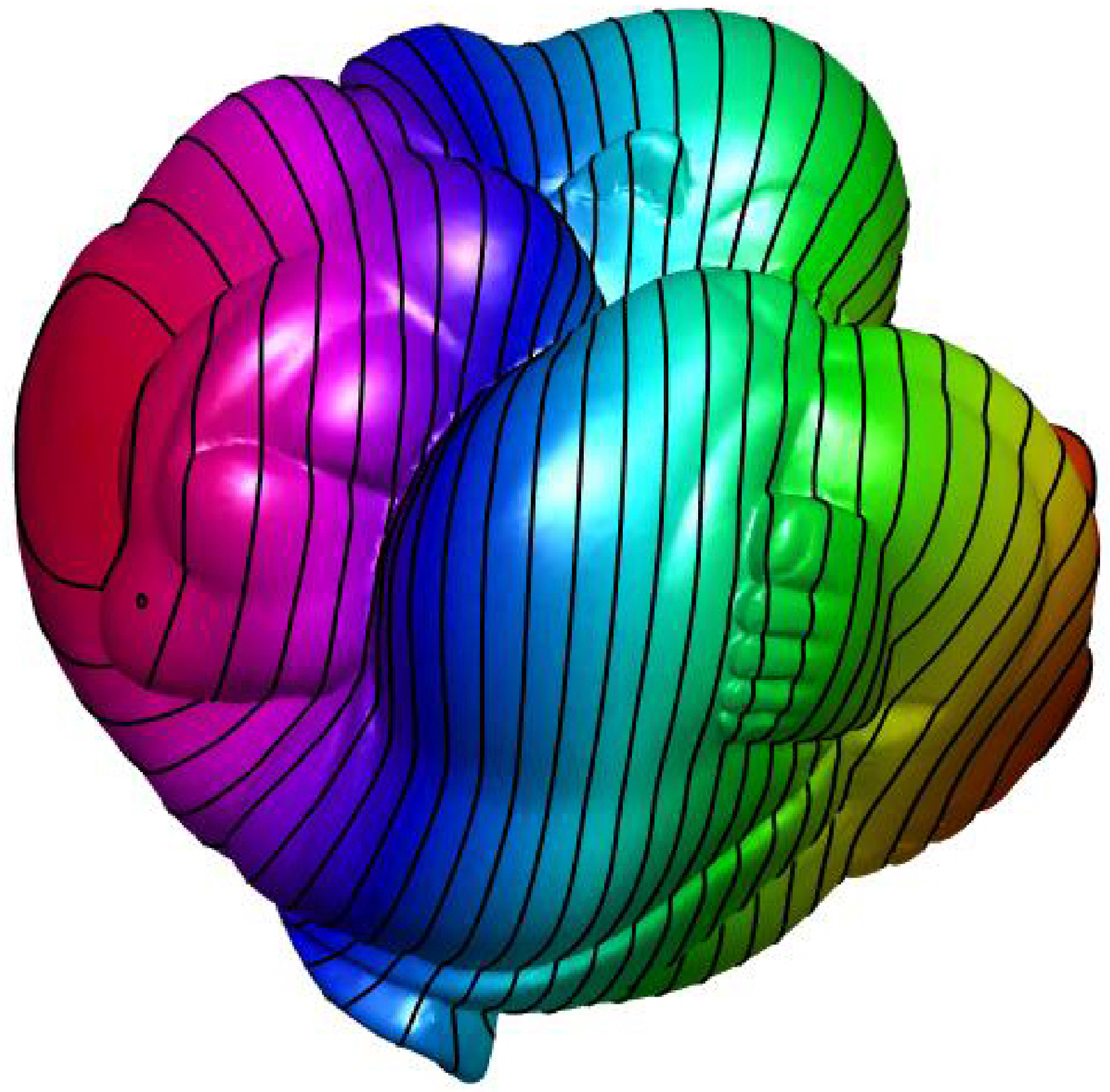}
&\includegraphics[height=50pt]{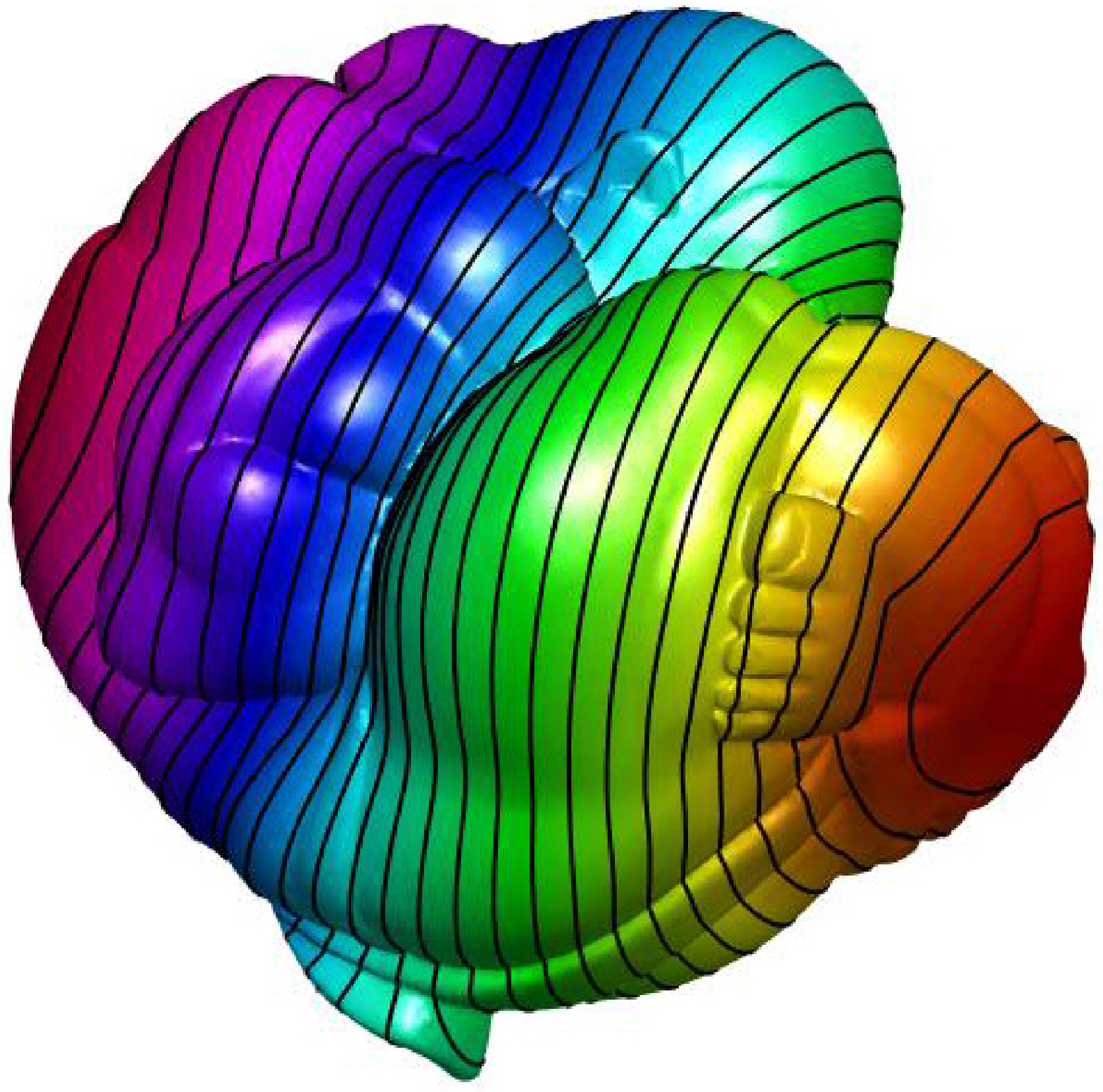}\\
$R_{1}(\Delta)\delta_{\mathbf{p}}$ &$R_{2}(\Delta)\delta_{\mathbf{p}}$
&$R_{3}(\Delta)\delta_{\mathbf{p}}$ &$R_{11}(\Delta)\delta_{\mathbf{p}}$\end{tabular}
\caption{Level-sets of rational graph filters centred at a seed point~$\mathbf{p}$ (black dot on the elbow) induced by the Chebyshev rational basis \mbox{$R_{k}(\Delta)\delta_{\mathbf{p}}$} with an increasing degree.\label{fig:RATIONAL-BASIS}}
\end{figure}

The \emph{Chebyshev rational polynomials} are defined as 
\begin{equation}\label{eq:RATIONAL-RECURSION-POLYNOMIAL}
R_{n}(x)
:=T_{n}\left(\frac{x-1}{x+1}\right)
=2\frac{x-1}{x+1}R_{n-1}(x)-R_{n-2}
\end{equation}
on the interval \mbox{$[0,+\infty)$}. These rational polynomials are orthogonal with respect to the weighted scalar product induced by \mbox{$w(x):=x^{-1/2}(1+x)^{-1}$}, according to the relation
\begin{equation*}\label{eq:CHEB-RATIONAL-ORTHO}
\langle R_{n},R_{m}\rangle_{w}
=\int_{0}^{+\infty}\frac{R_{n}(x)R_{m}(x)}{x^{1/2}(x+1)}dx
=\left\{
\begin{array}{ll}
0		&n\neq m;\\
\pi		&n=m=0;\\
\pi/2	&n=m=0.
\end{array}
\right.
\end{equation*}
For an arbitrary function \mbox{$f\in\mathcal{L}^{2}(\mathbb{R})$}, the orthogonality of the Chebyshev rational polynomials allows us to apply the relation \mbox{$f(x)=\sum_{n=0}^{+\infty} F_{n}R_{n}(x)$}, \mbox{$F_{n}:=\langle f, R_{n}\rangle_{w}$}. In Fig.~\ref{fig:RATIONAL-BASIS}, we show the level-sets of rational graph filters \mbox{$R_{k}(\Delta)\delta_{\mathbf{p}}$} (c.f., Eq. (\ref{eq:RATIONAL-RECURSION-POLYNOMIAL})) centred at a seed point~$\mathbf{p}$ (black dot) induced by Chebyshev rational polynomials with increasing degree.

\textbf{Computation}
Through the recursive relation (\ref{eq:RATIONAL-RECURSION-POLYNOMIAL}), the rational graph filter is efficiently evaluated as
\begin{equation}\label{eq:RATIONAL-RECURSION}
\begin{split}
f_{n}
:=R_{n}(\Delta)f
&=2(\Delta+\textrm{id})^{-1}(\Delta-\textrm{id})f_{n-1}-f_{n-2}\\
&=2g_{n-1}-f_{n-2},
\end{split}
\end{equation}
where~$g_{n-1}$ is the solution to \mbox{$(\Delta+\textrm{id})g_{n-1}=(\Delta-\textrm{id})f_{n-1}$}. We notice that~$g_{n-1}$ is uniquely defined, as the operator \mbox{$(\Delta+\textrm{id})$} is positive-definite.
\subsubsection{Approximation accuracy and convergence\label{sec:ACCURACY}}
Approximating the input filter with a rational polynomial~$\rho^{(r)}$ of degree \mbox{$(n,m)$} (e.g., \mbox{$m=n=r$}), the sequence \mbox{$(\Psi_{\rho^{(r)}}f)_{r=0}^{+\infty}$}, \mbox{$\Psi_{\rho^{(r)}}f
:=\sum_{n=0}^{+\infty}\rho^{(r)}(\lambda_{n})\langle f,\phi_{n}\rangle_{2}\phi_{n}$}, induced by the rational polynomial approximation~$\rho^{(r)}$ of~$\varphi$, converges to \mbox{$\Psi_{\varphi}f$}; in fact,
\begin{equation*}\label{eq:CONTINUOUS-APPROX-ACCURCACY}
\left\|\Psi_{\rho^{(r)}}f-\Psi_{\varphi}f\right\|_{2}^{2}
\leq \|\rho^{(r)}-\varphi\|_{\infty}^{2}\sum_{n=0}^{+\infty}\vert\langle f,\phi_{n}\rangle_{2}\vert^{2}
=\sigma_{r}^{2}\|f\|_{2}^{2},
\end{equation*}
where~$\sigma_{r}\approx\mathcal{O}(s^{m+n+1})$, \mbox{$s\rightarrow 0$}, is the approximation error between~$\varphi$ and~$\rho^{(r)}$.

\section{Numerical computation\label{sec:NUMERICAL-METHODS}}
Even though a central element in spectral graph processing is the evaluation of the spectrum, or equivalently, of the characteristic polynomial of the graph Laplacian, the fast computation of the characteristic polynomial and numerical instabilities associated with multiple or close eigenvalues has not been addressed in detail by previous work. As novel contributions, we focus on the fast approximation of the characteristic polynomial and on the definition of the pseudo-spectrum, which allows us to identify a subset of the eigenvalues that is robust with respect to data perturbation. To this end, we introduce the discrete spectral wavelets and kernels (Sect.~\ref{sec:DISCRETE-OPER-DISTANCE}); then, we define the pseudo-spectrum and spectral density for the approximation of the characteristic polynomial (Sect.~\ref{sec:POLY-APPROX}). Finally, we discuss the evaluation of the rational basis, the numerical stability and computational cost of the proposed approach (Sect.~\ref{sec:COMPUTATIONAL-COST}).

\subsection{Discrete spectral wavelets/kernels\label{sec:DISCRETE-OPER-DISTANCE}} 
\textbf{Graph Laplacian}
For the discretisation of the spectral operators, wavelets, and filtered convolution operators, let us consider a graph~$\mathcal{M}$ and a signal \mbox{$f:\mathcal{M}\rightarrow\mathbb{R}$} identified with the vector \mbox{$\mathbf{f}:=(f(\mathbf{p}_{i}))_{i=1}^{n}$} of~$f$-values at the nodes of the input graph. The \emph{graph Laplacian} is defined as \mbox{$\tilde{\mathbf{L}}:=\mathbf{D}^{-1}\mathbf{L}$}, where \mbox{$\mathbf{L}:=\mathbf{W}-\mathbf{D}$}. Here,~$\mathbf{W}$ is the weight matrix whose entry \mbox{$(i,j)$} is the weight associated with the corresponding edge, and~$\mathbf{D}$ is the diagonal matrix whose entries are the sum of the rows of~$\mathbf{L}$. Alternatively, we can consider the \emph{normalised graph Laplacian} \mbox{$\tilde{\mathbf{L}}:=\mathbf{D}^{-1/2}\mathbf{L}\mathbf{D}^{-1/2}$}. We notice that the graph Laplacian is~$\mathbf{D}$-adjoint, i.e., \mbox{$\langle\tilde{\mathbf{L}}\mathbf{f},\mathbf{g}\rangle_{\mathbf{D}}=\langle\mathbf{f},\tilde{\mathbf{L}}\mathbf{g}\rangle_{\mathbf{D}}$}. In our discussion, we focus on the graph Laplacian; analogous considerations can be derived for the normalised graph Laplacian. For the graph Laplacian, the spectral decomposition is \mbox{$\mathbf{L}\mathbf{X}=\mathbf{D}\mathbf{X}\Lambda$}, \mbox{$\mathbf{X}^{\top}\mathbf{D}\mathbf{X}=\mathbf{I}$}, where \mbox{$\mathbf{X}:=[\mathbf{x}_{1},\ldots,\mathbf{x}_{n}]$} is the eigenvectors' matrix and~$\Lambda$ is the diagonal matrix of the eigenvalues \mbox{$(\lambda_{i})_{i=1}^{n}$}. 

\textbf{Discrete spectral wavelets and kernels}
Noting that \mbox{$(\varphi(\lambda_{n}),\phi_{n})_{n=0}^{+\infty}$} is the eigensystem of~$\Psi_{\varphi}$, the spectral representation of the data-driven kernel is~$\mathbf{K}_{\varphi}$ such that 
\begin{equation*}
\varphi(\lambda_{i})
=\langle\Psi_{\varphi}\phi_{i},\phi_{j}\rangle_{2}
=\mathbf{x}_{i}^{\top}\mathbf{K}_{\varphi}^{\top}\mathbf{D}\mathbf{x}_{j},\quad \forall i=1,\ldots,n.
\end{equation*}
Indeed, \mbox{$\mathbf{K}_{\varphi}=\mathbf{X}\varphi(\Lambda)\mathbf{X}^{\top}\mathbf{D}$} is the \emph{spectral kernel}, which is a filtered version of the Laplacian matrix and~$\mathbf{D}$-adjoint. According to (\ref{eq:CONVOLUTION-BOUND}), the approximation of~$\mathbf{K}_{\rho}$ with a new kernel~$\mathbf{K}_{\varphi}$ reduces to the approximation of the corresponding filters. The approximation~$\varphi$ of~$\rho$ is computed on the interval \mbox{$[0,\lambda_{\max}(\tilde{\mathbf{L}})]$}, where the maximum Laplacian eigenvalue is evaluated by the Arnoldi method~\cite{GOLUB1989}, or is set equal to the upper bound~\cite{LEHOUCQ1996,SORENSEN1992}
\begin{equation*}
\lambda_{\max}(\tilde{\mathbf{L}})\leq \min\{\max_{i}\{\sum_{j}\tilde{L}(i,j)\},\max_{j}\{\sum_{i}\tilde{L}(i,j)\}\}. 
\end{equation*}
In the discrete setting, we proceed as done for the approximation and computation of~$\Phi_{\varphi}$ in Sect.~\ref{sec:DATA-DRIVEN-COMPUTATION},  by replacing the Laplace-Beltrami operator with the Laplacian matrix.
\begin{figure}[t]
\centering
\begin{tabular}{cc|cc}
\multicolumn{2}{c}{P.-C. approx.~$r:=7$}	&\multicolumn{2}{c}{Trunc. approx. (100 eigs)}\\
\hline
$t=0.01$		&$t=0.001$		&$t=0.01$		&$t=0.001$\\
\includegraphics[height=35pt]{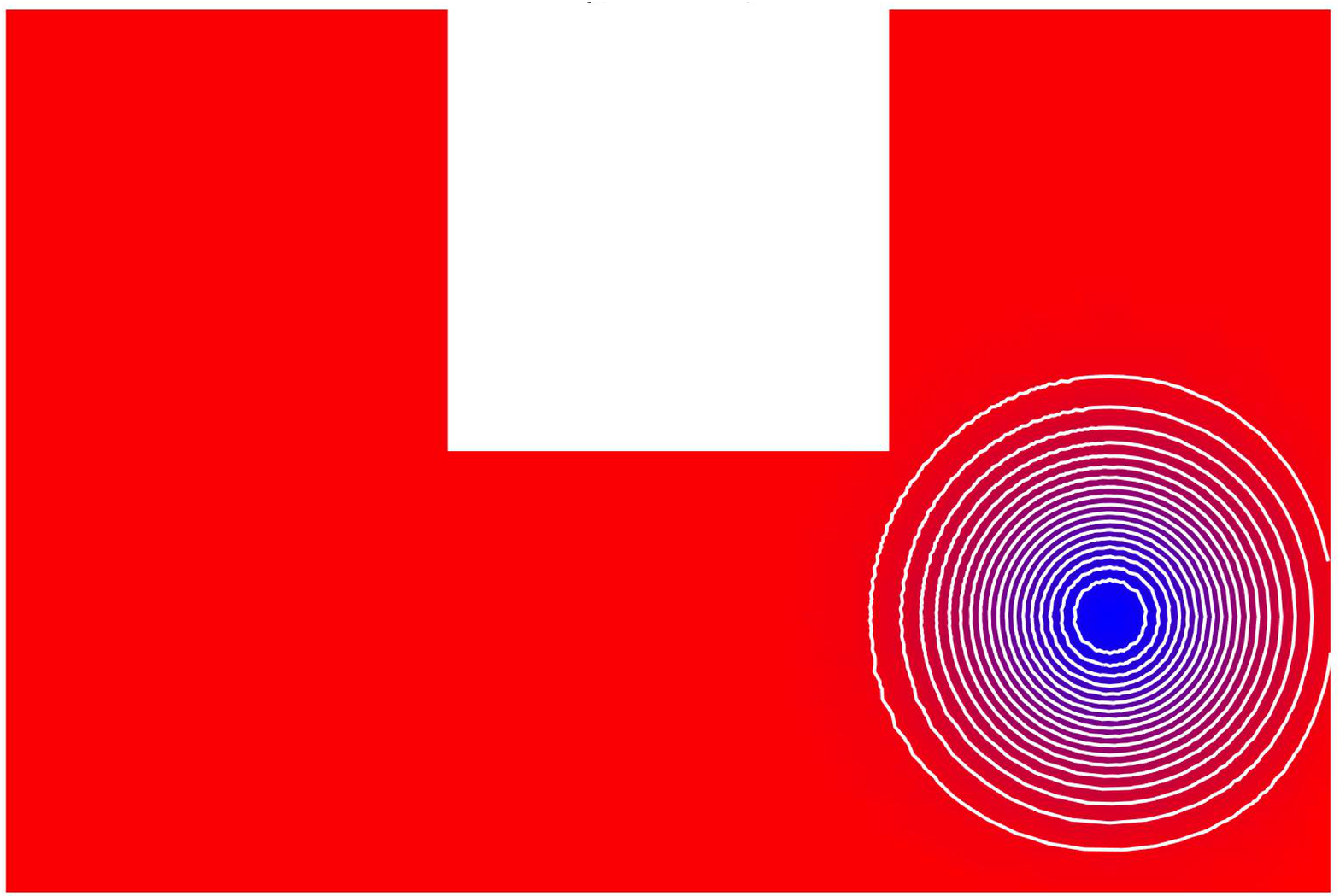}
&\includegraphics[height=35pt]{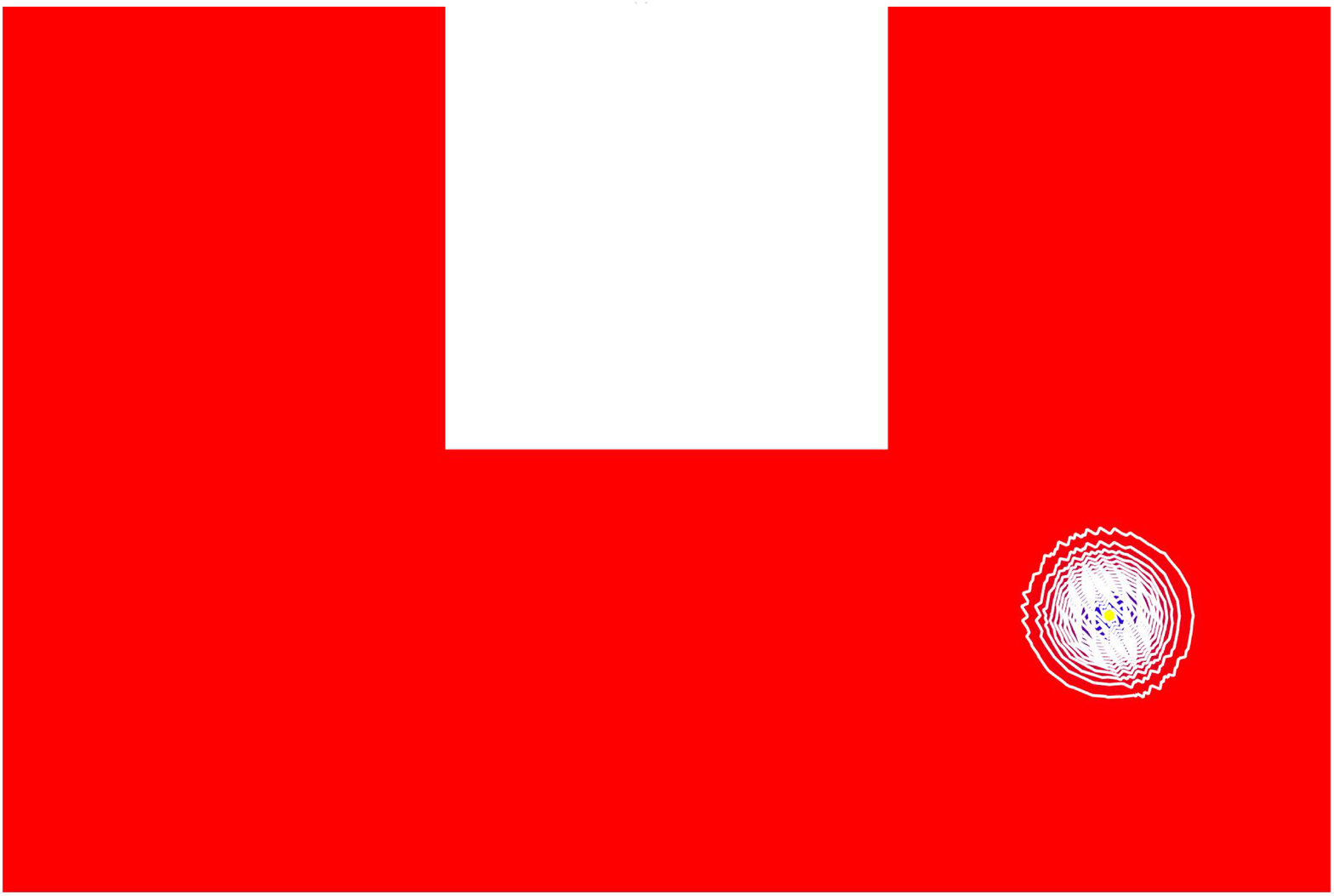}
&\includegraphics[height=35pt]{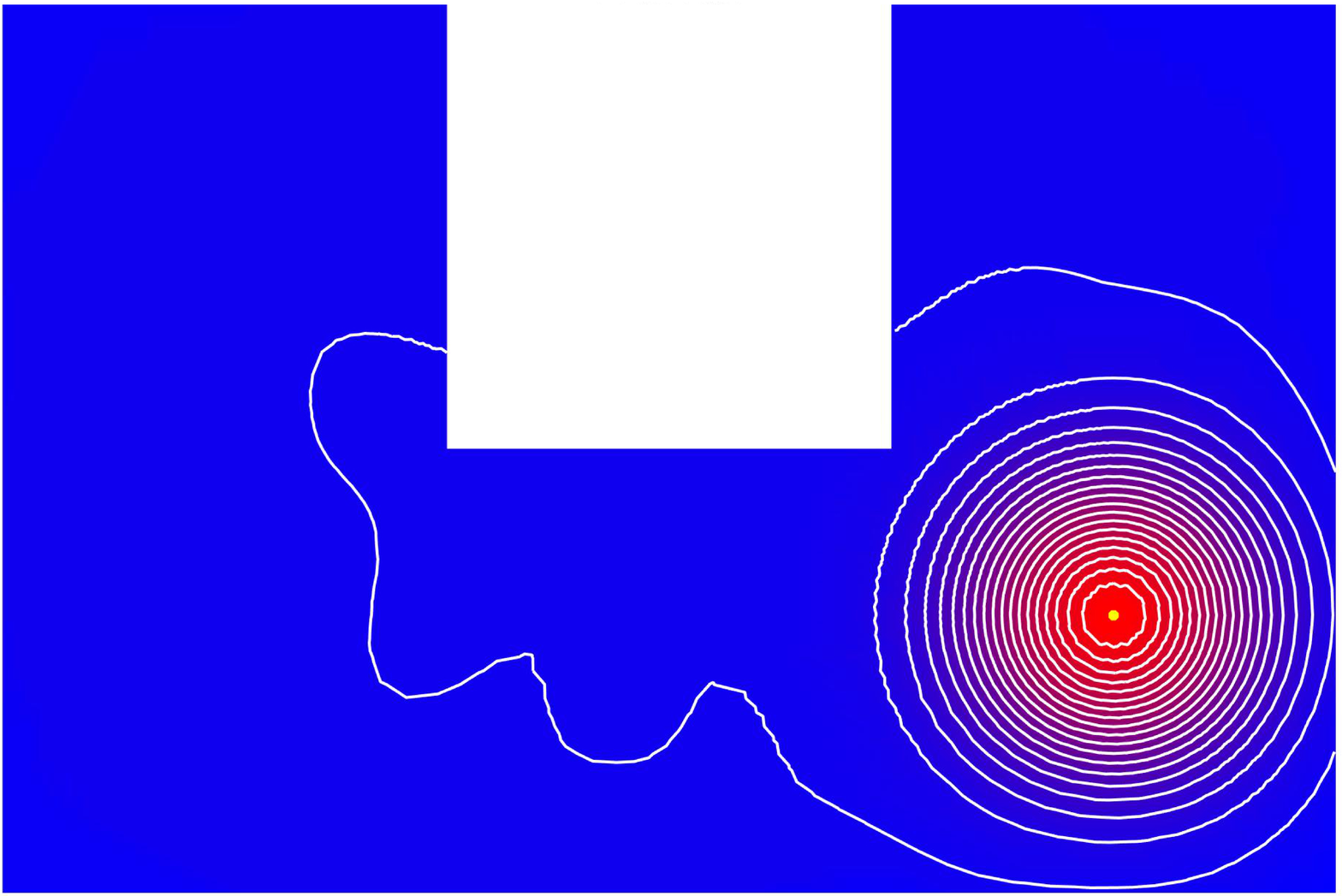}
&\includegraphics[height=35pt]{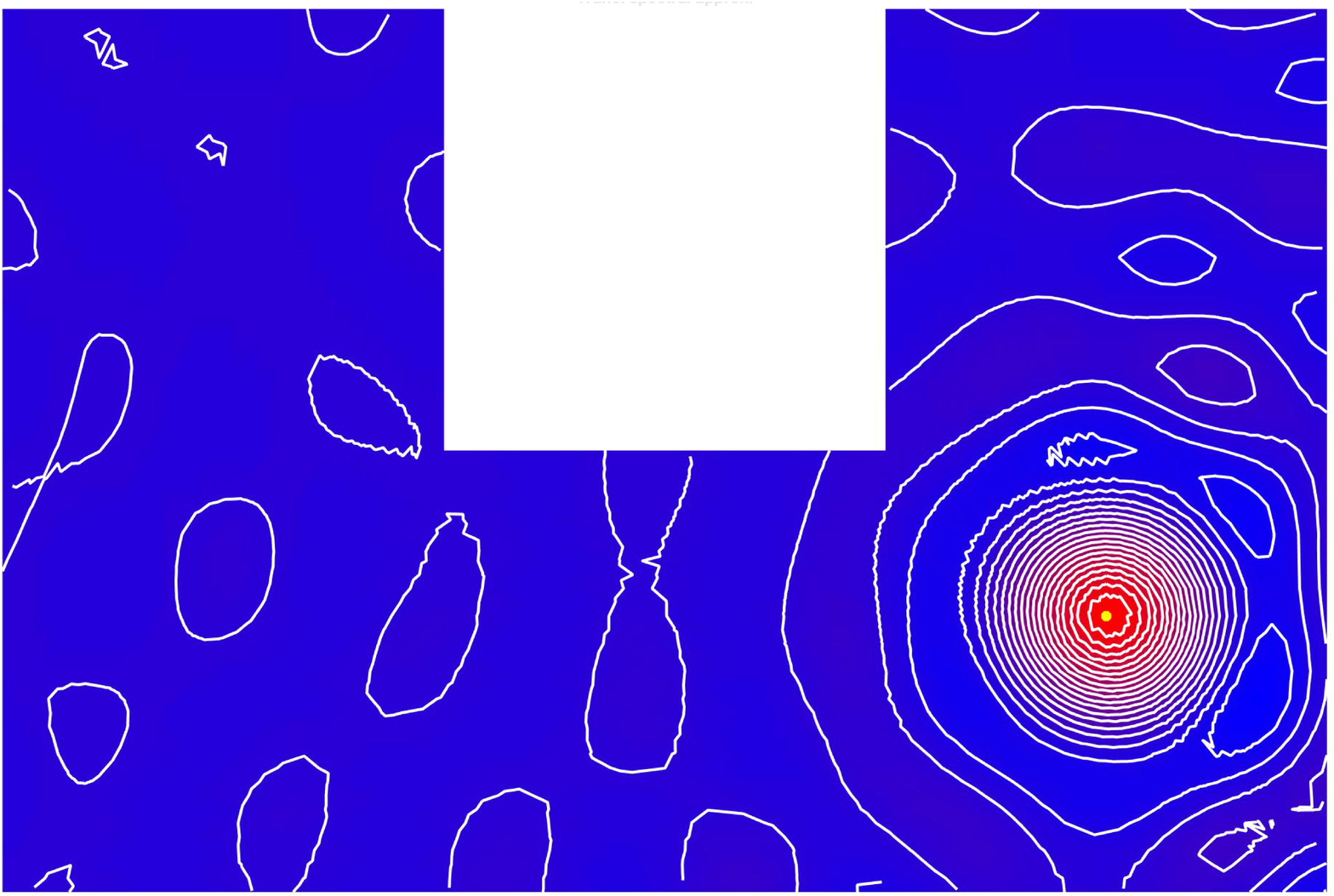} 
\end{tabular}
\caption{Colormap and level-sets of the diffusion wavelet at a source point computed with the P.-C. approximation and the trucated approximation with 100 eigenfunctions, which is affected by undulations far from the source point and at small scales.\label{fig:2D-ERROR-IPERBOLIC-HEAT}}
\end{figure}
\subsection{Polynomial filters: pseudo-spectrum and density\label{sec:POLY-APPROX}}
In spectral graph processing, the numerical stability of the spectrum and the evaluation of the characteristic polynomial of matrices associated with large graphs are two important aspects not addressed in detail by previous work. Indeed, we discuss the sensitivity of the computation of the spectrum with respect to the presence of multiple or close eigenvalues, the computation of the characteristic polynomials in case of multiple eigenvalues (Sect.~\ref{sec:EIG-SENSITIVENESS}), and the definition of the pseudo-spectrum as a way to compute a stable subset of the eigenvalues (Sect.~\ref{sec:LAPLACIAN-STABILITY}). Then, we address the approximation of the characteristic polynomial of a large matrix through spectral densities, which reduces to the computation of the trace of Chebyshev polynomial matrices. This approximation of the characteristic polynomial is necessary to apply the Cayley-Hamilton theorem for the reduction of the degree of polynomial filters (Sect.~\ref{sec:APPROX-SPECRTAL-DENSITY}).

\subsubsection{Characteristic polynomials: multiple eigenvalues\label{sec:EIG-SENSITIVENESS}}
One of the main issues with spectral graph processing is the presence of multiple or close Laplacian eigenvalues, which makes the computation of the spectrum and the evaluation of the spectral filters numerically unstable. Even though these situations are quite common in applications (e.g., for symmetric graph), previous work has paid a little attention to this problem that we address by discussing the computation of the characteristic polynomial in case of multiple eigenvalues. 

Given an arbitrary square matrix \mbox{$\mathbf{A}\in\mathbb{R}^{n\times n}$} (e.g., the Laplacian matrix, a kernel matrix \mbox{$\varphi(\tilde{\mathbf{L}})$}), or a kernel matrix), we consider its characteristic polynomial \mbox{$P(s):=det(\mathbf{A}-s\mathbf{I})$}. Indicating the eigenvalues of~$\mathbf{A}$ as \mbox{$(\lambda_{i})_{i=1}^{n}$}, previous work computes the coefficient of the characteristic polynomial \mbox{$P(s)=\sum_{i=0}^{n}\alpha_{i}s^{i}$} by imposing the interpolating conditions \mbox{$P(\lambda_{i})=0$}, \mbox{$\forall i$}, i.e., solving the homogeneous linear system \mbox{$\mathbf{V}\alpha=\mathbf{0}$}, where~$\mathbf{V}$ is the Vandermonde matrix and \mbox{$\alpha:=(\alpha_{i})_{i=1}^{n}$} is the unknown vector. In case of close or repeated eigenvalues, the Vandermonde coefficient matrix is singular, as it has multiple identical rows. In this case, for each repeated eigenvalue~$\lambda_{i}$ of multiplicity~$m$ the first \mbox{$(m-1)$} derivatives of the characteristic polynomial vanishes at~$\lambda_{i}$. Combining the interpolating conditions \mbox{$f(\lambda_{i})=R(\lambda_{i})$} at all the eigenvalues with the constraints \mbox{$\partial_{s^{k}}^{k}\varphi(\lambda_{i})=\partial_{s^{k}}^{k}R(\lambda_{i})$}, \mbox{$k=1,\ldots,m-1$, }on the derivatives at multiple eigenvalues, we get a non-singular linear system whose solution uniquely determines the coefficients of the remainder polynomial.
\begin{figure}
\centering
\includegraphics[height=65pt]{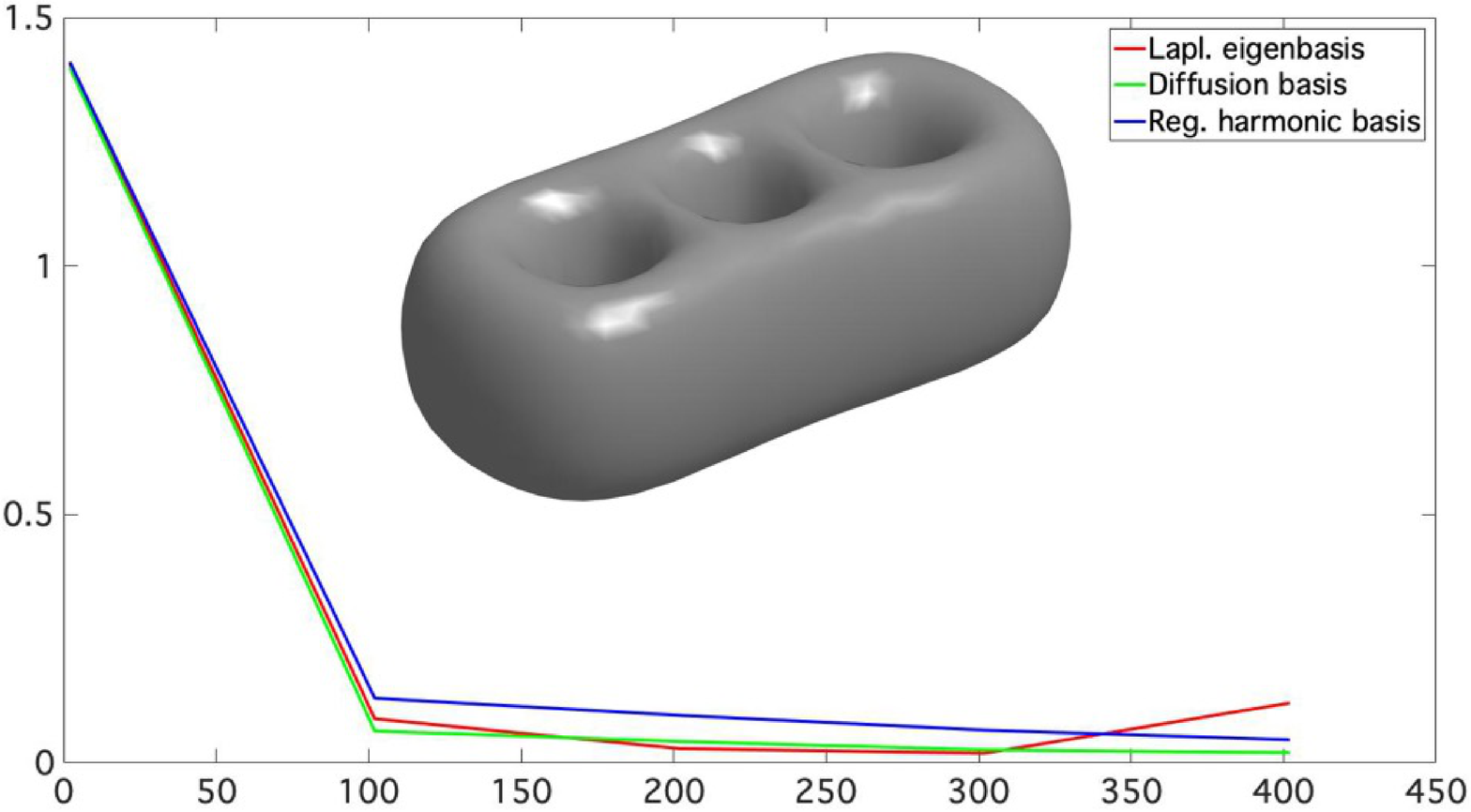}\\
\begin{tabular}{cc}
\includegraphics[height=65pt]{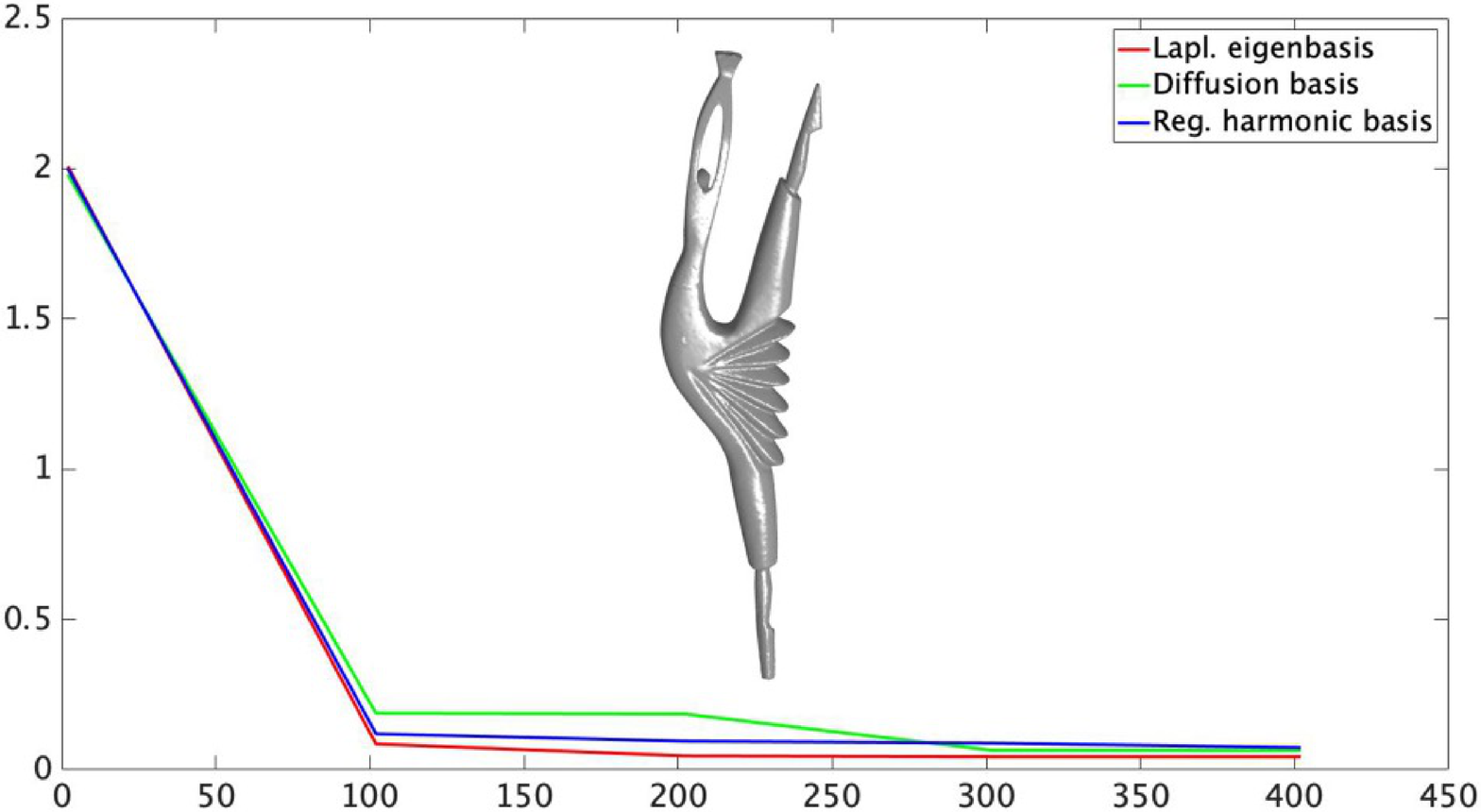}
&\includegraphics[height=65pt]{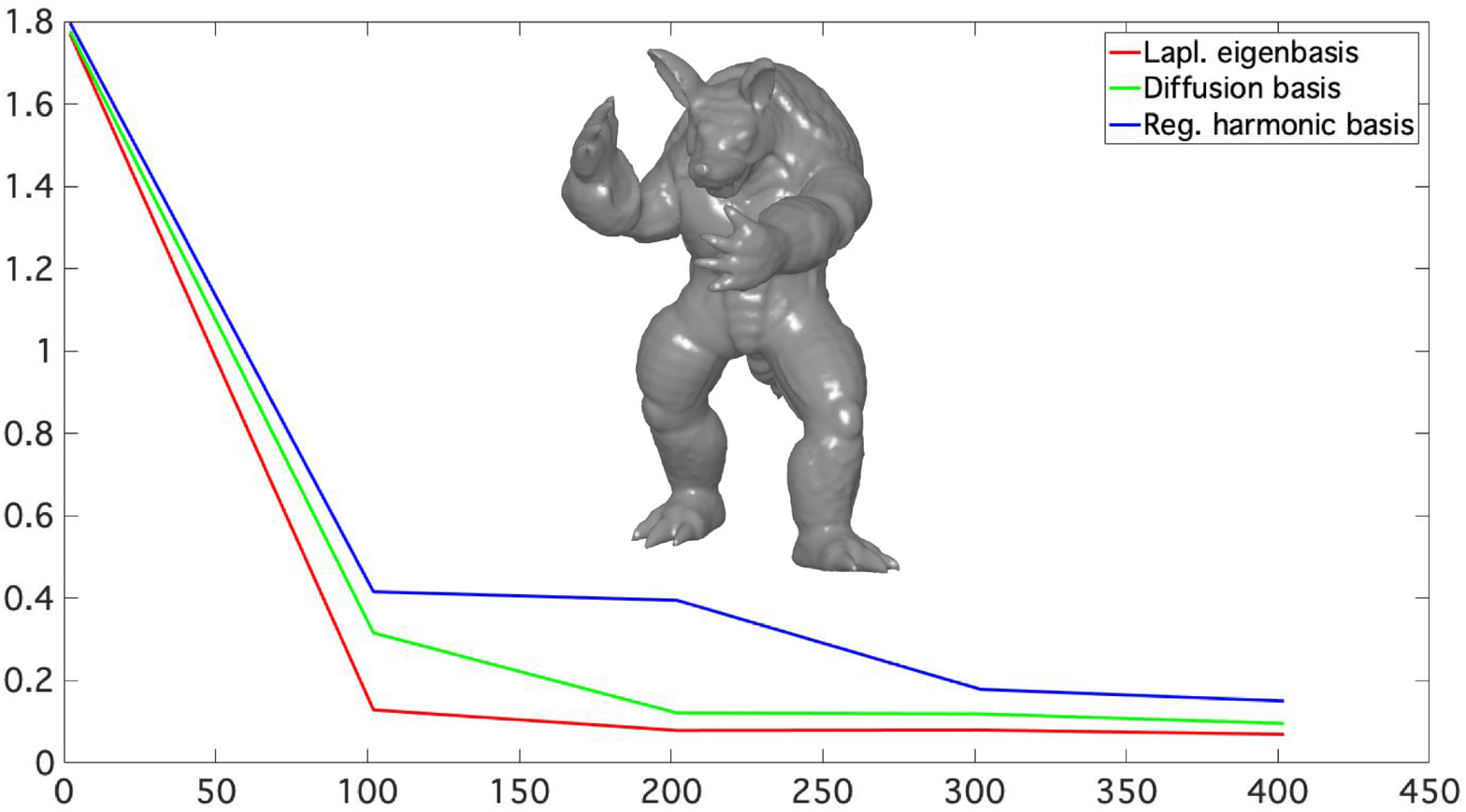}\\
\includegraphics[height=65pt]{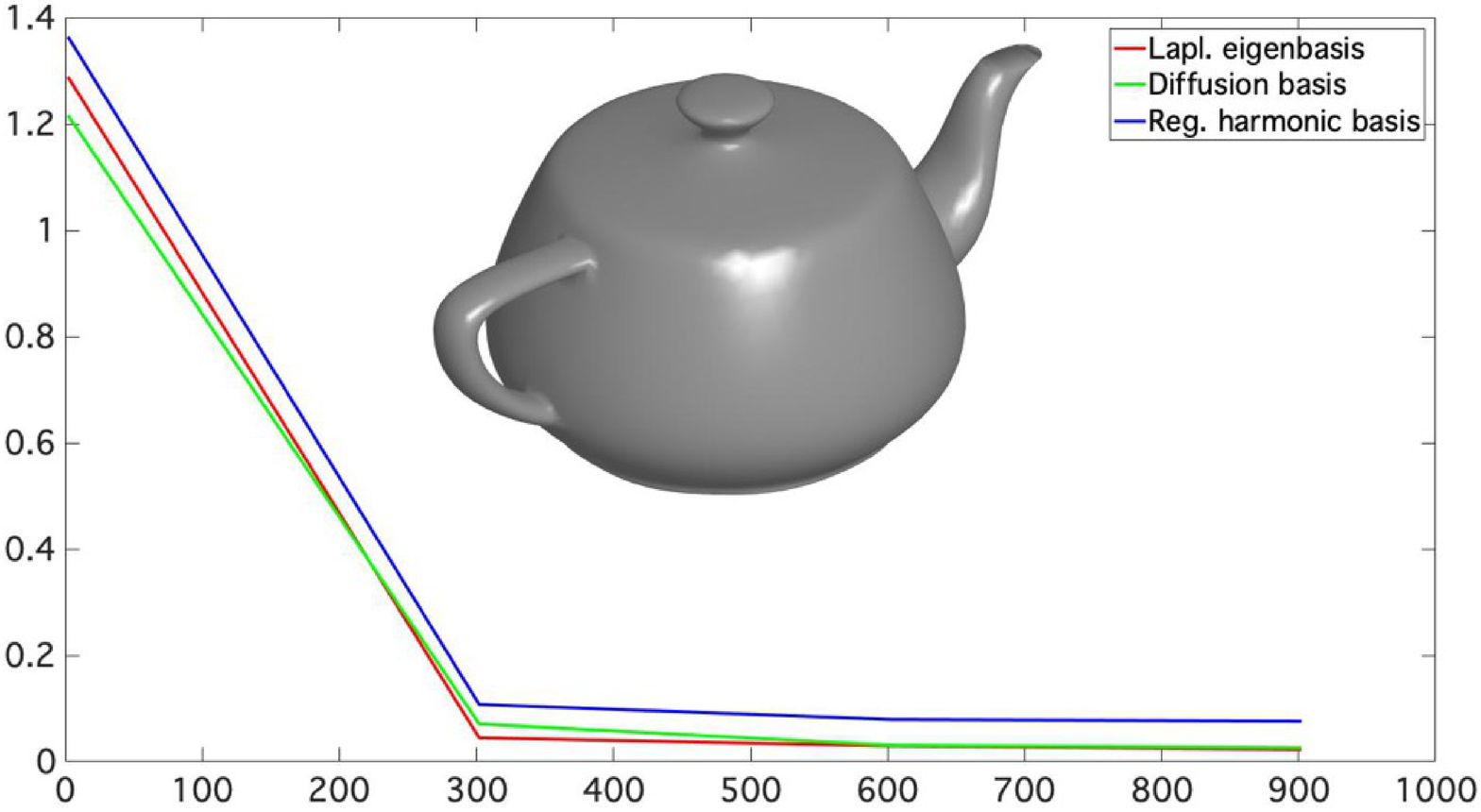}
&\includegraphics[height=65pt]{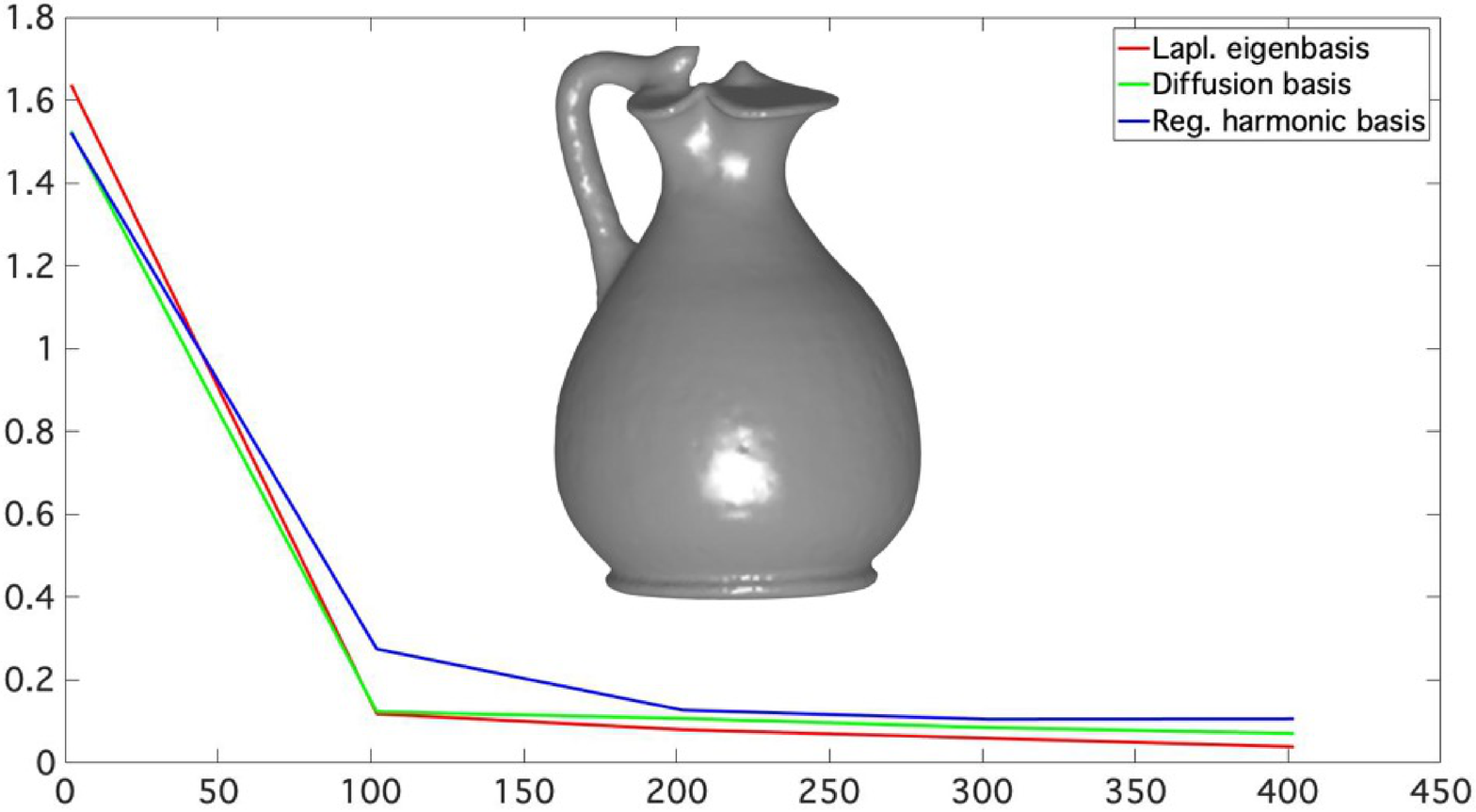}
\end{tabular}
\caption{$\ell_{\infty}$-approximation error ($y$-axis) of the reconstruction of the geometry of 3D shapes with respect to an increasing number ($x$-axis) of Laplacian eigenfunctions (red curve), harmonic (blue curve), and diffusion (green curve) kernels. Indeed, harmonic and diffusion kernels are a valid alternative to the eigenfunctions, with additional properties; in fact, they can be centred at any seed point and diffusion kernels have a multi-scale local behaviour.\label{fig:COMPRESSION}}
\end{figure}
\subsubsection{Stability of the eigenpairs and pseudo-spectrum\label{sec:LAPLACIAN-STABILITY}}
Assuming that \mbox{$\mu:=\rho(\lambda)$} is an eigenvalue of~$\varphi(\tilde{\mathbf{L}})$ with multiplicity~$m$ and rewriting the characteristic polynomial of \mbox{$\varphi(\tilde{\mathbf{L}})$} as \mbox{$P(s)=(s-\mu)^{m}Q(s)$}, where \mbox{$Q(\cdot)$} is a polynomial of degree \mbox{$(n-m)$} and \mbox{$q(\mu)\neq 0$}, we get that 
\begin{equation*}
(s-\mu)^{m}=\frac{P(s)}{Q(s)}
\approx \frac{O(\delta)}{Q(s)},\,
\delta\rightarrow 0,\,
\textrm{as }
P(s)\rightarrow 0,\,
s\rightarrow\mu,
\end{equation*}
i.e., \mbox{$s\approx\mu+O(\delta^{\frac{1}{m}})$}. Indeed, modifying the spectrum (i.e., the underlying graph) in such a way that the filtered eigenvalues are perturbed by \mbox{$\delta:=10^{-m}$} corresponds to a change of order~$0.1$ in \mbox{$\mu$} (i.e., \mbox{$s\approx\mu+0.1$}) and this amplification becomes larger as the multiplicity of the eigenvalue increases.

\textbf{Pseudo-spectrum}
To handle numerical instabilities associated with multiple or close eigenvalues (e.g., for symmetric graphs) of the Laplacian or spectral kernel matrix, we introduce the \emph{pseudo-spectrum}, which defines the eigenvalues of a matrix with respect to a threshold. Let~$\mathbf{A}$ be a \mbox{$n\times n$} symmetric matrix (or, more generally, a~$\mathbf{D}$-adjoint matrix as \mbox{$\varphi(\tilde{\mathbf{L}})$}) and \mbox{$\lambda_{\epsilon}(\mathbf{A}):=\{z\in\mathbb{R}:\,\|(z\mathbf{I}-\mathbf{A})^{-1}\|_{2}\geq 1/\epsilon\}$} the \emph{pseudo-spectrum} of~$\mathbf{A}$, i.e., the set of disks of radius~$\epsilon$ centred around the eigenvalues of~$\mathbf{A}$. Indeed,~$z$ is a pseudo-eigenvalues of~$\mathbf{A}$ if and only if \mbox{$(z\mathbf{I}-\mathbf{A})$} is sufficiently close to be singular (i.e., with respect to the threshold~$\epsilon$). In fact, 
\begin{equation*}\label{eq:SPECTRAL-RELATION}
\frac{1}{\epsilon}
\leq \|(z\mathbf{I}-\mathbf{A})^{-1}\|_{2}
=\vert\lambda_{\max}((z\mathbf{I}-\mathbf{A})^{-1})\vert
=\vert\lambda_{\min}(z\mathbf{I}-\mathbf{A})\vert^{-1},
\end{equation*}
i.e., \mbox{$\lambda_{\min}(z\mathbf{I}-\mathbf{A})\leq\epsilon$}. The pseudo-spectrum of~$\mathbf{A}$ generalises the spectrum (i.e., \mbox{$\lambda_{0}(\mathbf{A})=\lambda(\mathbf{A})$}) and \mbox{$z\in\lambda_{\epsilon}(\mathbf{A})$} if and only if \mbox{$z\in\lambda(\mathbf{A}+\mathbf{E})$} and \mbox{$\|\mathbf{E}\|_{2}\leq\epsilon$}. If \mbox{$\epsilon_{1}\geq\epsilon_{2}$}, then \mbox{$\lambda_{\epsilon_{1}}(\mathbf{A})\subseteq \lambda_{\epsilon_{2}}(\mathbf{A})$}.
\begin{figure}[t]
\centering
\begin{tabular}{ccc}
(a) $(f,g)$ &(b)~$\tilde{f}$ &(c)\\
\includegraphics[height=50pt]{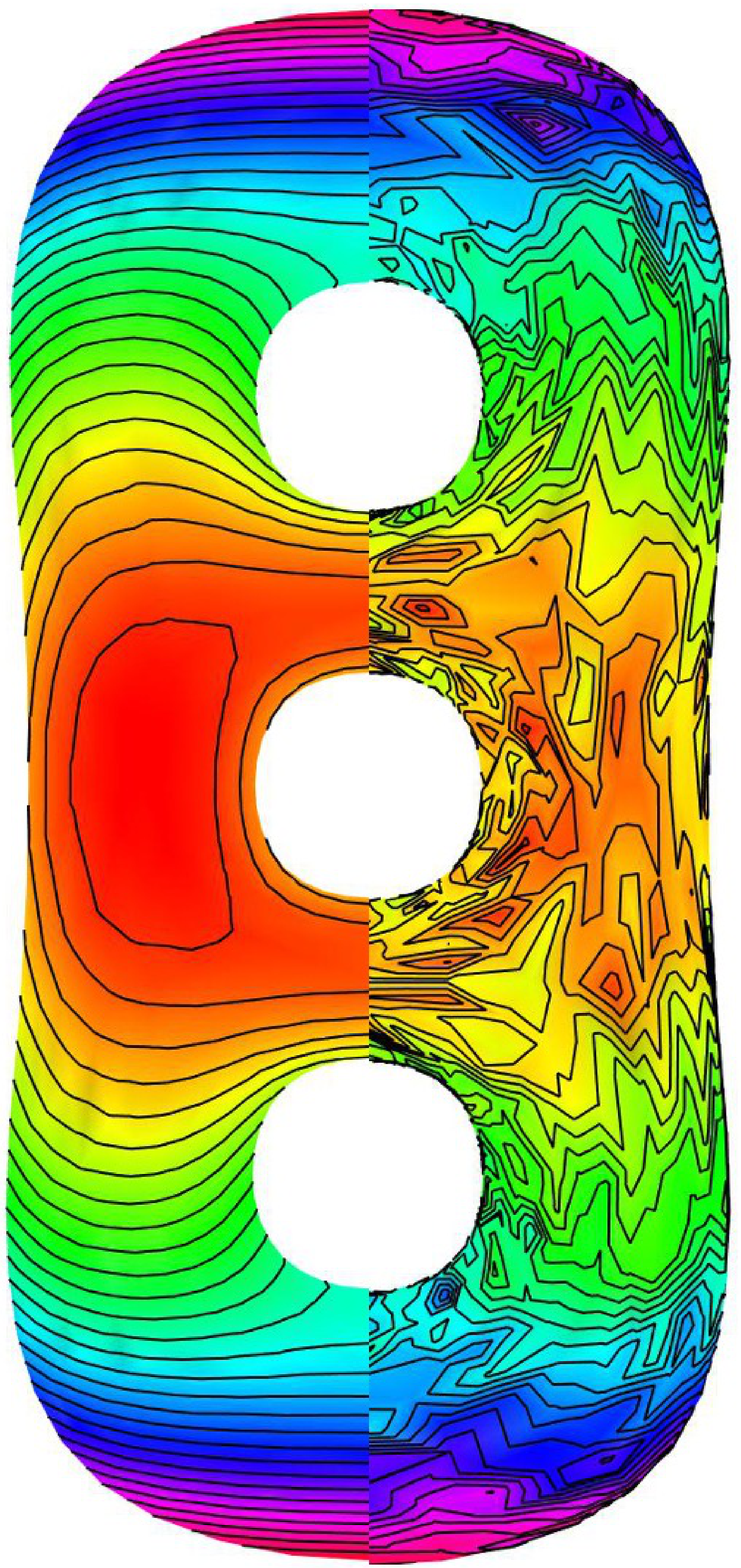}
&\includegraphics[height=50pt]{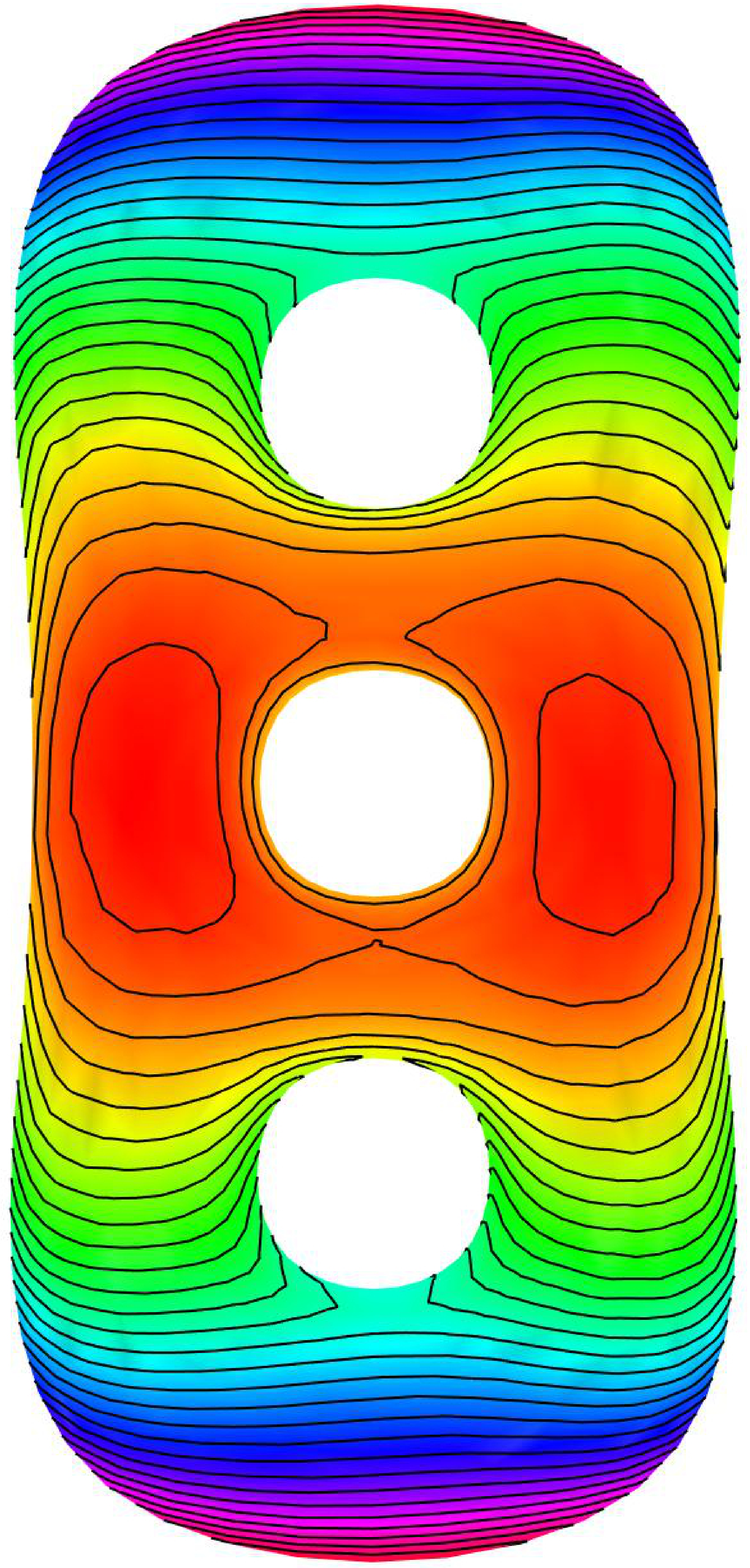}
&\includegraphics[height=60pt]{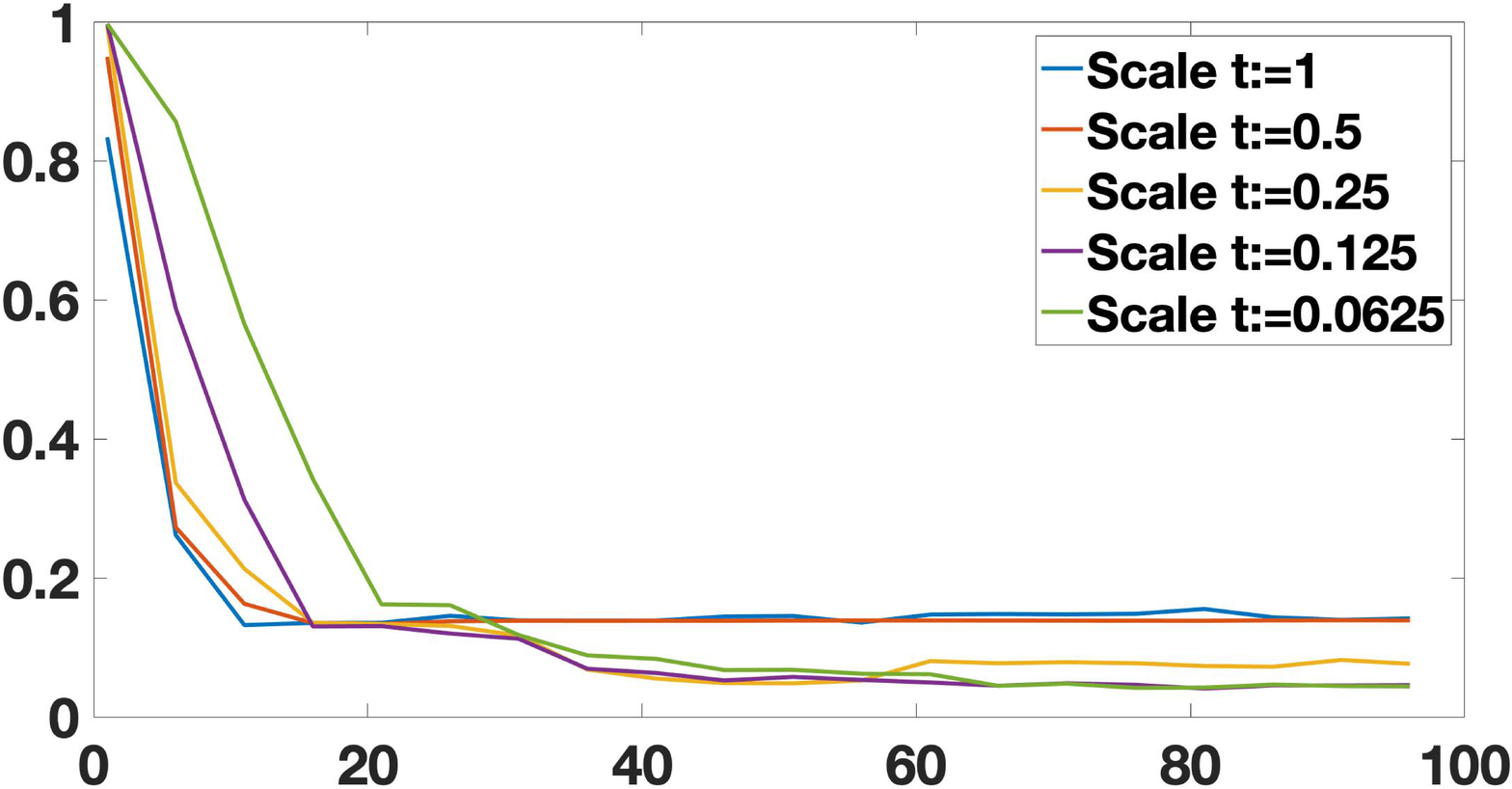}\\
\includegraphics[height=50pt]{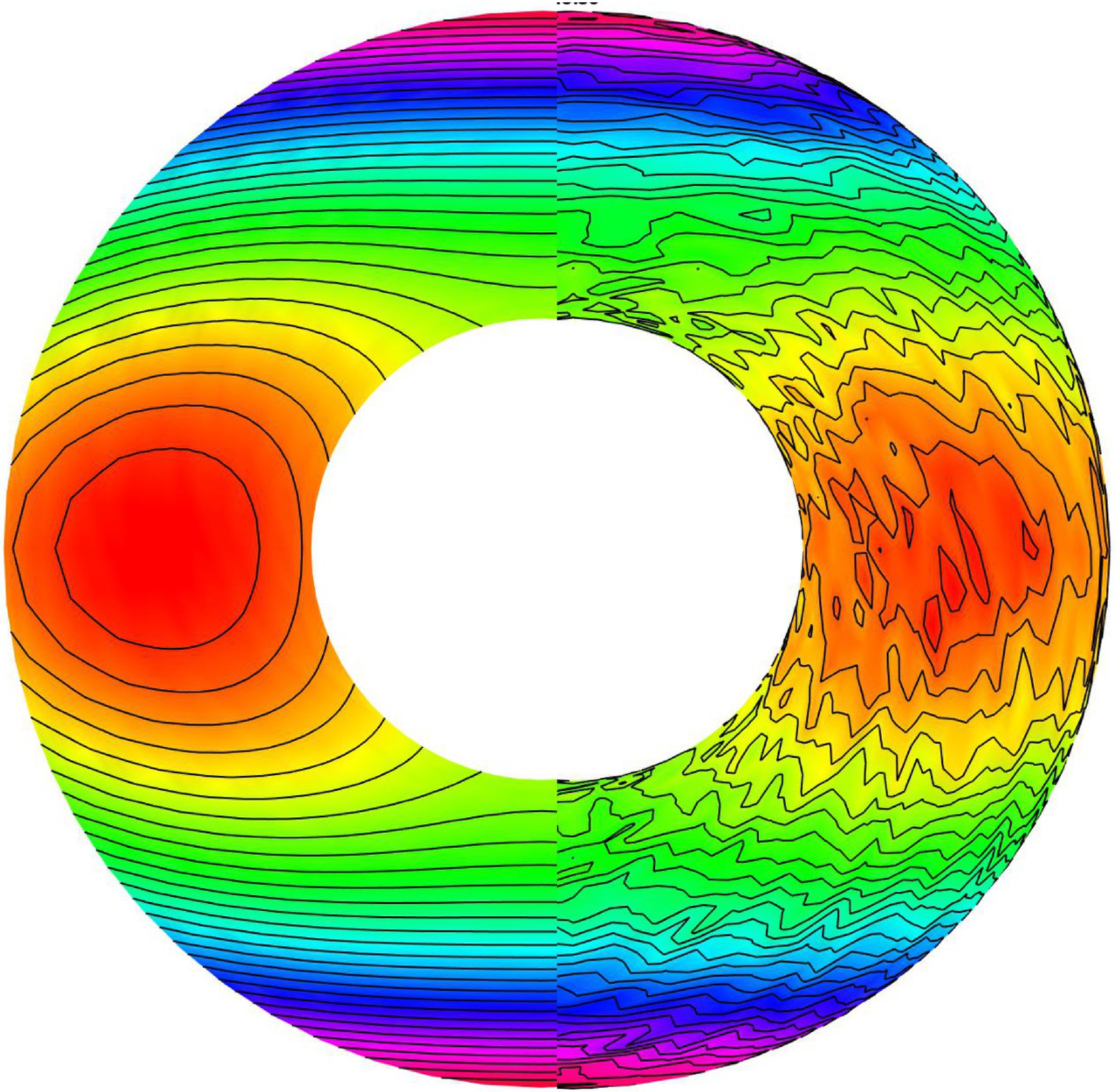}
&\includegraphics[height=50pt]{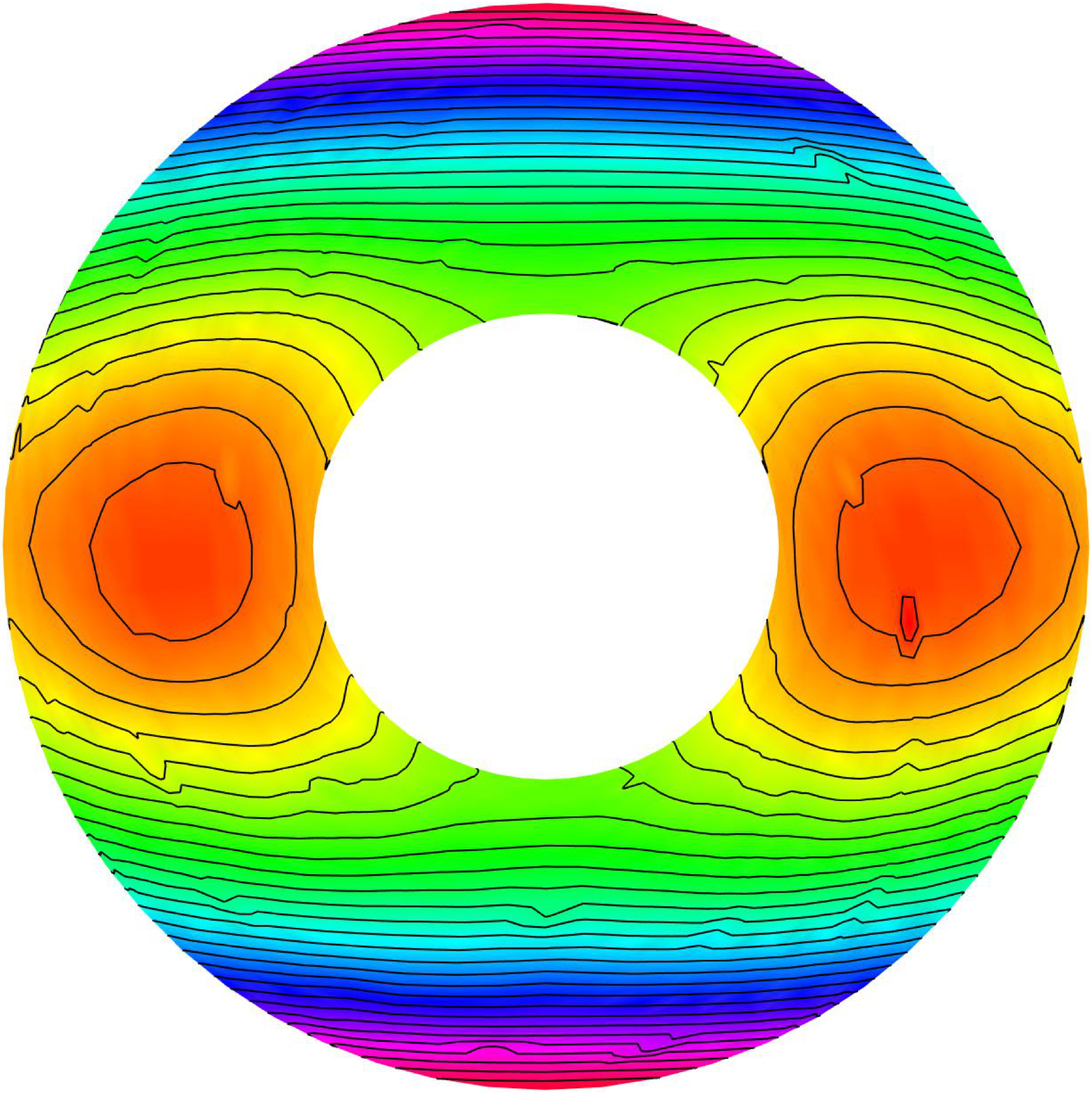}
&\includegraphics[height=60pt]{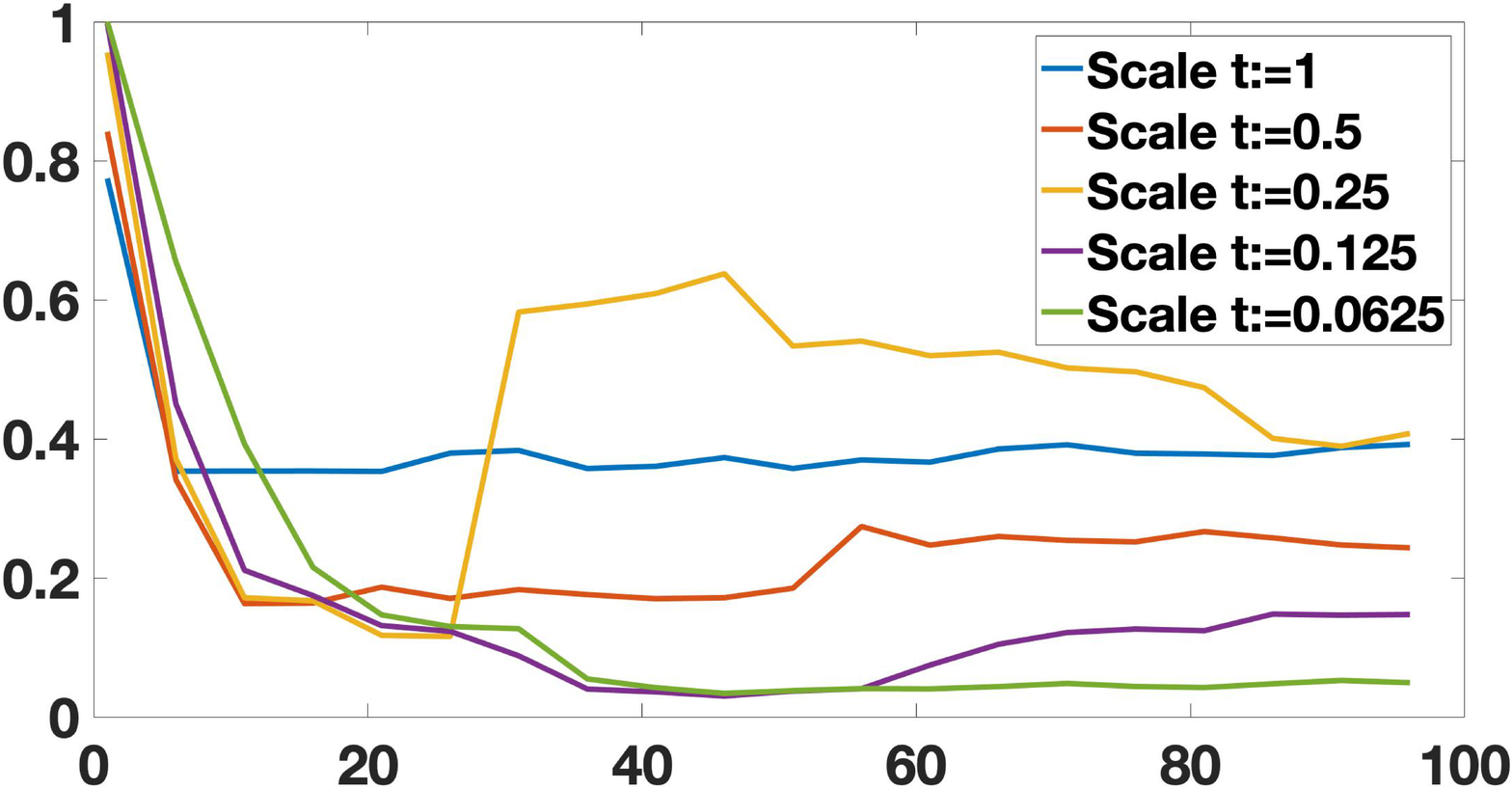}\\
\includegraphics[height=50pt]{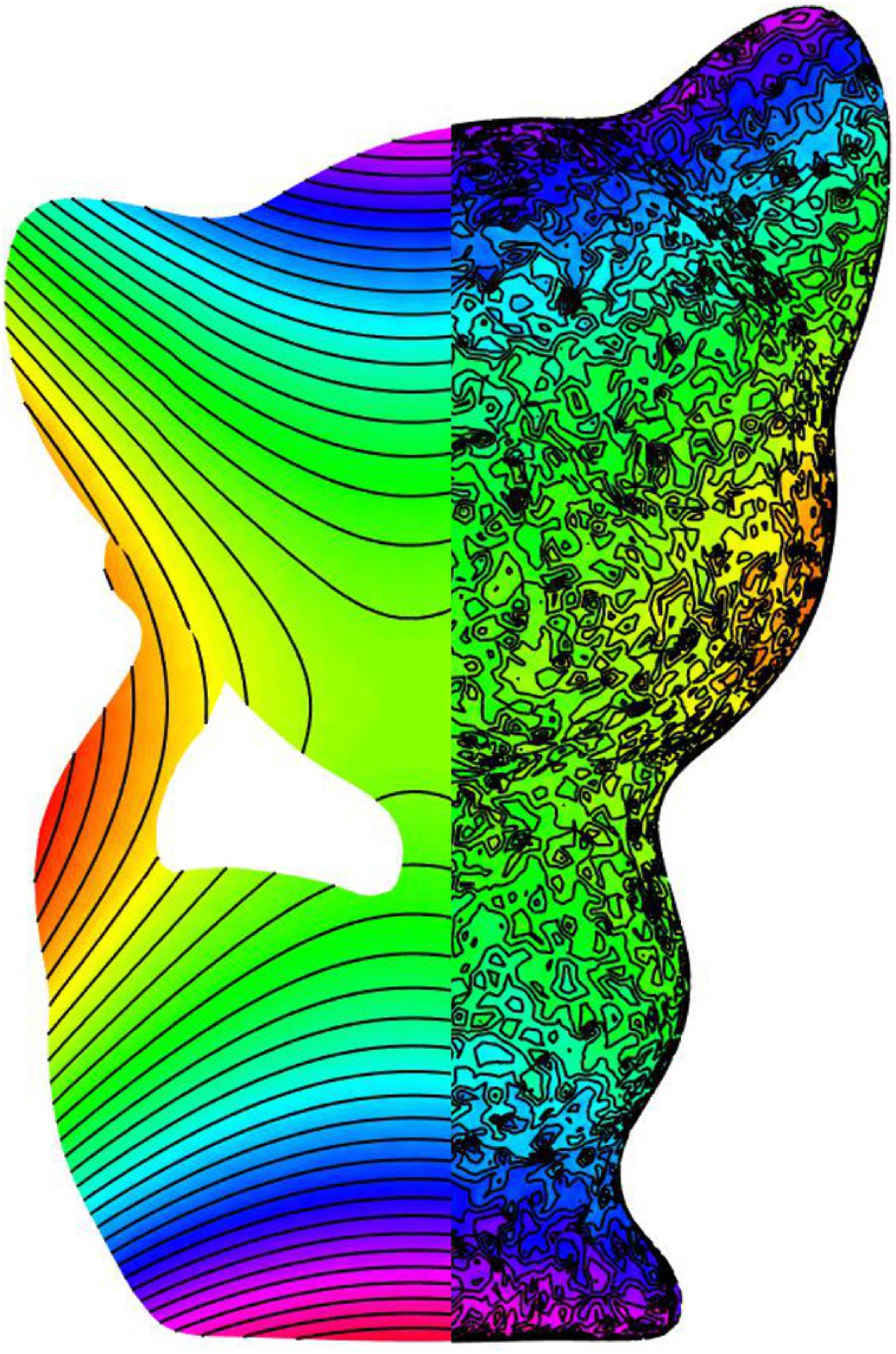}
&\includegraphics[height=50pt]{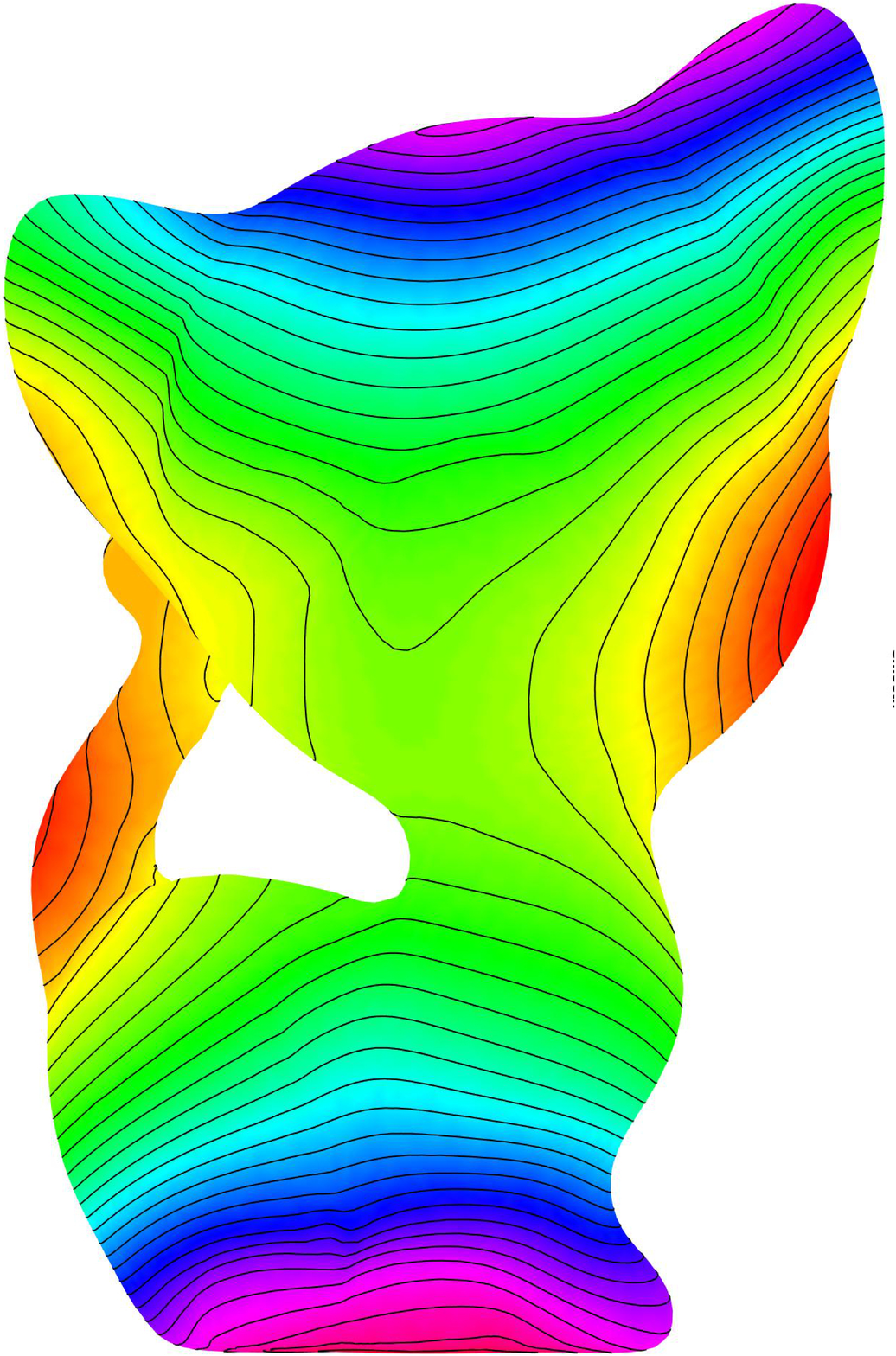}
&\includegraphics[height=60pt]{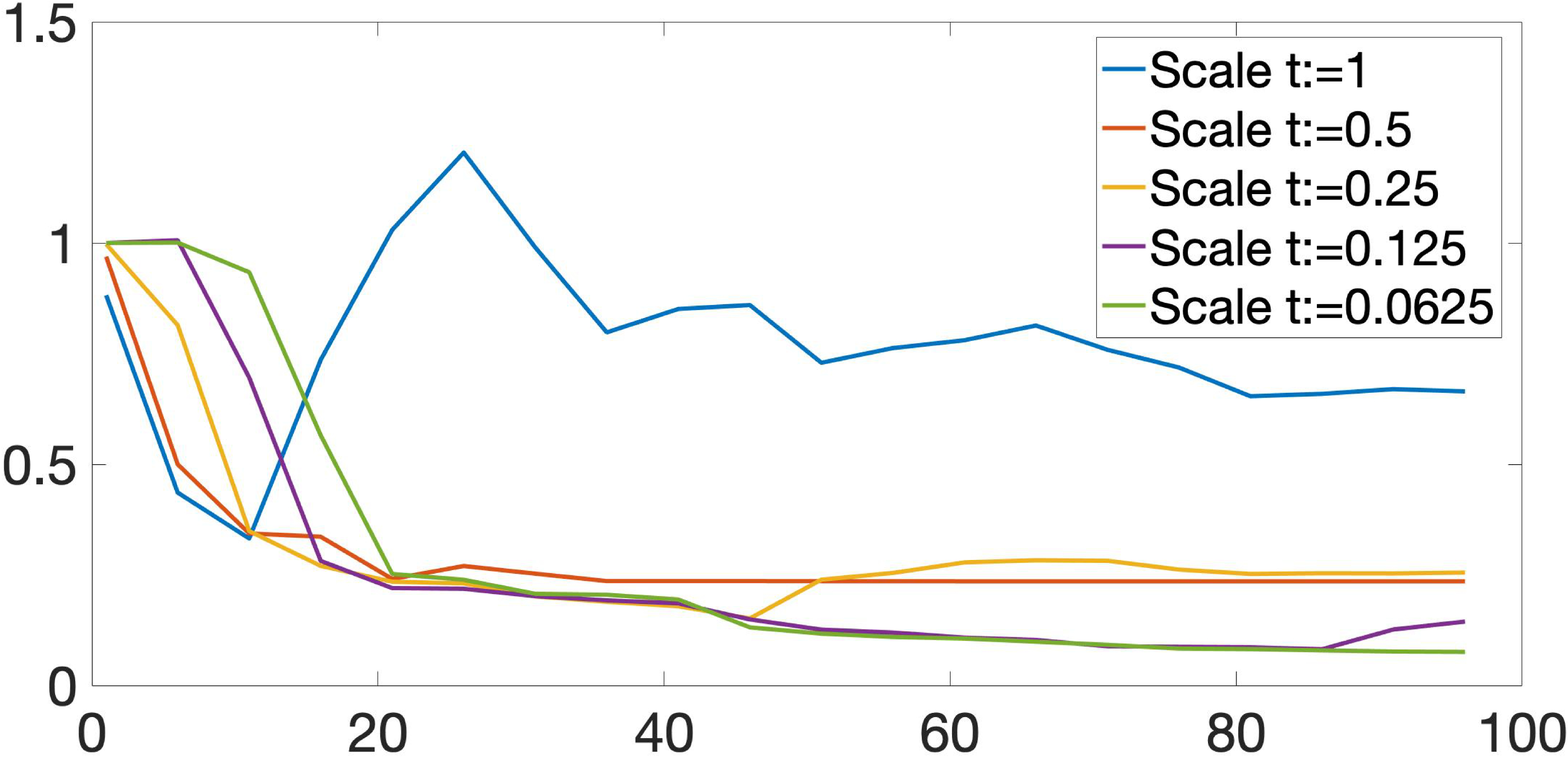}
\end{tabular}
\caption{Level-sets and colour-map of (a,right) a noisy function \mbox{$g:=f+\delta$} achieved by adding a Gaussian noise~$\delta$ to the function~$f$ in (a,left); (b) smoothed scalar function $\tilde{f}$ achieved with 100 diffusion functions. (c) Approximation error ($y$-axis) and number of diffusion functions ($x$-axis) at 5 different scales.\label{fig:3TORUS-DIFFUSION-BASIS}}
\end{figure}
\subsubsection{Approximated spectral densities\label{sec:APPROX-SPECRTAL-DENSITY}}
Given a matrix~$\mathbf{A}$ with eigenvalues \mbox{$(\lambda_{i})_{i=1}^{n}$}, its \emph{pointwise} and \emph{smooth spectral densities}~\cite{LIN2016} in an interval \mbox{$[a,b]$} are defined as the functions
\begin{equation*}
\delta:=\frac{1}{n}\int_{a}^{b}\sum_{i=1}^{n}\delta(s-\lambda_{i})ds,\quad
\delta(t):=\frac{1}{n\sigma}\sum_{i=1}^{n}h_{\sigma}(t-\lambda_{i}).
\end{equation*}
The pointwise spectral density measures the percentage of eigenvalues belonging to the selected interval, and the smooth density is defined by replacing the~$\delta$-function with a smooth hat function, such as the Gaussian function \mbox{$h_{\sigma}(t):=(2\pi\sigma^{2})^{-1/2}\exp(-t^{2}/\sigma^{2})$}. Here, larger values of the width~$\sigma$ provide smoother curves while reducing the interpolation of the eigenvalues. Assuming that the input matrix has been normalised in such a way that its eigevalues belong to the interval \mbox{$[-1,1]$}, we express the normalised spectral density \mbox{$\tilde{\delta}(t):=(1-t^{2})^{1/2}\delta(t)$} in terms of the Chebyshev polynomials as \mbox{$\tilde{\delta}=\sum_{n=0}^{+\infty}\mu_{n}T_{n}$}. Then,
\begin{equation*}
\begin{split}
\mu_{k}
&=\langle\tilde{\delta},T_{k}\rangle_{w,2}
=\frac{1}{n}\sum_{i=1}^{n}\int_{-1}^{1}\delta(s-\lambda_{i})T_{k}(s)ds\\
&=\frac{1}{n}\sum_{i=1}^{n}T_{k}(\lambda_{i})
=\frac{1}{n}\textrm{Trace}(T_{k}(\mathbf{A})).
\end{split}
\end{equation*}
Indeed, each coefficient reduces to the evaluation of the trace of the matrix achieved by evaluating the corresponding Chebyshev polynomial on the input matrix. For more details on the approximation of the spectral densities of large matrices with polynomial methods and spectroscopic approaches, we refer the reader to~\cite{LITVINOV1993}.

\subsection{Numerical stability and computational cost\label{sec:COMPUTATIONAL-COST}}
\textbf{Rational spectrum-free computation}
According to Eq. (\ref{eq:RATIONAL-RECURSION}), the vector \mbox{$R_{n+1}(\tilde{\mathbf{L}})\mathbf{f}$} is recursively computed as
\begin{equation}\label{eq:DISCRETE-RECURSION}
\mathbf{f}_{n+1}
:=R_{n+1}(\tilde{\mathbf{L}})\mathbf{f}
=2(\tilde{\mathbf{L}}+\mathbf{I})^{-1}(\tilde{\mathbf{L}}-\mathbf{I})\mathbf{f}_{n}-\mathbf{f}_{n-1}
=2\mathbf{g}_{n}-\mathbf{f}_{n-1},
\end{equation}
where the vector~$\mathbf{g}_{n}$ is the solution to the sparse, symmetric, and positive definite linear system \mbox{$(\tilde{\mathbf{L}}+\mathbf{I})\mathbf{g}_{n}=(\tilde{\mathbf{L}}-\mathbf{I})\mathbf{f}_{n}$}, or equivalently \mbox{$(\mathbf{L}+\mathbf{D})\mathbf{g}_{n}=(\mathbf{L}-\mathbf{D})\mathbf{f}_{n}$}. Since the coefficient matrix \mbox{$(\mathbf{L}+\mathbf{D})$} is independent of the iteration and positive-definite, it can be pre-factorised and its pre-factorisation is used for the computation of~$\mathbf{a}_{n}$ in linear time at each iteration. In an analogous way, we derive the discrete counterparts of Eqs. (\ref{eq:CANONICAL-POL-BASIS}), (\ref{eq:CHEBYSHEV-BASIS}). In Fig.~\ref{fig:MONKEY-DIFFUSION}, the diffusion kernel has been computed through the spectrum-free approximation with Chebyshev rational polynomials and the recursive relation in Eq. (\ref{eq:DISCRETE-RECURSION}). The shape and distribution of the level-sets confirm the high accuracy of this approximation at small and large scales.

\textbf{Conditioning of the spectral wavelet/kernel}
If~$\varphi$ is an increasing function (i.e.,~$\varphi$ is a low pass filter), then the conditioning number of the spectral kernel is bounded as
\begin{equation*}
\kappa_{2}(\mathbf{K}_{\varphi})
=\kappa_{2}(\varphi(\tilde{\mathbf{L}}))=\frac{\max_{i=1,\ldots,n}\{\varphi(\lambda_{i})\}}{\min_{i=1,\ldots,n}\{\varphi(\lambda_{i})\}}=\frac{\|\varphi\|_{\infty}}{\varphi(0)},
\end{equation*}
and it is ill-conditioned when \mbox{$\varphi(0)$} is close to zero or~$\varphi$ is unbounded. If~$\rho$ is bounded and \mbox{$\varphi(0)$} is not too close to~$0$, then the spectral kernel is well-conditioned. If \mbox{$\varphi(0)$} is null, then we consider the smallest and not null filtered Laplacian eigenvalue at the denominator of the previous relation.

\textbf{Computational cost}
The computational cost of the truncated spectral approximation of the kernel/wavelet depends on the sparsity degree of the Laplacian matrix and takes from \mbox{$\mathcal{O}(kn\log n)$} to \mbox{$\mathcal{O}(kn^{2})$} time, where~$k$ is the number of selected eigenpairs. Choosing a rational approximation of the input filter of degree \mbox{$(r,l)$}, the evaluation of the spectral kernel/wavelet is reduced to~$r$ linear systems whose coefficient matrix is~$\mathbf{L}$. Through iterative solvers, the computational cost is \mbox{$\mathcal{O}(r\tau(n))$}, where \mbox{$\tau(n)$} is the cost for the solution of a sparse linear system, which varies from \mbox{$\mathcal{O}(n)$} to \mbox{$\mathcal{O}(n^{2})$}, according to the sparsity of the coefficient matrix, and it is \mbox{$\mathcal{O}(n\log n)$} in the average case. Indeed, the polynomial and rational polynomial approximations have the same order of computational complexity.

Evaluating the spectral kernel/wavelet at one scale with the rational approximation is generally more efficient and accurate than the truncated spectral approximation, especially at small scales (Fig.~\ref{fig:2D-ERROR-IPERBOLIC-HEAT}). In the average case, the cost is \mbox{$\mathcal{O}(rn\log n)$} versus \mbox{$\mathcal{O}(kn\log n)$}, with \mbox{$r<<k$}, e.g., \mbox{$r:=5,7$} and \mbox{$k=50,100$}. In case of multiple scales, the eigensystem is computed only once and applied for the evaluation of spectral wavelets at all scales in linear time, while the linear systems associated with the rational approximation is solved for each scale. Assuming~$s$ scales, the computational cost of the rational approximation is competitive with respect to the truncated spectral approximation if \mbox{$sr<k$}, i.e., the number of scales is lower than the ration~$k/r$ between the number~$k$ of selected eigenpairs and the degree~$r$ of the rational polynomial. 
\begin{figure}
\centering
\begin{tabular}{c|c|c}
Mean corresp. err. &Diff. fun. &Lapl. fun.\\
\hline
\includegraphics[height=50pt]{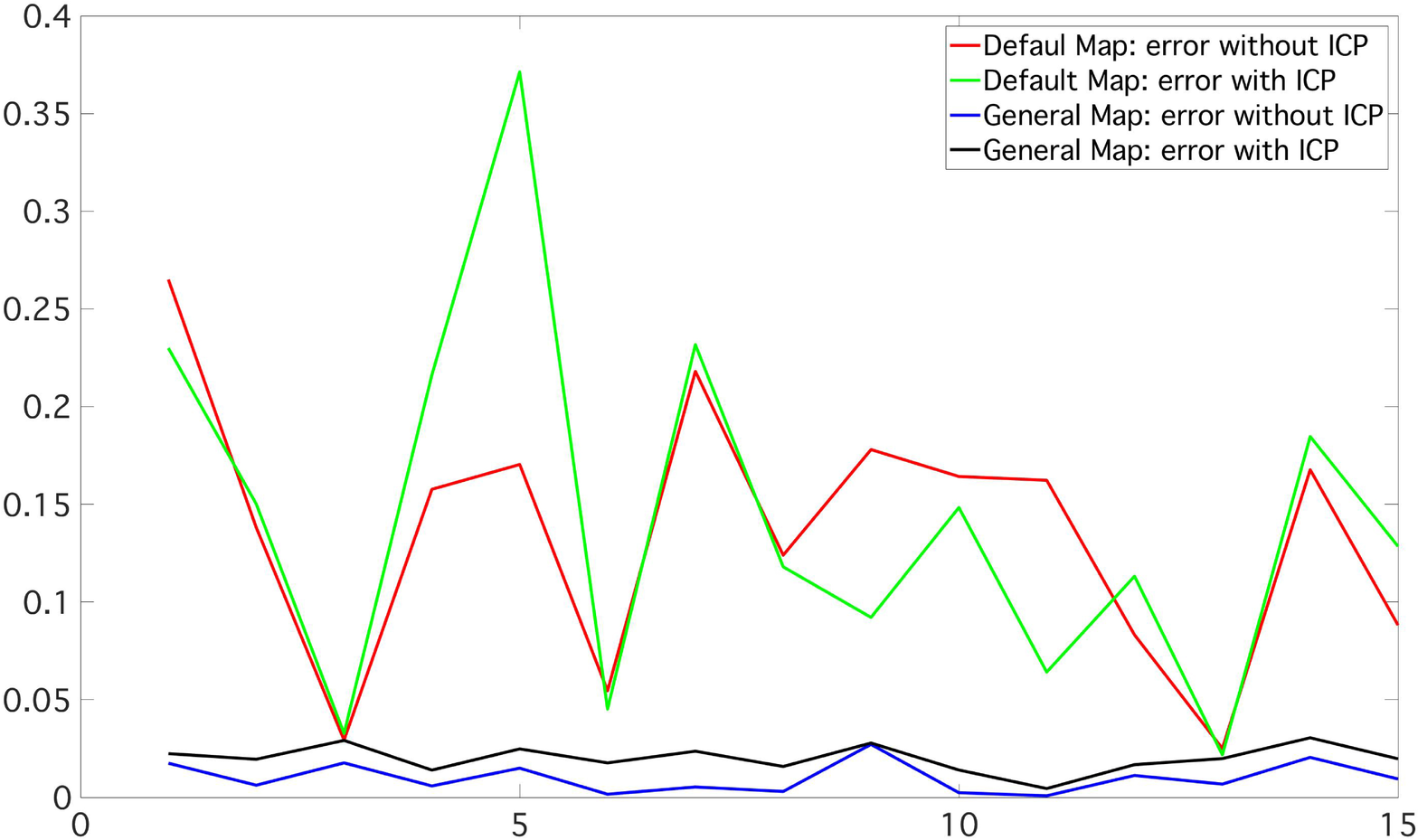}
&\includegraphics[height=40pt]{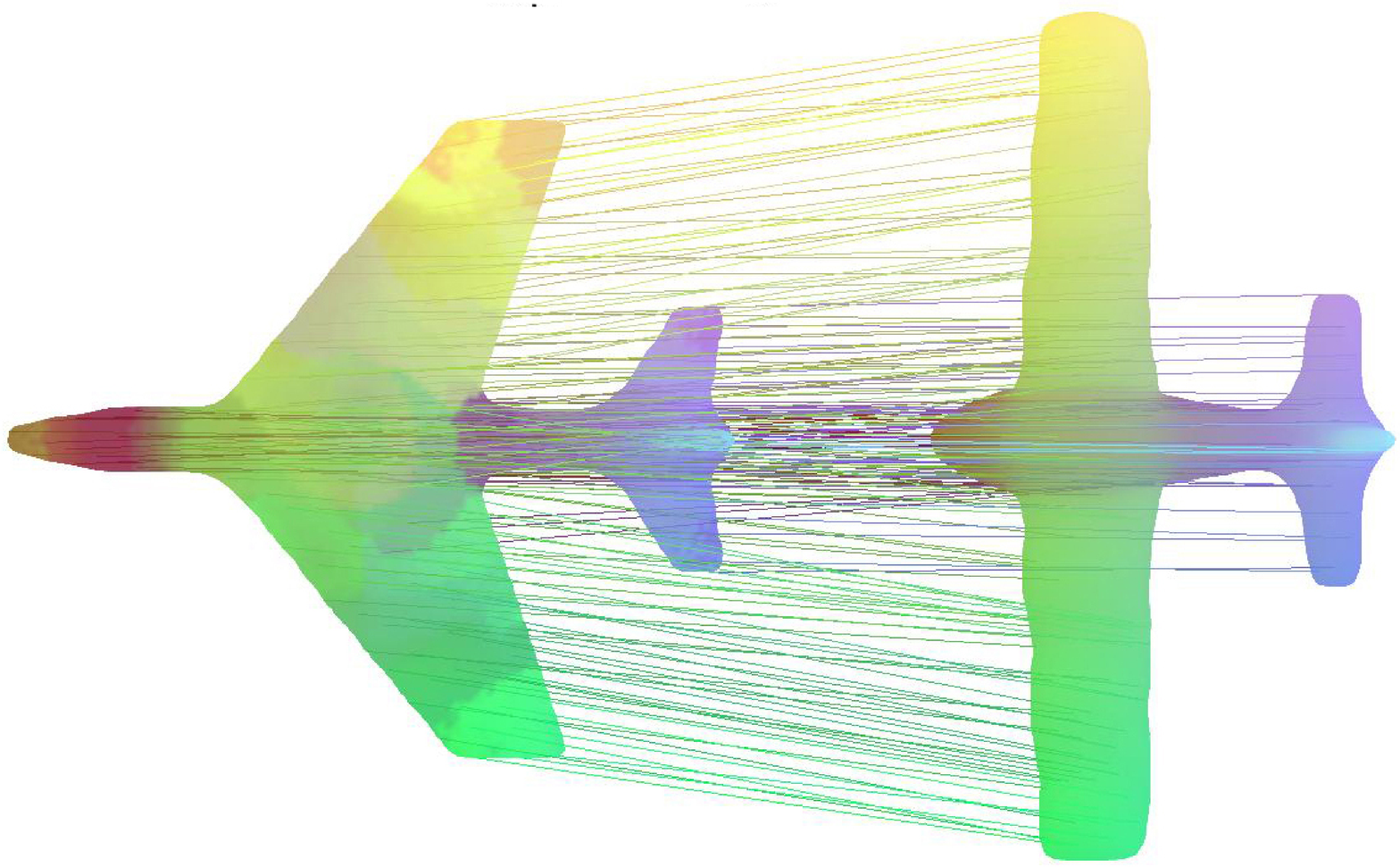}
&\includegraphics[height=40pt]{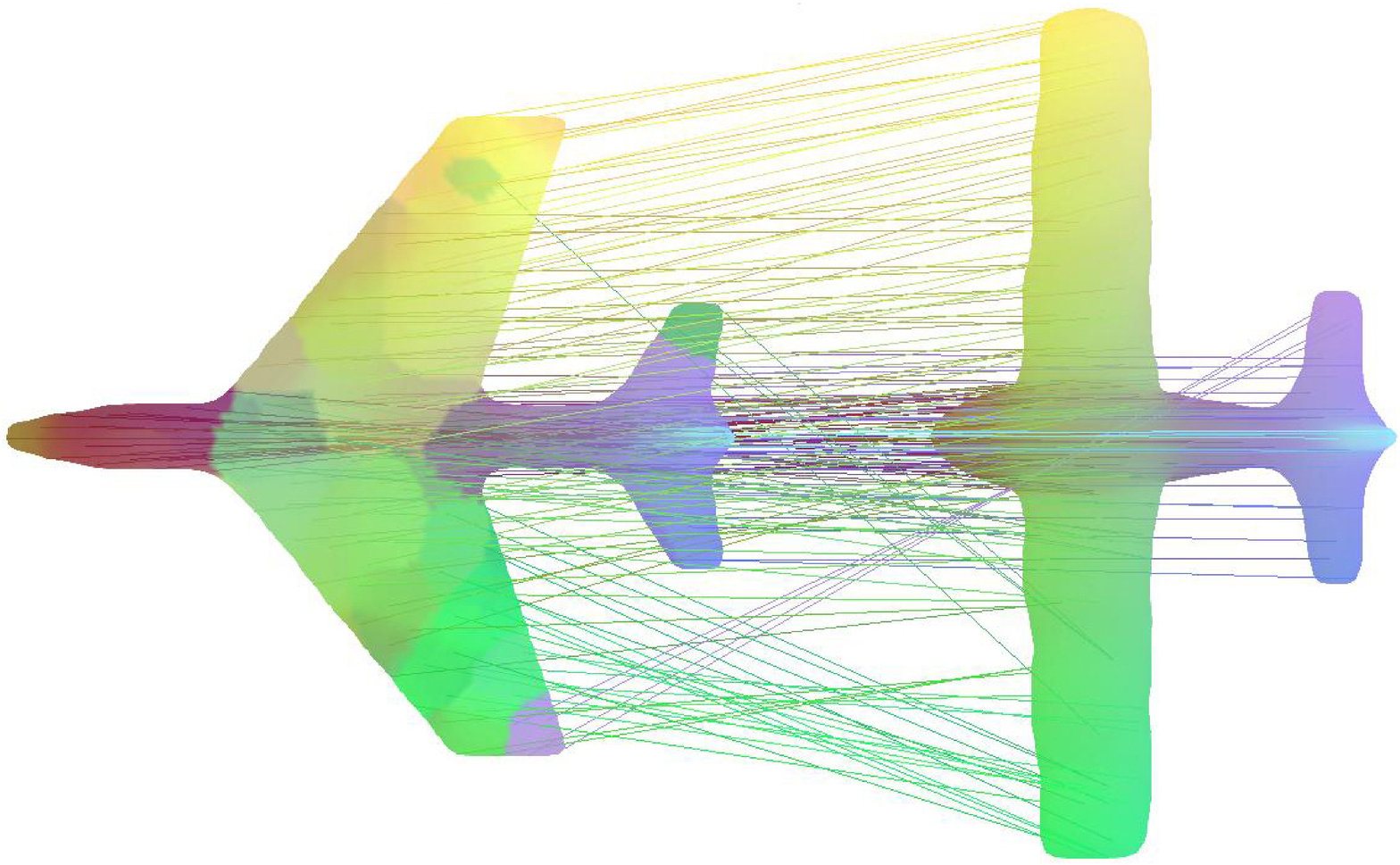}\\
\includegraphics[height=50pt]{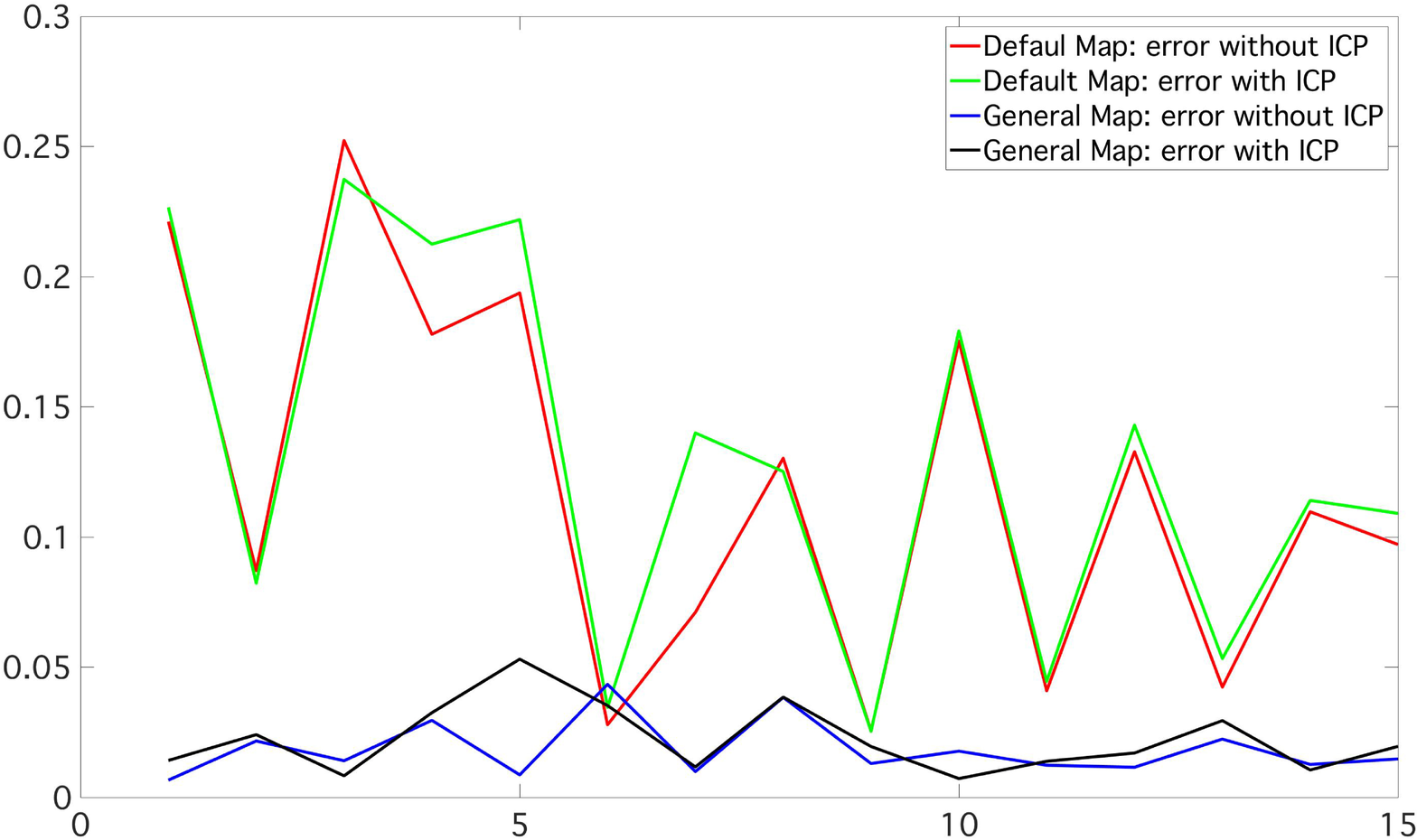}
&\includegraphics[height=40pt]{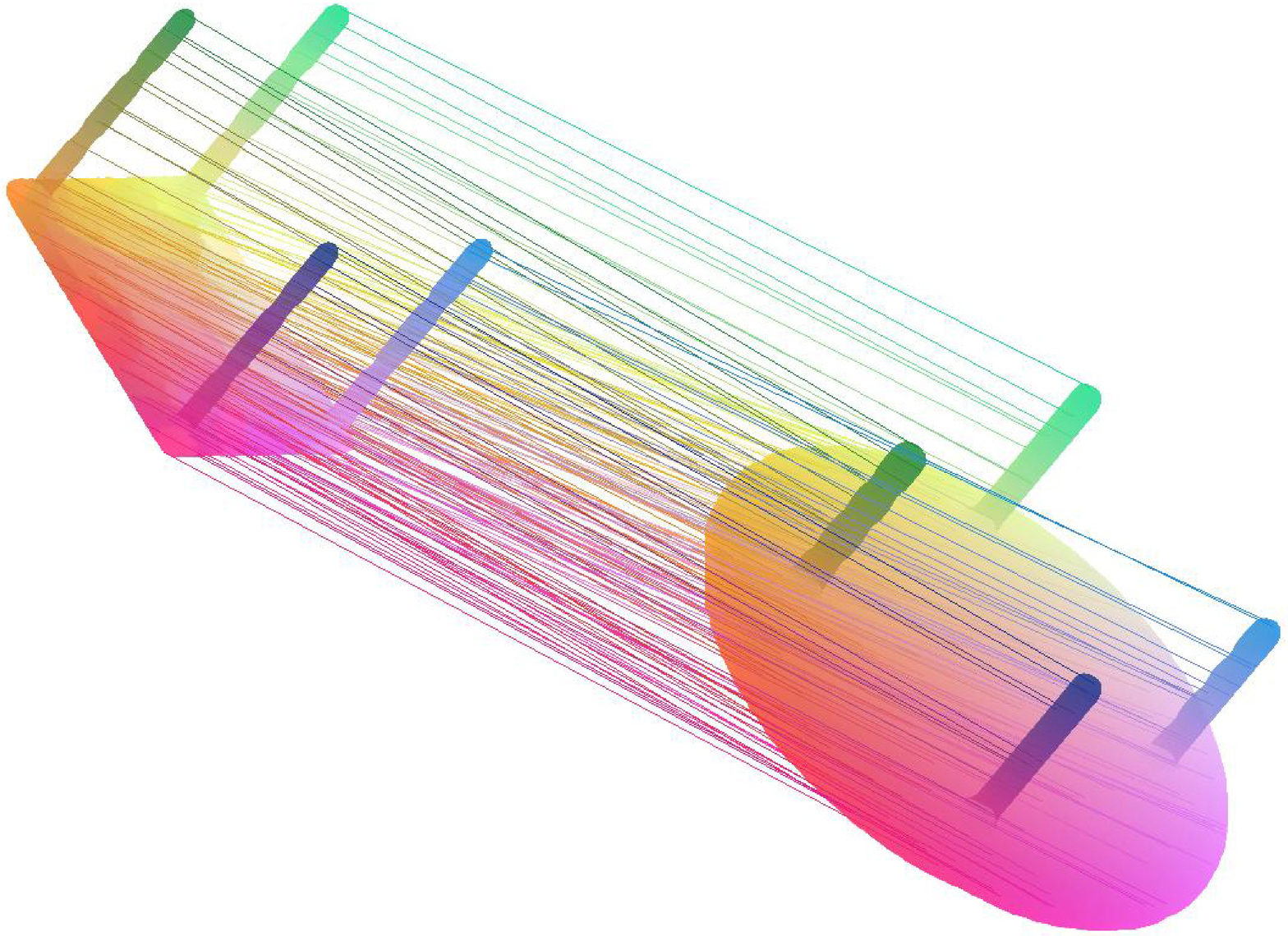}
&\includegraphics[height=40pt]{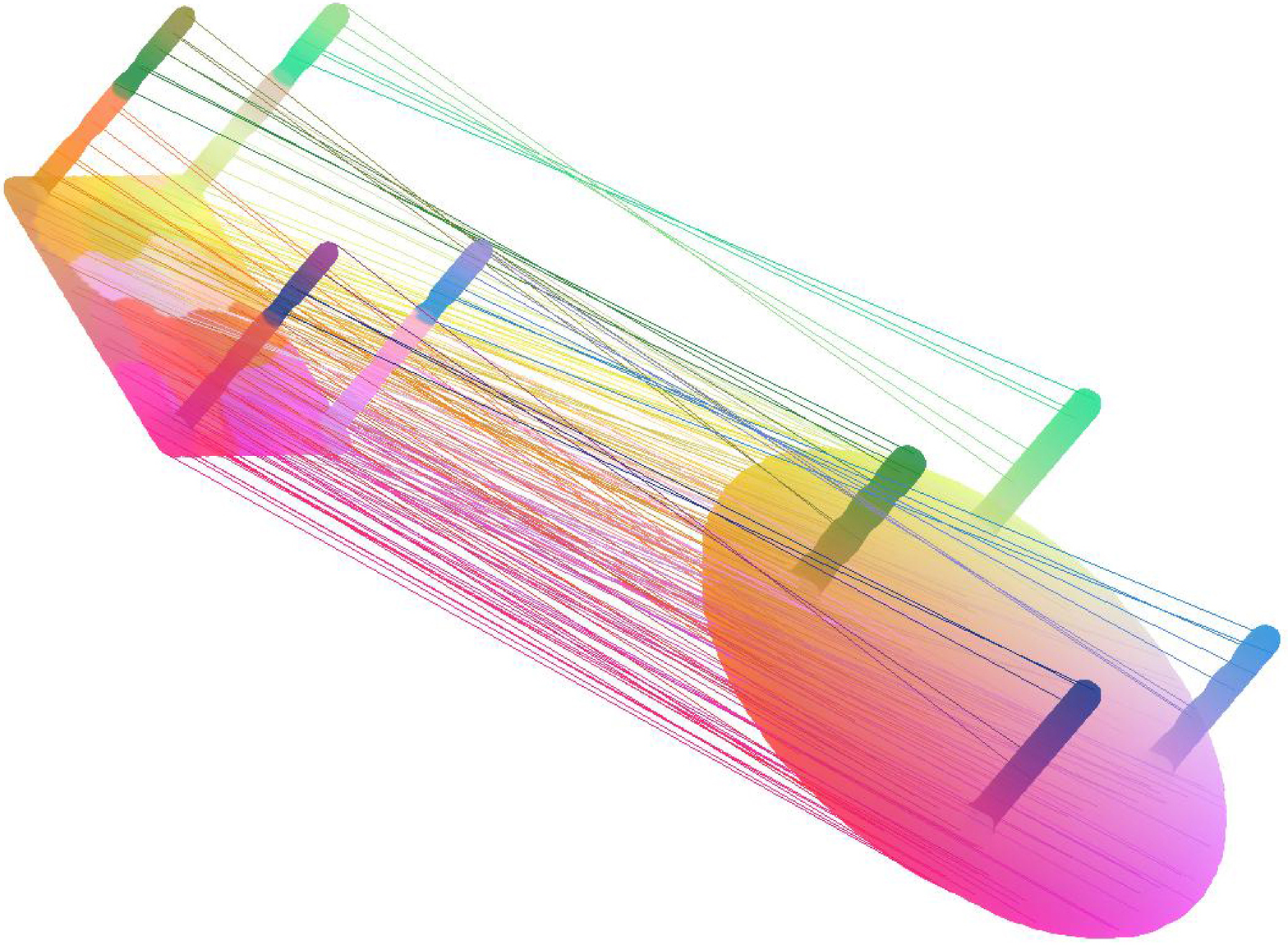}\\
\includegraphics[height=50pt]{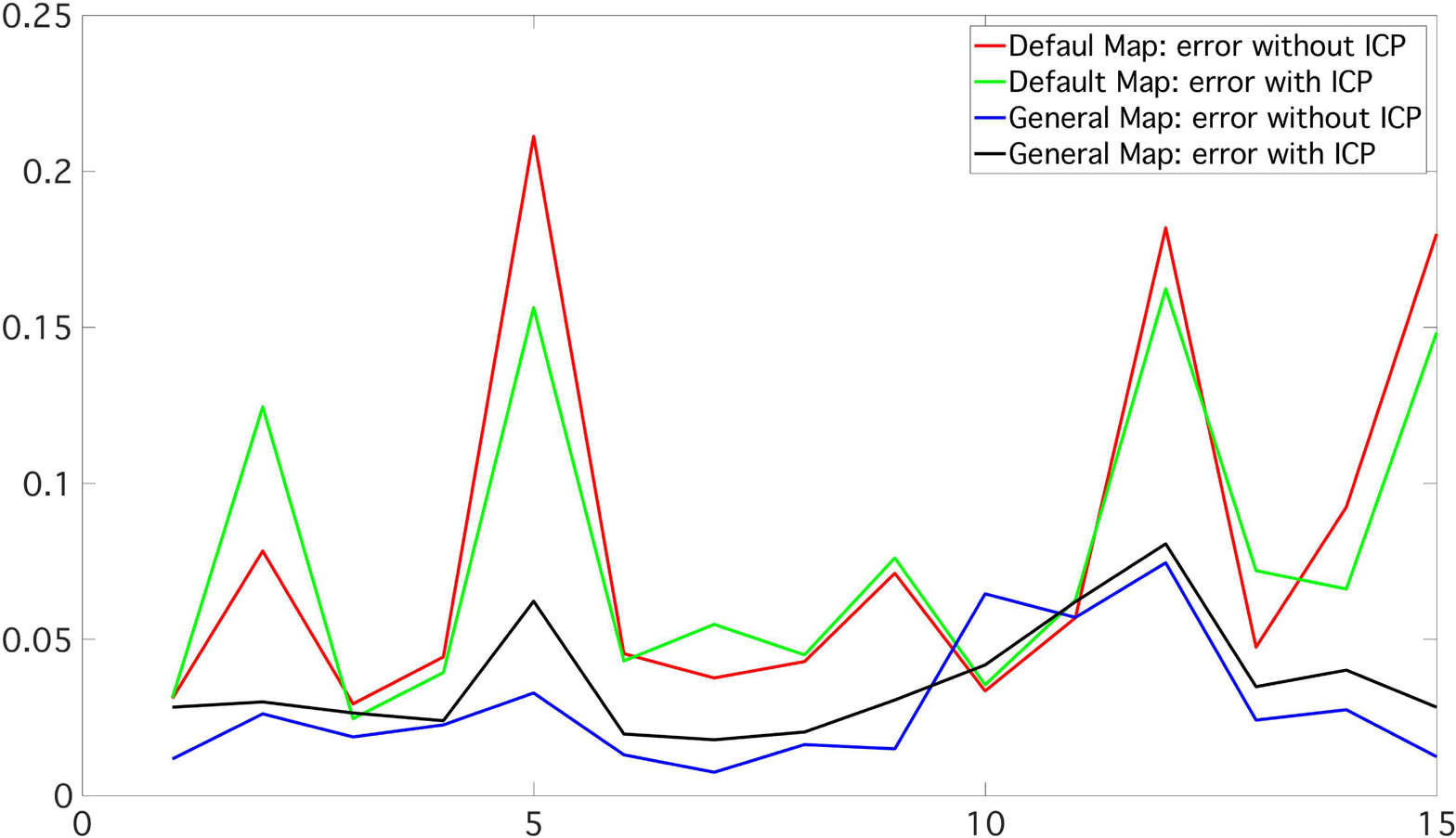}
&\includegraphics[height=40pt]{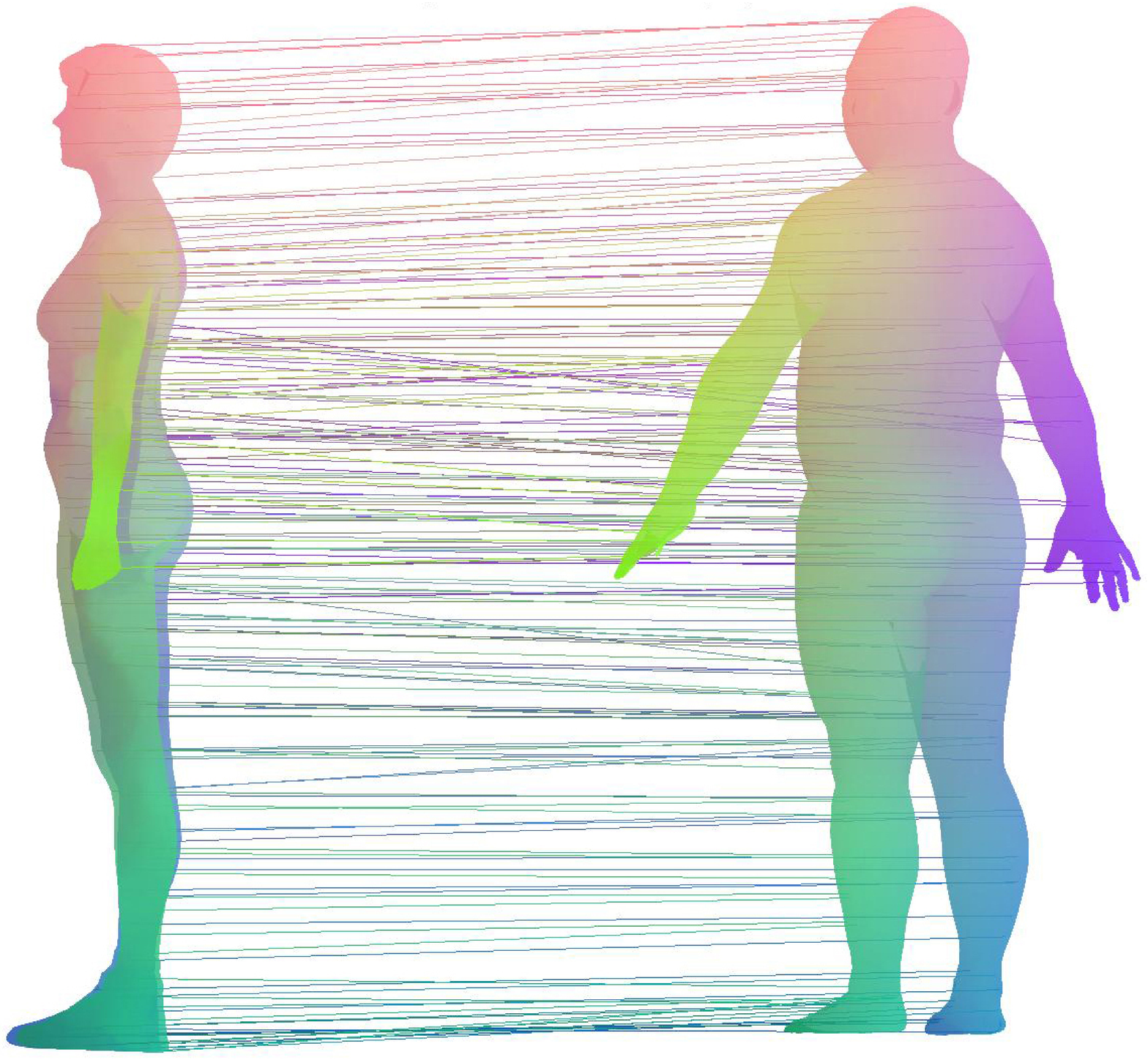}
&\includegraphics[height=40pt]{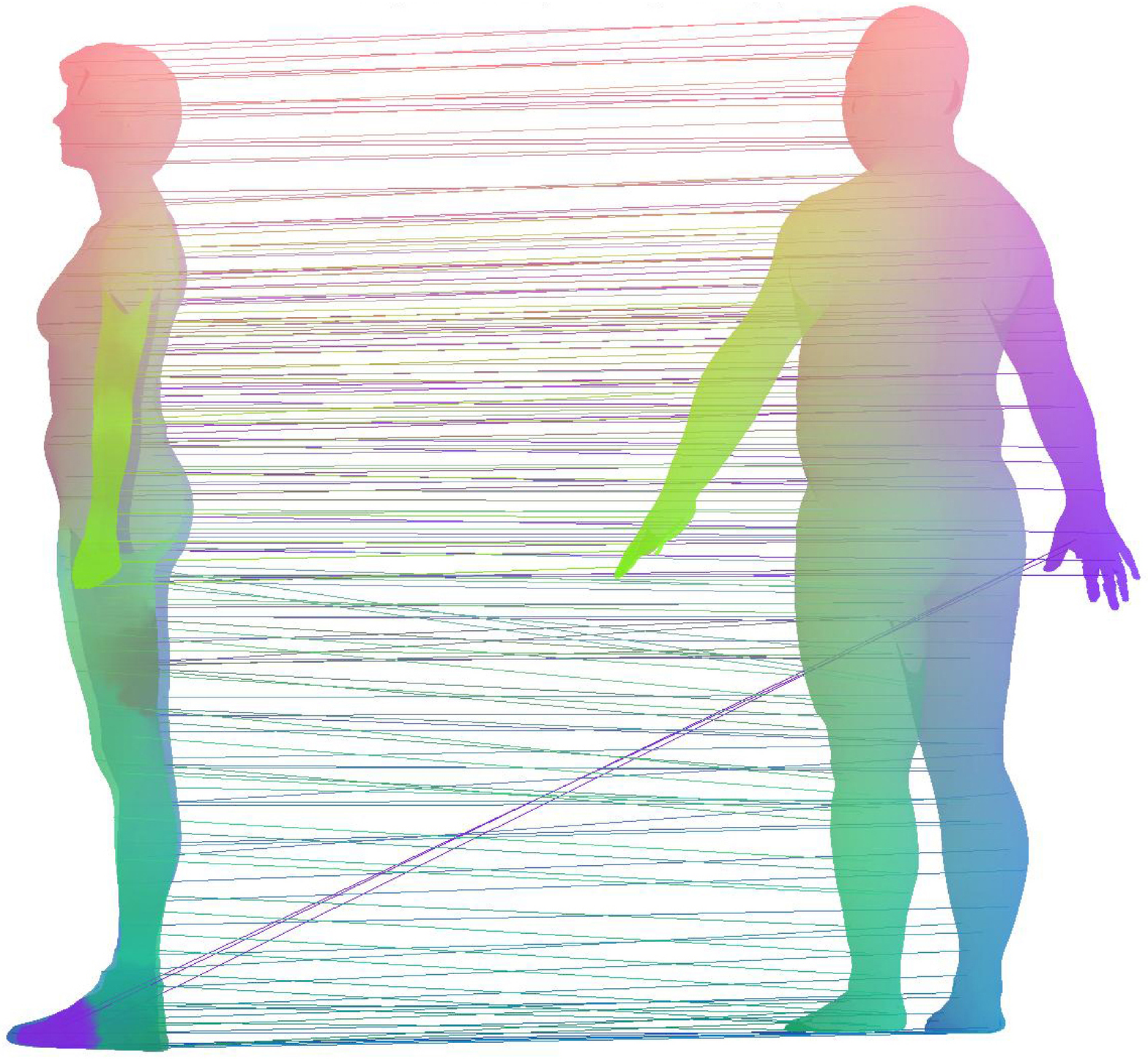}\\
\includegraphics[height=50pt,width=85pt]{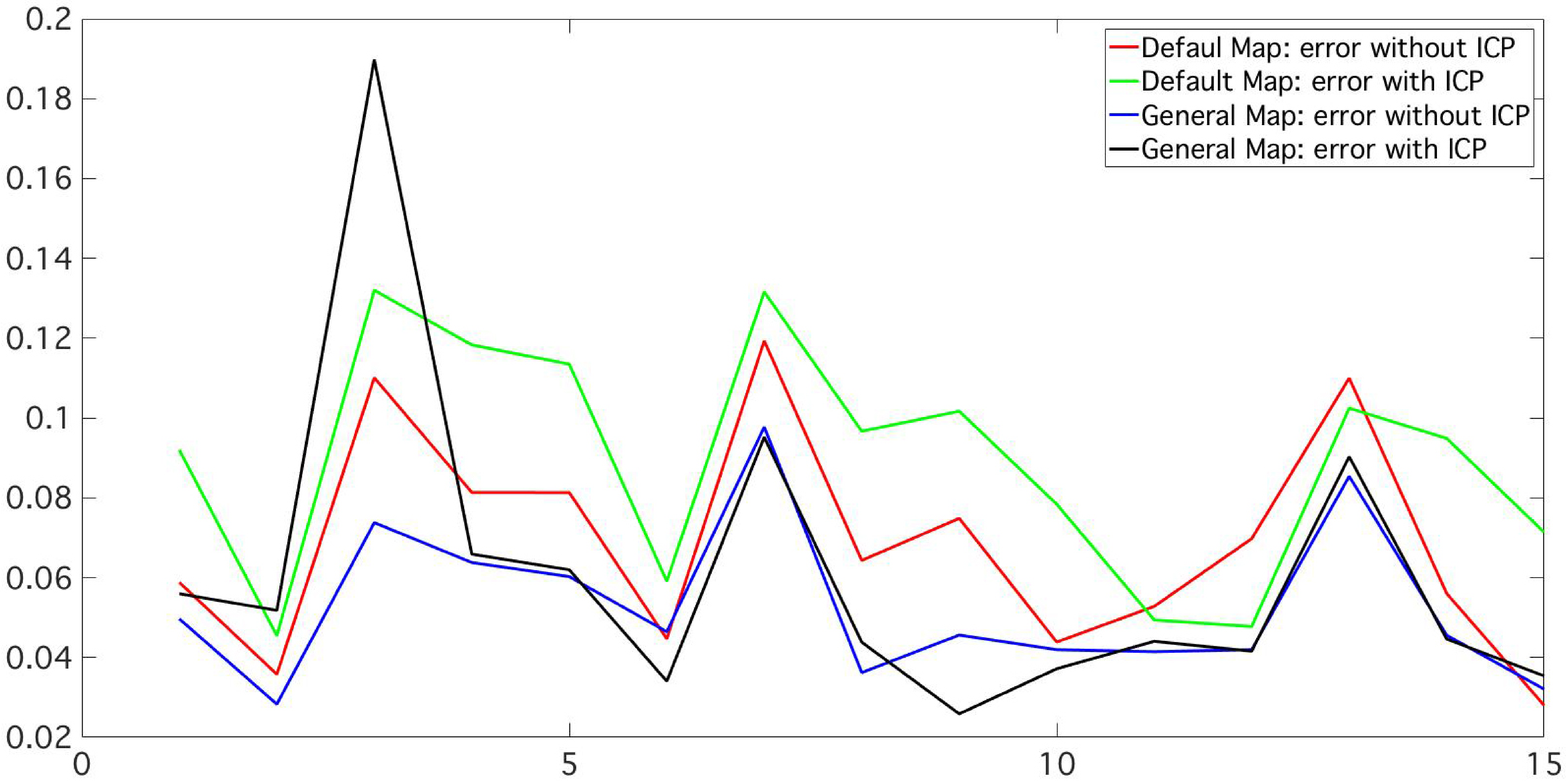}
&\includegraphics[height=30pt]{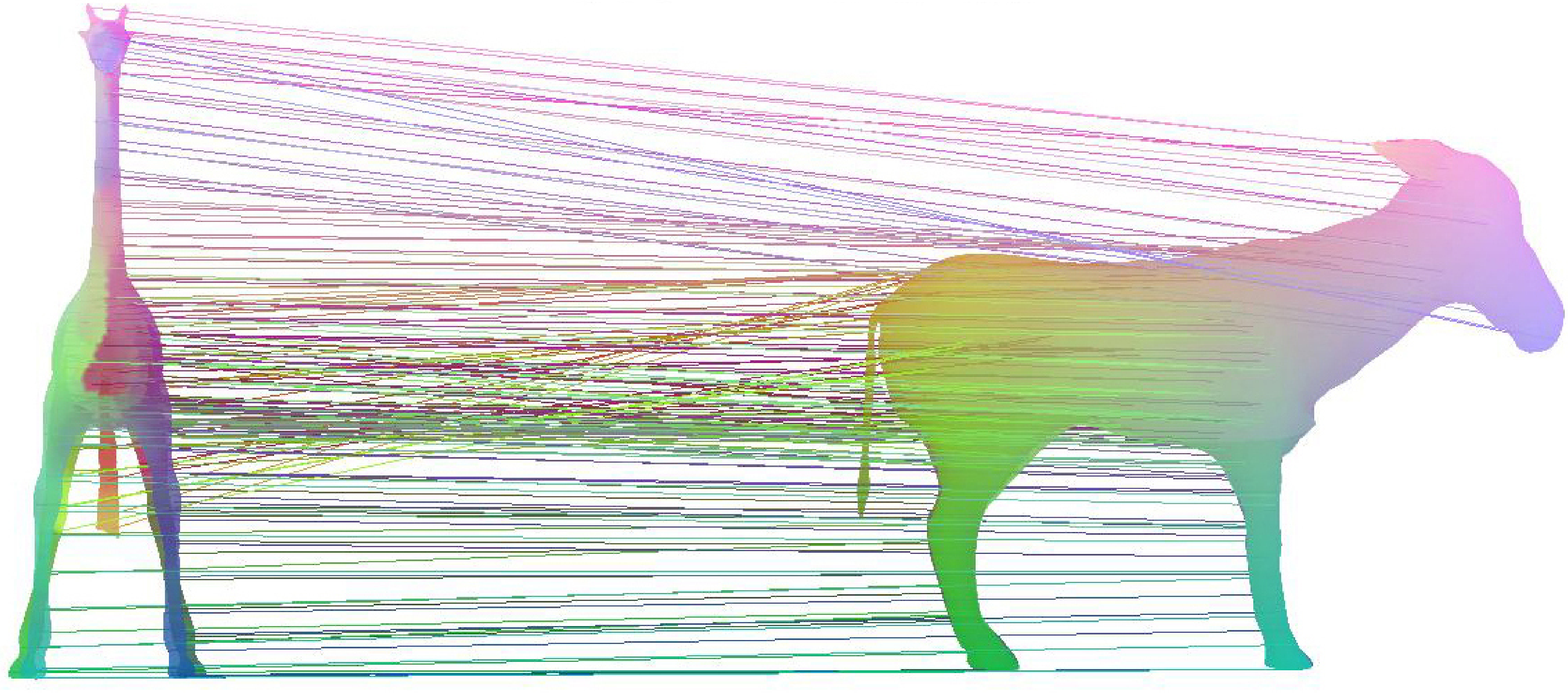}
&\includegraphics[height=30pt]{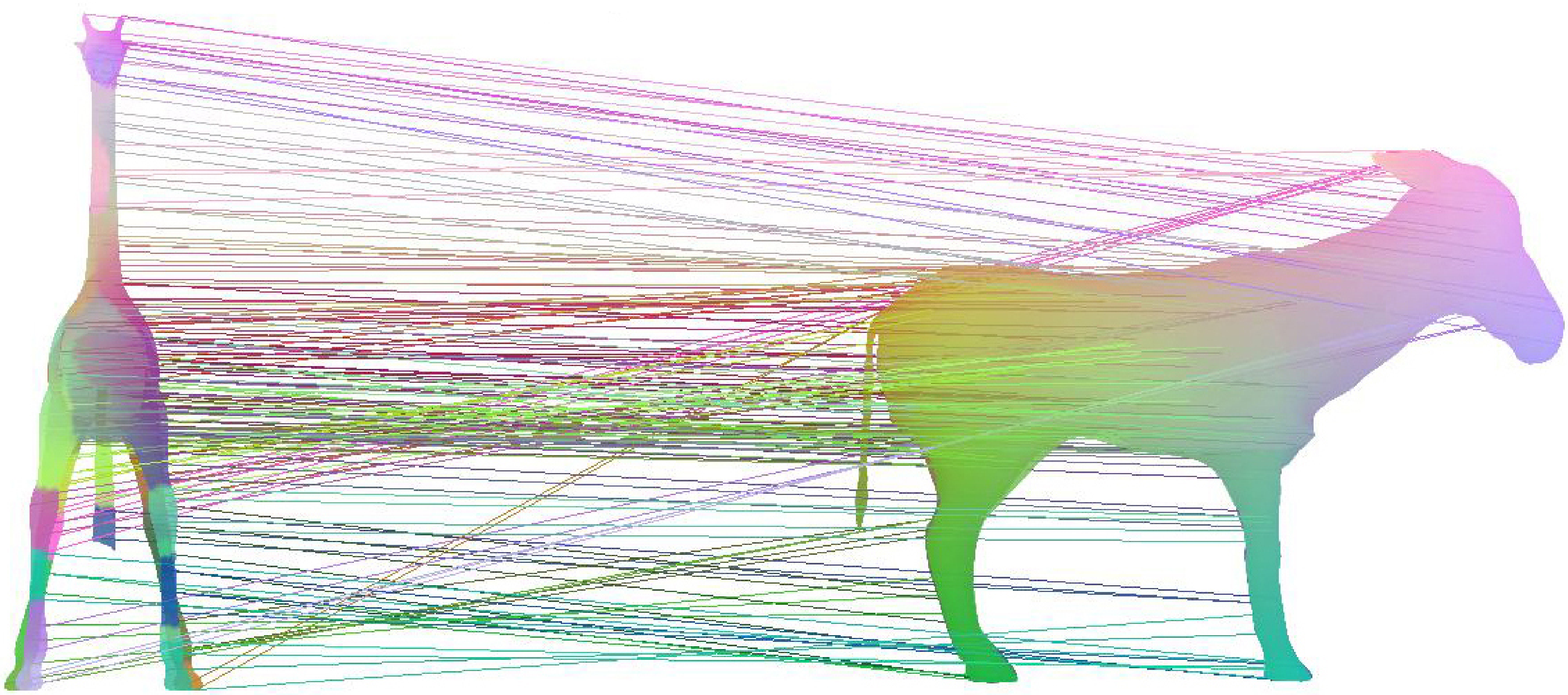}
\end{tabular}
\caption{Mean correspondence error ($y$-axis) on 15 couples ($x$-axis) of non-isometric SHREC'10 shapes~\cite{BRONSTEIN2010-SHREC-B} computed with the same number (i.e., \mbox{$k=60$}) of diffusion functions and Laplacian eigenfunctions. The diffusion functions generally provide a lower correspondence error before/after the correspondence optimisation through the Interior Closest Point (ICP) method and improve the quality of the correspondences, e.g., on the plane wings, the table legs, the legs and the harms, the tail and the legs.\label{fig:SHREC-DIFF-LAPL-SELECTION}}
\end{figure}
\section{Applications\label{sec:DAPPLICATIONS}}
Through Fourier-based and graph filters, we define the \emph{spectral kernels/functions} \mbox{$\Phi(\delta_{\mathbf{p}})=K_{\varphi}(\mathbf{p},\cdot)$} centred at a point~$\mathbf{p}$ as the action of the spectral operator on~$\delta_{\mathbf{p}}$. In this setting, we discuss applications of spectral kernels to \emph{signal reconstruction} and \emph{smoothing} and \emph{shape correspondence}. 

\textbf{Signal reconstruction and smoothing}
In Fig.~\ref{fig:COMPRESSION}, we report the~$\ell_{\infty}$ approximation error of the reconstruction of the geometry ($x$,~$y$,~$z$ coordinates) of different 3D shapes with respect to an increasing number~$k$ ($y$-axis) of regularised harmonic, Laplacian, and diffusion basis functions. The harmonic and diffusion basis functions are centred at~$k$ and~$k/4$ points, sampled with the geodesic farthest point method, and the diffusion basis are computed at 4 scales. The reconstruction error with different classes of spectral functions has an analogous behaviour, thus confirming their meaningfulness for signal approximation. Indeed, harmonic and diffusion functions are a valid alternative to the Laplacian eigenfunctions and have additional properties; in fact, they can be centred at any seed point and diffusion functions have a multi-scale local behaviour.

\textbf{Signal smoothing} For spectral smoothing, we consider a noisy signal \mbox{$\tilde{f}:=f+\delta$} (e.g.,  the~$x$,~$y$, and~$z$ coordinates of the pints on a surface), where~$\delta$ is a Gaussian noise, and a set~$\mathcal{B}$ of diffusion functions centred at \mbox{$k:=100$} samples, evaluated with the farthest point sampling from a seed point, and at \mbox{$s:=6$} scales. Then, we compute a smoothed signal~$g$ as the least-squares projection of~$\tilde{f}$ on~$\mathcal{B}$ and evaluate the corresponding approximation error as \mbox{$\|f-g\|_{\infty}/\|f\|_{\infty}$}. The smoothness order and the approximation accuracy (Fig.~\ref{fig:3TORUS-DIFFUSION-BASIS},~$y$-axis) increase as the number of functions at any selected scale. Best results are achieved by selecting diffusion wavelets at small scales in order to accurately recover the local and global details of the signal.
\begin{figure}[t]
\centering
\begin{tabular}{cc}
SHREC'16 cuts &FARM partial\\
\includegraphics[height=95pt]{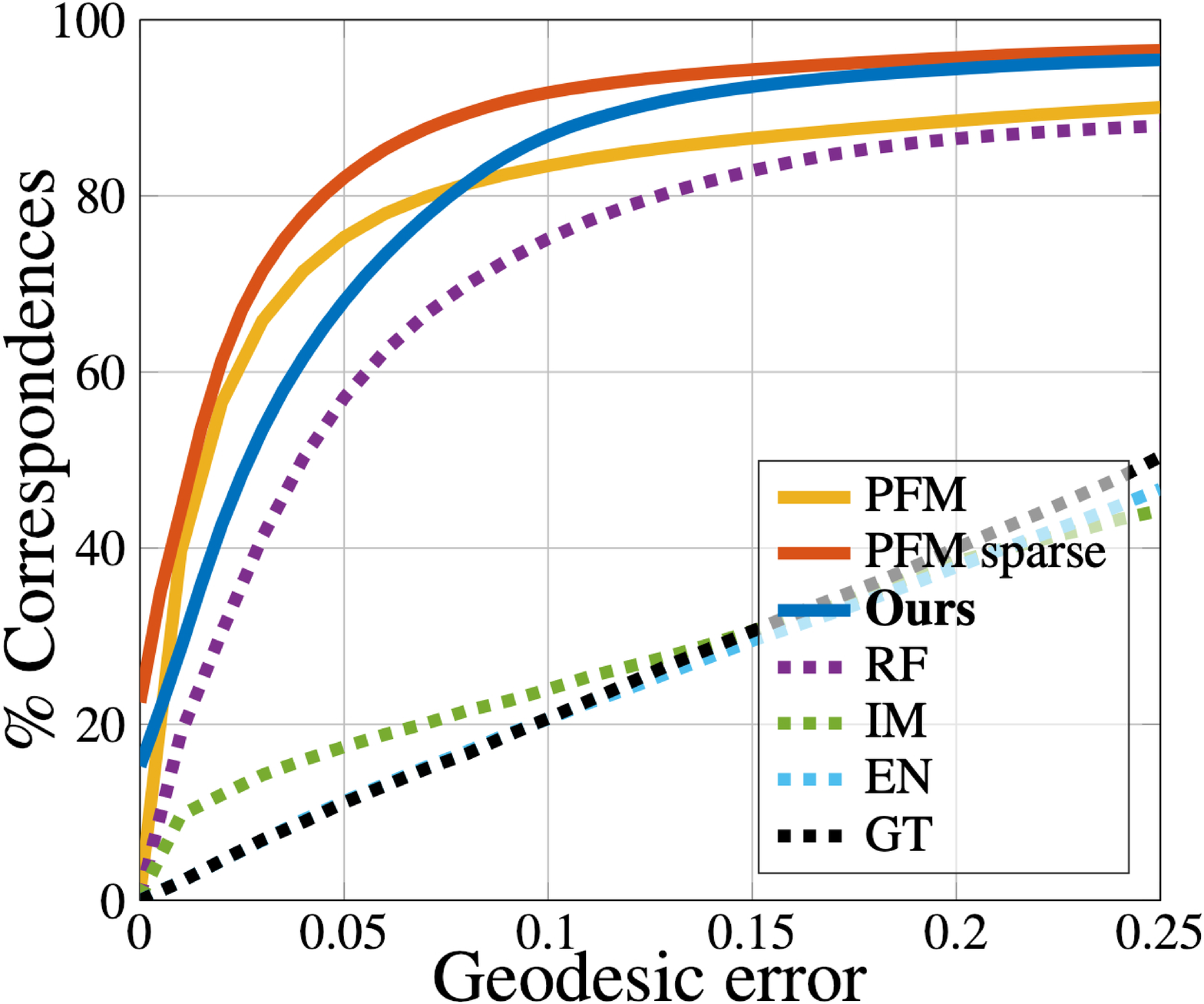}
&\includegraphics[height=95pt]{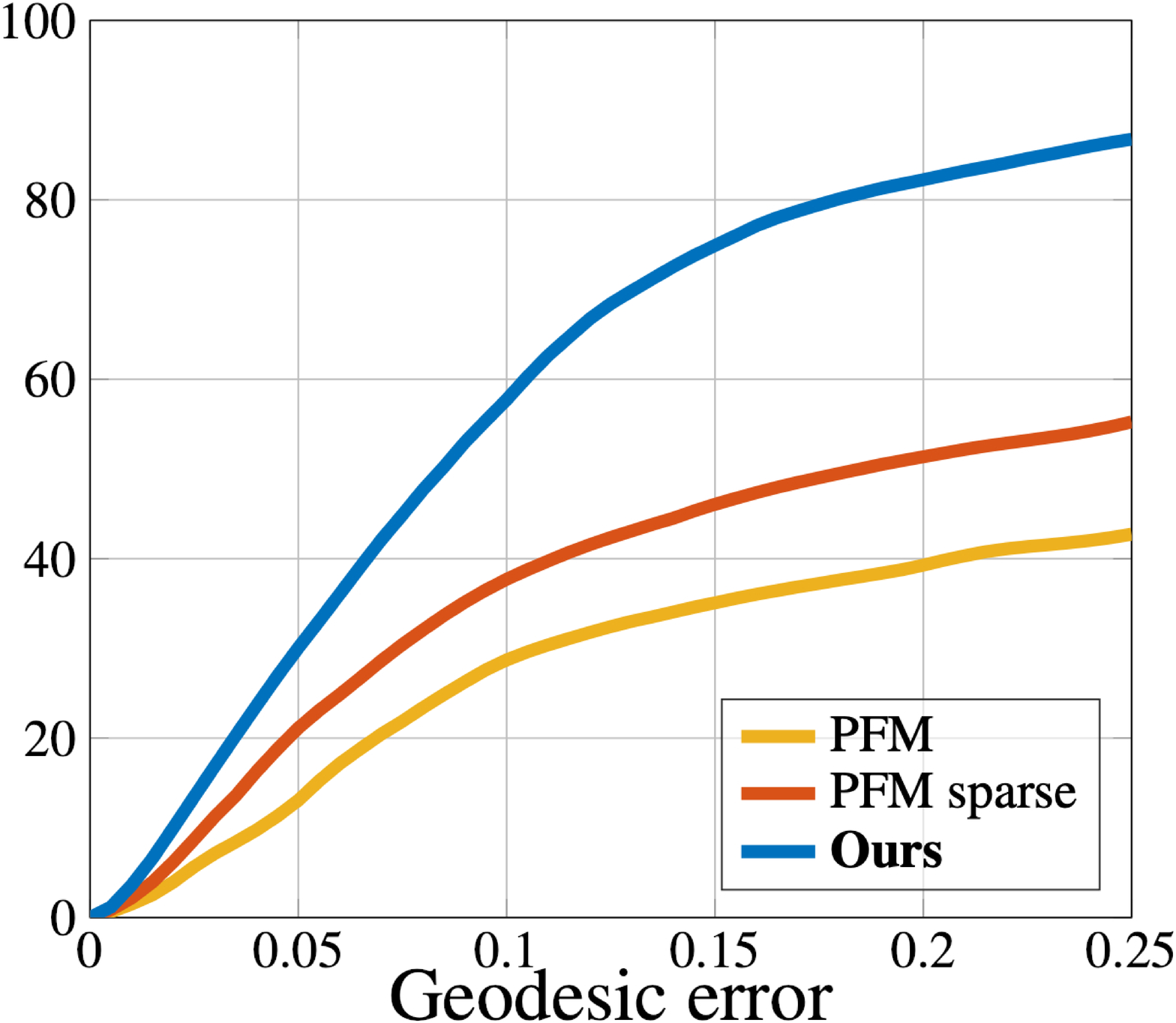}
\end{tabular}
\caption{Quantitative comparison on the SHREC'16 partial cut benchmark~\cite{shrec16partial} (similar connectivity) and on the FARM partial data set~\cite{SHREC19} (different connectivity): \textbf{PFM} (Partial functional maps)~\cite{rodola2017partial,OVSJANIKOV2012}, \textbf{PFM sparse} (PFM initialised with the same sparse correspondence), \textbf{RF} (Random Forest)~\cite{RODOLA2014}, \textbf{IM} (Scale-invariant isometric matching)~\cite{SUN2009}, \textbf{EN} (Elastic Net)~\cite{RODOLA2013}, \textbf{GT} (Game-theoretic matching)~\cite{RODOLA2012}, \textbf{our spectral wavelet} induced by \mbox{$\varphi(s):=\exp(-st)s$}. On the~$x$-axis, we report the mean geodesic distance to the ground truth.\label{fig:partial_quantitative}}
\end{figure}

\textbf{Shape correspondences\label{sec:DISCUSSION-CORRESPONDENCE}}
We now apply specific classes of the spectral kernels to high-level tasks, such as shape correspondence through the functional map framework~\cite{OVSJANIKOV2012}. To this end, we select a low number~$k$ of ground-truth landmarks, which are used to initialise the functional, and consider a set of local and multi-scaled filtered spectral kernels, which generate the sub-space on which the shape descriptors will be projected to define the functional map. Firstly, we compare the quality of the correspondence map computed on the function space generated by (i) \mbox{$k=60$} Laplacian eigenfunctions and (ii) the diffusion functions (i.e., 15 seed points, 4 scales) 5 couples of shapes belonging to 5 classes of the SHREC'10 data set. Indeed, we consider the same number of functions but with different properties in terms of locality and information encoding. For the rigid and articulated shapes (Fig.~\ref{fig:SHREC-DIFF-LAPL-SELECTION}), the computed correspondences correctly map local and global features on the source and target shape. 

To this end, we have selected 7 ground-truth landmarks, 5 uniformly sampled seed points and 4 scales on 15 shapes. The functional map induced by the corresponding 48 diffusion functions has been compared with the one induced by 60 Laplacian eigenfunctions. According to the variation of the mean correspondence error and the examples of correspondences, the diffusion basis functions generally provide a lower correspondence error before and after the optimisation based on the Iterative Closed Point (ICP, for short)~\cite{NOGNENG2017}. Furthermore, the diffusion functions improve the quality of the correspondences with respect to the Laplacian eigenbasis, e.g., between the legs of the giraffe and the tail of the cow, the legs of the giraffe and the horns of the goat, the legs of the dog and the horns of the cow. 

\textbf{Partial shape correspondences}
For comparison with previous work, we focus on partial shape correspondence. As descriptor, we select the spectral kernel \mbox{$K_{\varphi}(\mathbf{p},\cdot):=\mathcal{L}_{\varphi}\delta_{\mathbf{p}}$}, which is induced by spectral convolution operator \mbox{$\Psi_{\varphi}:=\exp(-t\Delta)\Delta$} and the filter \mbox{$\varphi(s):=\exp(-ts)s$}. Indeed, the spectral kernel \mbox{$K_{\varphi}(\mathbf{p},\cdot)$} is achieved by applying the diffusion operator \mbox{$\exp(-t\Delta)$} to the smooth approximation \mbox{$\Delta\delta_{\mathbf{p}}$} of~$\delta_{\mathbf{p}}$, which can be interpreted as a Mexican hat function. Then, the spectral operator is computed through the rational approximation \mbox{$\varphi(s)\approx (sP_{r}(s))/Q_{r}(s)$}, which~$P_{r}/Q_{r}$ \mbox{$(r,r)$} Pad\`e-Chebyshev approximation of thee exponential function. In Fig.~\ref{fig:partial_quantitative}, the spectral wavelet induced by the filter~$\varphi$ has been (i) applied to shape correspondence on partial 3D shapes with a similar (cut data set from SHREC'16~\cite{shrec16partial}) and irregular (FARM partial data set~\cite{SHREC19}) connectivity, and (ii) compared with state-of-the-art methods discussed in~\cite{shrec16partial}. The results achieved with the spectral wavelet are comparable with the best current method~\cite{rodola2017partial} for partial functional maps (PFM) on 3D shape with similar connectivity (Fig.~\ref{fig:partial_quantitative}a); contrary to PFM, our results remain reliable when matching shapes with a highly different connectivity (Fig.~\ref{fig:partial_quantitative}a). For more details, we refer the reader to our recent work presented in~\cite{KIRGO2020}.

\section{Conclusions and future work\label{sec:DISCUSSION}}
This paper has discussed the definition of novel Fourier-based and rational spectral operators for graph processing, which generalise the notion of polynomial spectral filters and Fourier transform to non-Euclidean domains. As future work, we plan to investigate the usefulness of rational filters for the definition of a family of spectral bases for data analysis. In fact, several signal processing techniques generally represent the data in terms of a given basis in order to highlight a given class of underlying properties or features, e.g., localising content in both space and frequency through wavelet basis. Rational filters, and in particular rational Chebyshev polynomials, are particularly useful to enlarge the class of learning networks, as a generalisation of networks based on polynomial filters, such as PolyNet~\cite{ZHANG2017}, ChebNet~\cite{KIPF2016}, CayleyNet~\cite{LEVIER2019}, in order to improve the discriminative capabilities of networks in the context of 3D geometric deep learning.\\

\paragraph*{\textbf{Acknowledgments}}
We thank the Reviewers for their thorough review and constructive comments, which helped us to improve the technical part and presentation of the paper. Graphs are courtesy of the ``\emph{Gaimc: Graph Alg.}'' Library. 
%

%
\begin{IEEEbiography}
[{\includegraphics[width=1in,height=1.10in,clip,keepaspectratio]{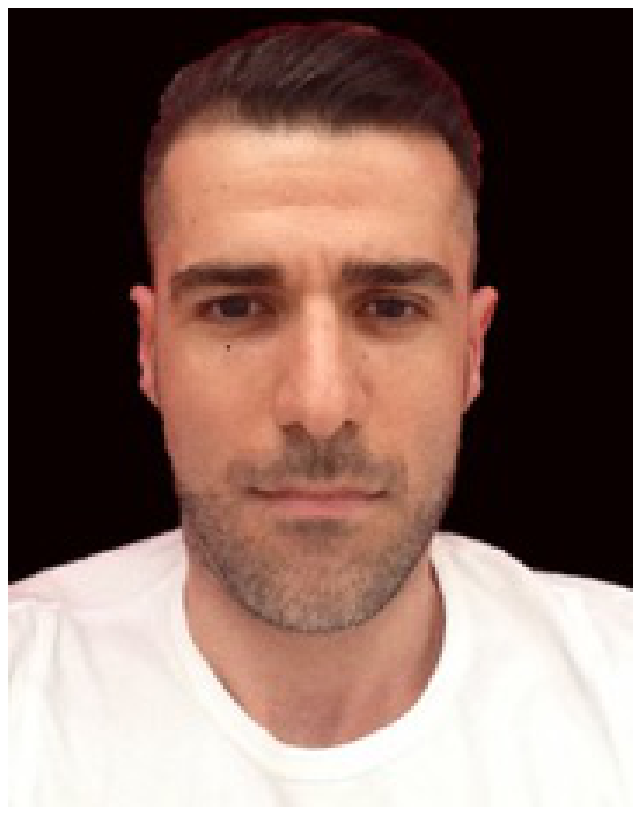}}]{Giuseppe Patan\`e}
is senior researcher at CNR-IMATI. Since 2001, his research is mainly focused on Data Science. He obtained the National Scientific Qualification as Full Professor of Computer Science. He is author of scientific publications on international journals and conference proceedings, and tutor of Ph.D. and Post.Doc students. He is responsible of R$\&$D activities in national and European projects.
\end{IEEEbiography}
\end{document}